%% file: paper.tex
\documentclass[lettersize,journal]{IEEEtran}

\RequirePackage[normalem]{ulem} 
\RequirePackage{color}\definecolor{RED}{rgb}{1,0,0}\definecolor{BLUE}{rgb}{0,0,1} 

\usepackage[pdftex]{graphicx}
\graphicspath{{images}}
\DeclareGraphicsExtensions{.pdf,.jpeg,.png}

\usepackage{amsmath}
\usepackage{array}

\usepackage[font=small]{caption}
\usepackage{subcaption}
\usepackage{setspace}
\let\labelindent\relax
\usepackage{enumitem}
\usepackage{url}
\usepackage{amssymb}
\usepackage{bm}
\usepackage{graphicx}
\usepackage{float}
\usepackage[normalem]{ulem}
\usepackage{listings}
\usepackage{gensymb}
\usepackage{array}
\usepackage{pbox}
\usepackage{mathtools}
\usepackage{breqn}
\usepackage[ruled,vlined]{algorithm2e}
\usepackage{resizegather}
\usepackage{changepage}
\usepackage{titlesec}
\usepackage{geometry}
\usepackage{nicefrac}
\usepackage{cases}
\usepackage{makecell}
\usepackage{nicematrix}
\usepackage{orcidlink}
\DeclareFontEncoding{LS1}{}{}
\DeclareFontSubstitution{LS1}{stix2}{m}{n}
\DeclareSymbolFont{letters-candra}{LS1}{stix2}{m}{it}
\DeclareMathAccent{\candra}{\mathalpha}{letters-candra}{"8B}

\titlespacing\section{0pt}{10pt minus 2pt}{5pt minus 2pt}
\titlespacing\subsection{0pt}{10pt minus 2pt}{5pt minus 2pt}
\newlength{\offsetpage}
\newenvironment{widepage}{\begin{adjustwidth}{-\offsetpage}{-\offsetpage}%
    \addtolength{\textwidth}{2\offsetpage}}%
{\end{adjustwidth}}
\usepackage{xpatch}
\xpretocmd{\algorithm}{\hsize=\linewidth}{}{}
\graphicspath{
    {images/}
}

\newcommand{\diag}{\mathrm{diag}}
\newcommand{\figref}{Fig.~\ref}
\newcommand{\tabref}{Table~\ref}

\newcommand{\mbs}{\bm}
\newcommand{\mbf}{\mathbf}

\newcommand{\bbm}{\begin{bmatrix}}
\newcommand{\ebm}{\end{bmatrix}}
\newcommand{\col}{\mathbin{:}}
\DeclareMathOperator*{\argmin}{\arg\!\min}
\usepackage[T3,T1]{fontenc}
\DeclareSymbolFont{tipa}{T3}{cmr}{m}{n}
\DeclareMathAccent{\invbreve}{\mathalpha}{tipa}{16}

\newcolumntype{L}{>{$}l<{$}}

\usepackage{xcolor}

\newcommand \reviewcomment[1]{{#1}}
\newcommand \reviewtemp[1]{}
\newcommand \reviewdelmath[1]{}  
\newcommand \reviewdel[1]{}  

\newcommand\highlightReference[1]{%
  \expandafter\newcommand\csname highlightReference-#1\endcsname{}%
}
\let\oldbibitem\bibitem
\def\bibitem#1 #2\par{%
  \expandafter\ifx\csname highlightReference-#1\endcsname\relax
    \oldbibitem{#1}#2\par
  \else
    \oldbibitem{#1}\reviewcomment{#2}\par
  \fi
}

\makeatletter
\newcommand{\removelatexerror}{\let\@latex@error\@gobble}
\makeatother

\begin{document}

\title{Data-Driven Batch Localization and SLAM Using Koopman Linearization}

\author{{Zi~Cong~Guo}~\orcidlink{0000-0002-2789-007X},~\IEEEmembership{Student Member,~IEEE}, {Frederike~Dümbgen}~\orcidlink{0000-0002-7258-9753},~\IEEEmembership{Member,~IEEE}, {James~R.~Forbes}~\orcidlink{0000-0002-1987-9268},~\IEEEmembership{Member,~IEEE}, {Timothy~D.~Barfoot}~\orcidlink{0000-0003-3899-631X},~\IEEEmembership{Fellow,~IEEE} \vspace{-30pt}
\thanks{Manuscript received February 13, 2024; accepted July 17, 2024. This paper was recommended for publication by Editor Javier Civera upon evaluation of the reviewers' comments. This work was funded in part by the Swiss National Science Foundation, Postdoc Mobility under Grant 206954, and in part by the Natural Sciences and Engineering Research Council of Canada (NSERC).}%
\thanks{Zi Cong Guo, Frederike Dümbgen, and Timothy D. Barfoot are with the University of Toronto Robotics Institute, Toronto, Ontario, Canada (email: zc.guo@mail.utoronto.ca; frederike.dumbgen@utoronto.ca; tim.barfoot@utoronto.ca).}
\thanks{James R. Forbes is with the Department of Mechanical Engineering, McGill University, Montreal, Quebec, Canada (email: james.richard.forbes@mcgill.ca).}
}
\maketitle
\input{sections/0_Abstract}
\IEEEpeerreviewmaketitle
\input{sections/1_Intro}
\input{sections/2_RelatedWork}
\input{sections/3_LiftingTheSystem}
\input{sections/4_SystemIdentification}
\input{sections/5_UKL}
\input{sections/6_CKL}

\input{sections/7_RCKL}
\input{sections/8_ExperimentsResults}
\input{sections/9_Conclusion}


\highlightReference{dahdah_pykoop_2022}
\highlightReference{pysindy}
\highlightReference{hyper-opt}
\highlightReference{arXiv-version-rckl}

\bibliographystyle{IEEEtran}
{
\singlespacing
\bibliography{refs}
}
\input{sections/Appendix_arxiv}

\end{document}

%% file: sections/0_Abstract.tex
\begin{abstract}
We present a framework for model-free batch localization and SLAM. We use lifting functions to map a control-affine system into a high-dimensional space, where both the process model and the measurement model are rendered bilinear. During training, we solve a least-squares problem using groundtruth data to compute the high-dimensional model matrices associated with the lifted system purely from data. At inference time, we solve for the unknown robot trajectory and landmarks through an optimization problem, where constraints are introduced to keep the solution on the manifold of the lifting functions. The problem is efficiently solved using a sequential quadratic program (SQP), where the complexity of an SQP iteration scales linearly with the number of timesteps. Our algorithms, called Reduced Constrained Koopman Linearization Localization (RCKL-Loc) and Reduced Constrained Koopman Linearization SLAM (RCKL-SLAM), are validated experimentally in simulation and on two datasets: one with an indoor mobile robot equipped with a laser rangefinder that measures range to cylindrical landmarks, and one on a golf cart equipped with RFID range sensors. We compare RCKL-Loc and RCKL-SLAM with classic model-based nonlinear batch estimation. While RCKL-Loc and RCKL-SLAM have similar performance compared to their model-based counterparts, they outperform the model-based approaches when the prior model is imperfect, showing the potential benefit of the proposed data-driven technique.

\end{abstract}

%% file: sections/1_Intro.tex
\section{Introduction}
\IEEEPARstart{S}TATE estimation, and simultaneous localization and mapping (SLAM) in particular, are of prime importance for many robotics systems. Most state-estimation methods rely on complex modelling and/or problem-specific techniques, and the procedure often involves linearization. Moreover, when process or measurement models are inaccurate, or even unavailable, model-based state estimation methods can struggle to produce accurate and consistent navigation results. For many applications~\cite{uwb-loc},~\cite{rfid-loc},~\cite{manipulator-model}, the procedure for modelling and/or solving the estimation problem is much more involved than the data-gathering process.

\begin{figure}[!t]
 \begin{widepage}
 \centering
\includegraphics[width=0.99\columnwidth,trim={0 0 0 0},clip]{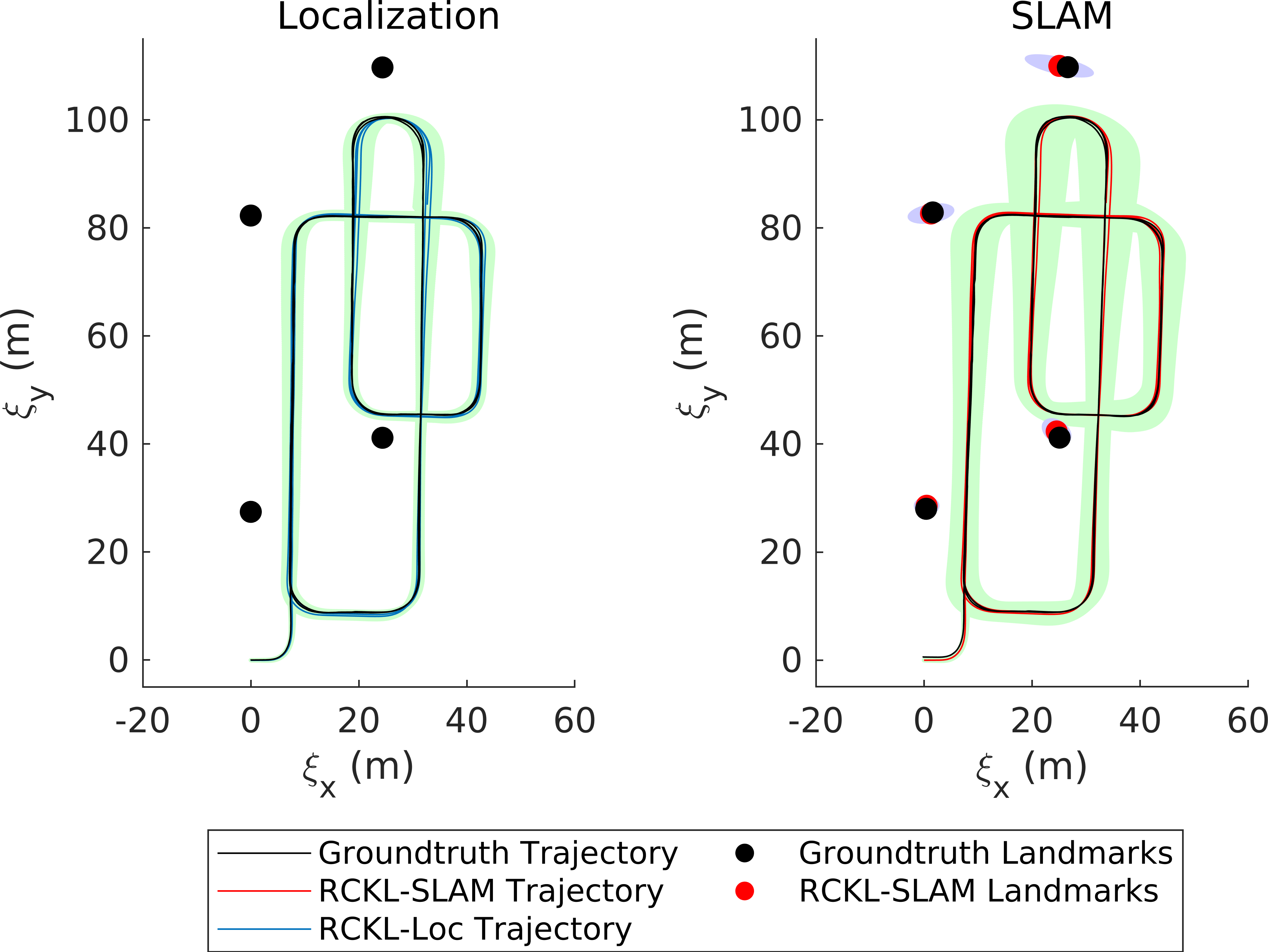}
\end{widepage}
\caption{Visualization of the trajectory output of RCKL-Loc (left), and the trajectory and landmark output of RCKL-SLAM (right), on Experiment 2 described in Section \ref{sec:exp-golf-cart}, showing the estimators' mean states and mean landmark positions compared to the groundtruth. The green regions and the grey regions are, respectively, the $3\sigma$ covariances of the trajectory and of the landmarks. The estimators' trajectories and landmarks are close to the groundtruth and are within the estimated $3\sigma$ bounds, despite
the algorithm having no prior knowledge of the system models.}
\label{fig:gesling_redkooploc_redkoopslam}
\end{figure}
In this work, we propose the Reduced Constrained Koopman Linearization (RCKL) framework for data-driven localization and SLAM. The method
\begin{itemize}
 \item learns a high-dimensional (bi)linear system model from data without requiring prior models,
 \item is applicable to systems with control-affine dynamics,
 \item solves for the unknown states and landmark positions with an inference cost that scales linearly with the number of timesteps, and
 \item has a training cost that scales linearly with the amount of data and an inference cost that is independent of the amount of training data. 
\end{itemize}

RCKL is an extension of the Koopman State Estimator (KoopSE)~\cite{kooplin}, a framework for data-driven localization where the landmarks are at the same positions at test time as during training. RCKL extends KoopSE by allowing for localization with landmarks at different positions at test time, and allowing for SLAM when the landmarks are unknown at test time. Specifically, the main novelties of RCKL over KoopSE are (i) a formulation that allows for the joint estimation of poses and landmarks, (ii) the introduction of constrained optimization over the manifolds defined by the lifting functions, and (iii) a formulation of a Sequential Quadratic Program (SQP)~\cite{nocedal} to efficiently perform this optimization. Given a robot with unknown nonlinear dynamics and measurements, we can first learn its high-dimensional model in a factory setting with groundtruth, and then deploy the robot in the field and use RCKL-SLAM to map the new landmarks. Once the landmarks are known, we can then potentially use RCKL-Loc to perform batch (or recursive) localization against the map.

This paper is structured as follows. We review related work in Section~\ref{sec:related-work}, summarize themes in our notation in Section~\ref{sec:notation}, and discuss system lifting with Koopman linearization in Sections~\ref{sec:koopman-lift} and~\ref{sec:prelim}. We outline the model-learning procedure in Section~\ref{sec:system-id}. We then build up the design of RCKL through Sections~\ref{sec:batch-est}-\ref{sec:redkoop}, where we discuss the procedure and limitations of Unconstrained Koopman Linearization (UKL) in Section~\ref{sec:batch-est}, Constrained Koopman Linearization (CKL) in Section~\ref{sec:constrained-koopman}, and finally RCKL in Section~\ref{sec:redkoop}. We present experimental results in Section~\ref{sec:experiments} and conclude in Section~\ref{sec:discussion_conclusion}. 

%% file: sections/2_RelatedWork.tex
\section{Related Work}
\label{sec:related-work}
Two common concepts for data-driven algorithms are kernel embeddings~\cite{Hofmann} and the Koopman operator~\cite{Koopman}, both used to transform a system into a simpler form in a high-dimensional space. Kernel methods lift the system by embedding probability distributions in high-dimensional spaces~\cite{song}, and there has been some work on kernelized state estimation and SLAM for limited model types using Gaussian processes~\cite{GPBayes},~\cite{wifi-slam} and kernel embeddings~\cite{ksmoother}. However, pure kernel methods scale poorly, typically cubically with the amount of training data. Koopman-based methods, on the other hand, work by lifting the variables directly, and both the model-learning process and the inference process typically scale well with data~\cite{dmd-big-book-compact}. There is some work on Koopman-based state estimation, validated for small problems in simulation~\cite{koopman-const-convex},~\cite{koopman-nonlinear-se}. However, validations on real-world datasets are limited, and efficient learning and inference techniques have yet to be developed for large-scale state-estimation problems such as SLAM. Rather, most of the attention on Koopman in robotics has been on filtering~\cite{koopman-estimation-power-systems} or model-predictive control for nonlinear systems~\cite{koopman-control},~\cite{mauroy_2020_koopman},~\cite{koopman-mpc}, which are applications with short time horizons. In general, there is limited work on using the Koopman operator to lift and solve large batch optimization problems, including SLAM.

There is a large body of work on system identification with kernels~\cite{kernel-system-id}, \cite{rkhs-inverse-dynamics} and Koopman techniques~\cite{reproject-eg}, \cite{koopman-system-id}, \cite{koopman-system-id-soft-robot}. These methods solve linear-like optimization problems a lifted space and typically have better convergence properties than classic methods. However, there is little work on actually employing the identified lifted models for state estimation. Converting them back into the original state space and using classic nonlinear estimation methods would still be challenging, especially if the models are noisy or high dimensional~\cite{barfoot-txtbk}. Instead, we seek a data-driven method that both identifies the models and employs them for estimation efficiently in the lifted space.

One method for data-driven state estimation is KoopSE~\cite{kooplin}, which lifts a control-affine system such that standard batch linear-Gaussian state estimation methods can be used. KoopSE is validated on a mobile robot dataset. However, KoopSE incorporates all of the measurements as part of one large model for the whole environment. Thus, when carrying out localization, the landmarks are required to be at the same positions at test time as during training. With no explicit specification for the landmarks, this formulation also cannot be used for SLAM. This paper aims to begin filling this gap by formulating data-driven bilinear localization and SLAM for control-affine systems.

%% file: sections/3_LiftingTheSystem.tex
\vspace{-5pt}
\section{Notation}
\label{sec:notation}
We denote matrices with capital boldface letters, $\mbf{A}$, vectors with lowercase boldface letters, $\mbf{a}$, and scalars with normal-faced letters, $a$. We use $\text{diag}(\mbf{A}_1,\dots,\mbf{A}_N)$ to denote the block-diagonal matrix with the blocks being $\mbf{A}_1,\dots,\mbf{A}_N$. We denote quantities in the original, unlifted space with Greek letters, $\mbs{\xi}$, and quantities in the lifted space with Roman letters, $\mbf{x}$. We use the same letter of the two alphabets to denote connected quantities whenever possible, e.g., $\mbs{\xi}$ for the original state and $\mbf{x}$ for the lifted state. For the lifted quantities, we denote batch terms with italics, $\mbs{x}$, and non-batch terms without italics.

\section{Koopman Lifting of Process Models}
\label{sec:koopman-lift}
In this section, we briefly introduce the concept of Koopman lifting~\cite{Koopman} of discrete process models. For a comprehensive review of the Koopman operator, see~\cite{dmd-big-book-compact}, \cite{mauroy_2020_koopman}.

The Koopman framework's main idea is to lift a nonlinear autonomous system into a high-dimensional space where the system becomes linear. Suppose a noiseless autonomous system (i.e., no inputs) is governed by $\mbs{\xi}_k = \mbf{f}(\mbs{\xi}_{k-1})$, where $\mbs{\xi}_k$ is the state at timestep $k$ and $\mbf{f}(\mbs{\xi})$ is the process model. Then, there exists an embedding, $\mbf{p}$, and a linear operator (the Koopman operator), $\mbs{\mathcal{K}}$, such that $\mbf{p}(\mbs{\xi}_k) = \mbs{\mathcal{K}} \mbf{p}(\mbs{\xi}_{k-1})$. We call $\mbf{x}_k = \mbf{p}(\mbs{\xi}_k)$ the embedded state. The lifted system dynamics become $\mbf{x}_k = \mbs{\mathcal{K}} \mbf{x}_{k-1}$, which is linear. Although a finite-dimensional $\mbs{\mathcal{K}}$ and $\mbf{p}$ may be hard (or impossible; see~\cite{koop-inv-with-sindy}) to find, this transformation always exists. As an example from~\cite{koop-inv-with-sindy}, suppose that a 2-dimensional system is governed by the nonlinear dynamics
\begin{gather}
 \bbm \xi_{1,k} \\ \xi_{2,k} \ebm
 =
 \mbf{f}\left(\bbm \xi_{1,k-1} \\ \xi_{2,k-1} \ebm \right)
 =
 \bbm \lambda \xi_{1,k-1} \\ \mu\xi_{2,k-1} + (\lambda^2-\mu) \xi_{1,k-1}^2 \ebm,
\end{gather}
for some parameters $\lambda$ and $\mu$. In this case, the nonlinear system can be transformed into a linear system with the finite-dimensional lifting function $\mbf{p}(\mbs{\xi}_k) = \bbm \xi_{1,k} & \xi_{2,k} & \xi_{1,k}^2 \ebm^T = \bbm x_{1,k} & x_{2,k} & x_{3,k} \ebm^T$, yielding
\begin{gather}
 \bbm x_{1,k} \\ x_{2,k} \\ x_{3,k} \ebm
 =
 \bbm \lambda & 0 & 0 \\ 0 & \mu & \lambda^2-\mu \\ 0 & 0 & \lambda^2 \ebm
 \bbm x_{1,k-1} \\ x_{2,k-1} \\ x_{3,k-1} \ebm.
\end{gather}
In cases where an exact finite-dimensional Koopman operator cannot be found, using a truncated approximation can still be reasonable~\cite{koop-inv-with-sindy}. This concept can be generalized to non-autonomous systems~\cite{koopman-dmdc}. However, there is limited work on identifying and using the Koopman operator for estimation in the presense of noise, inputs, and measurements. In the next section, we generalize the Koopman lifting framework to our systems of interest, which include noisy process and measurement models.

\section{Lifting the Full System}
\label{sec:prelim}
We focus on systems with process and measurement models of the form
\begin{subequations}\label{eq:control-affine}
\begin{align}
 \mbs{\xi}_{k} &= \mbf{f}(\mbs{\xi}_{k-1}, \mbs{\nu}_k, \mbs{\omega}_k) \\
 &= \mbf{f}_0(\mbs{\xi}_{k-1}) + \sum_{i=1}^{N_\nu} \mbf{f}_i(\mbs{\xi}_{k-1})\nu_{k,i} + \mbs{\omega}_k, \label{eq:control-affine-1}\\
 \mbs{\gamma}_{k,j} &= \mbf{g}(\mbs{\xi}_k, \mbs{\psi}_j, \mbs{\eta}_{k,j}) \\
 &= \sum_{i=1}^{N_{g,\xi}} \sum_{m=1}^{N_{g,\psi}} \mbf{g}_{\xi,i} (\mbs{\xi}_k) {\pi}_{\psi,m} (\mbs{\psi}_j) + \mbs{\eta}_{k,j}, \label{eq:control-affine-2}
\end{align}
\end{subequations}
where $\mbs{\xi}_k \in \mathbb{R}^{N_\xi}$ is the robot state, $\mbs{\nu}_k \in \mathbb{R}^{N_\nu}$ the control input, $\mbs{\omega}_k \in \mathbb{R}^{N_\omega}$ the process noise, all at timestep $k$. $\mbs{\psi}_j \in \mathbb{R}^{N_\psi}$ is the position of landmark $j$, $\mbs{\gamma}_{k,j} \in \mathbb{R}^{N_\gamma}$ the measurement of $\mbs{\psi}_j$ at timestep $k$, and $\mbs{\eta}_{k,j} \in \mathbb{R}^{N_\eta}$ the measurement noise for $\mbs{\gamma}_{k,j}$. $\mbf{f}_i$ are the various components of the process model, and $\mbf{g}_{\xi,i},{\pi}_{\psi,m}$ are the various components of the measurement model. We have chosen to focus on systems with a control-affine process model~\eqref{eq:control-affine-1} since the model can be written exactly as a lifted bilinear model when the system is noiseless~\cite{control-affine-to-bilin}. In the same spirit, we have chosen a measurement model of the form~\eqref{eq:control-affine-2}, where it is nonlinear in the state and the landmark position but not both, such that the model can be exactly written as a lifted bilinear model. Many common robot process and measurement models can be written in this form.

The problem of localization is to estimate the states, $\mbs{\xi}_k$, given a series of inputs, $\mbs{\nu}_k$, landmark positions, $\mbs{\psi}_j$, and measurements, $\mbs{\gamma}_{k,j}$. The problem of SLAM is to estimate the states while moving the landmark positions to the list of unknowns: estimate $\mbs{\xi}_k$ and $\mbs{\psi}_j$ given only $\mbs{\nu}_k$ and $\mbs{\gamma}_{k,j}$. Both problems are difficult when the models, $\mbf{f}_i$, $\mbf{g}_{\xi,i}$, $\pi_{\psi,m}$, are nonlinear, and especially so if the models are inaccurate or unknown. Rather than solving these problems directly, we first embed each of the states, inputs, landmarks, and measurements into high-dimensional spaces,
\begin{subequations}\label{eq:embeddings}
\begin{gather}
\mbf{x}_k = \mbf{p}_{\mbs{\xi}} (\mbs{\xi}_k ), \quad
\mbf{u}_k  = \mbf{p}_{\mbs{\nu}}(\mbs{\nu}_k), \\
\mbs{\ell}_j = \mbf{p}_{\mbs{\psi}}(\mbs{\psi}_j), \quad
\mbf{y}_{k,j} = \mbf{p}_{\mbs{\gamma}}(\mbs{\gamma}_{k,j}),
\end{gather}
\end{subequations}
where 
\begin{subequations}\label{eq:embeddings-spaces}
 \begin{gather}
  \mbf{p}_{\mbs{\xi}}: \mathbb{R}^{N_\xi}\rightarrow\ \mathcal{X}, \quad
  \mbf{p}_{\mbs{\nu}}: \mathbb{R}^{N_\nu}\rightarrow \mathcal{U}, \\
  \mbf{p}_{\mbs{\psi}}: \mathbb{R}^{N_\psi}\rightarrow \mathcal{L}, \quad
  \mbf{p}_{\mbs{\gamma}}: \mathbb{R}^{N_\gamma}\rightarrow \mathcal{Y},
 \end{gather}
\end{subequations}
are the embeddings associated with the manifolds $\mathcal{X}$, $\mathcal{U}$, $\mathcal{L}$, and $\mathcal{Y}$, respectively. These embeddings are user-defined and can be customized for specific systems. We discuss some choices for these embeddings later in the Experiments (Section~\ref{sec:experiments}). For the rest of the paper, we assume that the manifolds are within finite-dimensional vector spaces:
\begin{gather}\label{eq:embeds-vs}
 \mathcal{X} \subseteq \mathbb{R}^{N_x}, \quad
 \mathcal{U} \subseteq \mathbb{R}^{N_u}, \quad
 \mathcal{L} \subseteq \mathbb{R}^{N_\ell}, \quad
 \mathcal{Y} \subseteq \mathbb{R}^{N_y},
\end{gather}
and let $N=\max(N_x,N_u,N_{\ell},N_y)$.

For the process model, we follow the same lifting procedure as~\cite{kooplin},~\cite{control-affine-to-bilin}, where the control-affine model becomes a bilinear-Gaussian model in the lifted space. That is,~\eqref{eq:control-affine-1} becomes
\begin{gather}
 \mbf{x}_{k} = \mbf{A}\mbf{x}_{k-1} + \mbf{B}\mbf{u}_k + \mbf{H} \left(\mbf{u}_k \otimes \mbf{x}_{k-1} \right) + \mbf{w}_k,
\end{gather}
where $\otimes$ represents the tensor product, equivalent to the Kronecker product for a finite-dimensional $\mathcal{X}$. The reasoning is that a deterministic control-affine model (i.e., $\mbs{\omega}_k=\mbf{0}$) involves products of nonlinear functions of the robot state and the control input individually, but not nonlinear functions of both quantities. Thus, it can be written exactly as a lifted bilinear model~\cite{control-affine-to-bilin}. When there is system noise,~\cite{kooplin} suggest that a stochastic control-affine model can be modelled similarly with an additive Gaussian noise. In the same spirit, we now lift the measurement model in~\eqref{eq:control-affine-2} so that it also becomes bilinear in the deterministic case,
\begin{gather}
 \mbs{\gamma}_{k,j} = \mbf{g}(\mbs{\xi}_k, \mbs{\psi}_{j}, \mbf{0}) \; \Rightarrow \; \mbf{y}_{k,j} = \mbf{C} (\mbs{\ell}_j \otimes \mbf{x}_k),
\end{gather}
and assume that the noise from a stochastic measurement model can also be modelled as additive-Gaussian noise. This results in the full lifted system,
\begin{subequations}
\label{eq:bilin-inv}
\begin{align}
 \mbf{x}_{k} &= \mbf{A}\mbf{x}_{k-1} + \mbf{B}\mbf{u}_k + \mbf{H} \left(\mbf{u}_k \otimes \mbf{x}_{k-1} \right) + \mbf{w}_k, \label{eq:bilin-inv-1} \\
 \mbf{y}_{k,j} & = \mbf{C} (\mbs{\ell}_{j} \otimes \mbf{x}_k) + \mbf{n}_{k,j}, \label{eq:bilin-inv-2}
\end{align}
\end{subequations}
where $\mbf{w}_k \sim \mathcal{N}(\mbf{0},\mbf{Q})$ and $\mbf{n}_{k,j} \sim \mathcal{N}(\mbf{0}, \mbf{R})$ are the process and measurement noises, respectively, and 
\begin{subequations}
 \begin{gather}
  \mbf{w}_k \in \mathcal{X}, \quad
  \mbf{n}_{k,j} \in \mathcal{Y}, \\
  \mbf{A} \col \mathcal{X} \to \mathcal{X}, \quad
  \mbf{B} \col \mathcal{U} \to \mathcal{X}, \\
  \mbf{H} \col \mathcal{U} \otimes \mathcal{X} \to \mathcal{X}, \quad
  \mbf{C} \col \mathcal{L} \otimes \mathcal{X} \to \mathcal{Y}, \\
  \mbf{Q} \in \mathcal{X} \times \mathcal{X}, \quad
  \mbf{R} \in \mathcal{Y} \times \mathcal{Y},
 \end{gather}
\end{subequations}
where $\mbf{Q}$ and $\mbf{R}$ are positive definite. Note that for a closer parallel to the process model, the measurement model would contain three terms, $\mbf{y}_{k,j} = \mbf{C}_1\mbf{x}_k + \mbf{C}_2\mbs{\ell}_{j} + \mbf{C}_3(\mbs{\ell}_{j} \otimes \mbf{x}_k) + \mbf{n}_{k,j}$, and the subsequent derivations would still follow through. However, the extra terms had little effects during the experimental evaluation, and therefore the one-term version is presented for simplicity.

As we will see in Section~\ref{sec:batch-est}, it is significantly easier to work with the lifted bilinear system in~\eqref{eq:bilin-inv} than with the original nonlinear system in~\eqref{eq:control-affine}. However, we first need to learn the lifted models from data.

%% file: sections/4_SystemIdentification.tex
\section{System Identification}
\label{sec:system-id}

\subsection{Lifted Matrix Form of Dataset}
Our objective is to learn the lifted system matrices $\mbf{A},\mbf{B},\mbf{H},\mbf{C},\mbf{Q},\mbf{R}$ in~\eqref{eq:bilin-inv} from data. We assume a dataset of the control-affine system in~\eqref{eq:control-affine}, including the ground-truth state transitions with their associated control inputs for $P$ states: $\{ \invbreve{\mbs{\xi}}_i, \mbs{\xi}_i, \mbs{\nu}_i \}_{i=1}^P$. Here, $\invbreve{\mbs{\xi}}_i$ transitions to $\mbs{\xi}_i$ under input $\mbs{\nu}_i$. If the dataset consists of a single trajectory of $P+1$ states and $i$ represents the timestep, then we would set $\invbreve{\mbs{\xi}}_i = \mbs{\xi}_{i-1}$. We also assume a dataset of the associated landmark measurements: $\{\{ \mbs{\gamma}_{i,j}, \mbs{\psi}_{i,j} \}_{j=1}^{\beta_i}\}_{i=1}^P$. Here, $\beta_i$ is the number of landmarks seen at timestep $i$, $\mbs{\psi}_{i,j}$ is the position of the $j$th landmark at timestep $i$, and $\mbs{\gamma}_{i,j}$ is the measurement received from landmark $\mbs{\psi}_{i,j}$ at timestep $i$. This format allows for data from one or multiple training trajectories to be used at once, and also allows us to combine datasets with different landmark positions. 
We write the data neatly in block-matrix form:
\begin{subequations}\label{eq:block-matrix-train}
\begin{alignat}{2}
 \mbs{\Xi} &= \begin{bmatrix} \mbs{\xi}_1 & \cdots & \mbs{\xi}_P \end{bmatrix}, \quad
 \invbreve{\mbs{\Xi}} = \begin{bmatrix} \invbreve{\mbs{\xi}}_1 & \cdots & \invbreve{\mbs{\xi}}_P \end{bmatrix}, \\
 \mbs{\Upsilon} &= \begin{bmatrix} \mbs{\nu}_1 & \cdots & \mbs{\nu}_P \end{bmatrix}, \\
  \mbs{\Gamma} &= \left[ \begin{array}{@{}ccc|c|ccc@{}} \mbs{\gamma}_{1,1} & \cdots & \mbs{\gamma}_{1,\beta_1} & \cdots & \mbs{\gamma}_{P,1} & \cdots & \mbs{\gamma}_{P,\beta_P} \end{array} \right], \\
  \mbs{\Psi} &= \left[ \begin{array}{@{}ccc|c|ccc@{}} \mbs{\psi}_{1,1} & \cdots & \mbs{\psi}_{1,\beta_1} & \cdots & \mbs{\psi}_{P,1} & \cdots & \mbs{\psi}_{P,\beta_P} \end{array} \right].
\end{alignat}
\end{subequations}
The data lifts to $\{ \invbreve{\mbf{x}}_i, \mbf{x}_i, \mbf{u}_i \}_{i=1}^P$ and $\{\{ \mbf{y}_{i,j}, \mbs{\ell}_{i,j} \}_{j=1}^{\beta_i}\}_{i=1}^P$ in the embedded space such that
\begin{subequations}\label{eq:train-model-eqns}
\begin{alignat}{2}
&\mbf{x}_i&& = \mbf{A}\invbreve{\mbf{x}}_i + \mbf{B}\mbf{u}_i + \mbf{H} \left(\mbf{u}_i \otimes \invbreve{\mbf{x}}_i \right) + \mbf{w}_i,\\
&\mbf{y}_{i,j} && = \mbf{C} (\mbs{\ell}_{j} \otimes \mbf{x}_i) + \mbf{n}_{i,j},
\end{alignat}
\end{subequations}
for some unknown noise, $\mbf{w}_i\sim\mathcal{N}(\mbf{0},\mbf{Q})$, $\mbf{n}_{i,j} \sim \mathcal{N}(\mbf{0}, \mbf{R})$. We rewrite the lifted versions of the data and the noises in block-matrix form:
\begin{subequations}\label{eq:block-matrix-train-lifted}
{
\begin{align}
 \mbf{X} &= \begin{bmatrix} \mbf{x}_1 & \cdots & \mbf{x}_P \end{bmatrix}, \quad
 \invbreve{\mbf{X}} = \begin{bmatrix} \invbreve{\mbf{x}}_1 & \cdots & \invbreve{\mbf{x}}_P \end{bmatrix}, \\
 \mbf{U} &= \begin{bmatrix} \mbf{u}_1 & \cdots & \mbf{u}_P \end{bmatrix}, \quad
 \mbf{W} = \begin{bmatrix}  \mbf{w}_1 & \cdots & \mbf{w}_P \end{bmatrix}, \\
  \mbf{Y} &= \left[ \begin{array}{@{}ccc|c|ccc@{}} \mbf{y}_{1,1} & \cdots & \mbf{y}_{1,\beta_1} & \cdots & \mbf{y}_{P,1} & \cdots & \mbf{y}_{P,\beta_P} \end{array} \right], \\
 \mbf{L} &= \left[ \begin{array}{@{}ccc|c|ccc@{}} \mbs{\ell}_{1,1} & \cdots & \mbs{\ell}_{1,\beta_1} & \cdots & \mbs{\ell}_{P,1} & \cdots & \mbs{\ell}_{P,\beta_P} \end{array} \right], \\
  \mbf{N} &= \left[ \begin{array}{@{}ccc|c|ccc@{}} \mbf{n}_{1,1} & \cdots & \mbf{n}_{1,\beta_1} & \cdots & \mbf{n}_{P,1} & \cdots & \mbf{n}_{P,\beta_P} \end{array} \right], \\ 
  \candra{\mbf{X}} &= \left[ \begin{array}{@{}c|c|c@{}} \smash[b]{\underbrace{\begin{matrix} \mbf{x}_1 & \cdots & \mbf{x}_1 \end{matrix}}_{\beta_1}} & \cdots & \smash[b]{\underbrace{\begin{matrix} \mbf{x}_P & \cdots & \mbf{x}_P \end{matrix}}_{\beta_P}} \end{array} \right],
\end{align}
}
\end{subequations}
\\
where we defined $\candra{\mbf{X}}$ as the states duplicated based on the number of landmarks seen at each timestep. We also define $S = \sum_{i=1}^P \beta_i$ as the total number of measurements. Then, $\mbf{X}$, $\invbreve{\mbf{X}}$, $\mbf{U}$, $\mbf{W}$ are all $P$ columns wide, while $\mbf{Y}$, $\mbf{L}$, $\mbf{N}$, $\candra{\mbf{X}}$ are all $S$ columns wide. With these definitions, the lifted matrix form of the system for~\eqref{eq:train-model-eqns} is
\begin{subequations}
\begin{align}
\mbf{X} &= \mbf{A}\invbreve{\mbf{X}} + \mbf{B}\mbf{U} + \mbf{H}\left(\mbf{U} \odot \invbreve{\mbf{X}} \right) + \mbf{W}, \\
\mbf{Y} &= \mbf{C}(\mbf{L} \odot \candra{\mbf{X}}) + \mbf{N},
\end{align}
\end{subequations}
where $\odot$ denotes the Khatri-Rao (column-wise) tensor product.

\subsection{Loss Function}
\label{sec:loss-fn}
The model-learning problem is posed as
\begin{gather}
 \left\{\mbf{A}^\star,\mbf{B}^\star,\mbf{H}^\star,\mbf{C}^\star,\mbf{Q}^\star,\mbf{R}^\star\right\} = \argmin_{\{\mbf{A},\mbf{B},\mbf{H},\mbf{C},\mbf{Q},\mbf{R}\}} V(\mbf{A},\mbf{B},\mbf{H},\mbf{C},\mbf{Q},\mbf{R}),
\end{gather}
where the loss function, $V=V_1+V_2$, is the sum of 
\begin{subequations}\label{eq:loss-fn}
\begin{align}
V_1 &= \frac{1}{2} \left\| \mbf{X} - \mbf{A} \invbreve{\mbf{X}} - \mbf{B} \mbf{U} - \mbf{H}( \mbf{U} \odot \invbreve{\mbf{X}} ) \right\|^2_{\mbf{Q}^{-1}} \nonumber \\
&+ \frac{1}{2} \left\| \mbf{Y} - \mbf{C}(\mbf{L} \odot \candra{\mbf{X}} ) \right\|^2_{\mbf{R}^{-1}} - \frac{1}{2} {P} \ln \left| \mbf{Q}^{-1} \right| - \frac{1}{2} {S} \ln \left| \mbf{R}^{-1} \right|, \label{eq:loss-fn-1} \\
V_2 &= \frac{1}{2} {P} \tau_A \left\| \mbf{A} \right\|^2_{\mbf{Q}^{-1}}  + \frac{1}{2} {P} \tau_B \left\| \mbf{B} \right\|^2_{\mbf{Q}^{-1}}  +  \frac{1}{2} {P} \tau_H \left\| \mbf{H} \right\|^2_{\mbf{Q}^{-1}} \nonumber \\
&+  \frac{1}{2} {S} \tau_C \left\| \mbf{C} \right\|^2_{\mbf{R}^{-1}} + \frac{1}{2} {P} \tau_Q \, \mbox{tr}(\mbf{Q}^{-1}) +  \frac{1}{2} {S} \tau_R \, \mbox{tr}(\mbf{R}^{-1}), \label{eq:loss-fn-2}
\end{align}
\end{subequations}
where $|\mbf{X}|$ represents the determinant of $\mbf{X}$, and the norm is a weighted Frobenius matrix norm: $\left\| \mbf{X} \right\|_{\mbf{W}} = \sqrt{\mbox{tr}\left( \mbf{X}^T \mbf{W} \mbf{X} \right)}$. Similar to~\cite{kooplin}, $V_1$ represents the negative log-likelihood of the Bayesian posterior, and  $V_2$ contains prior terms to penalize the description lengths of $\mbf{A}$, $\mbf{B}$, $\mbf{H}$, and $\mbf{C}$ and inverse-Wishart (IW) priors for the covariances, $\mbf{Q}$ and $\mbf{R}$. The regularizing hyperparameters, $\tau_A,\tau_B,\tau_H,\tau_C,\tau_Q,\tau_R$, can be tuned to maximize performance if desired.

We find the critical points by setting the derivatives of $V$ with respect to the model parameters $(\frac{\partial V}{\partial \mbf{A}}$, $\frac{\partial V}{\partial \mbf{B}}$, $\frac{\partial V}{\partial \mbf{H}}$, $\frac{\partial V}{\partial \mbf{C}}$, $\frac{\partial V}{\partial \mbf{Q}^{-1}}$, and $\frac{\partial V}{\partial \mbf{R}^{-1}})$ to zero. We define
\begin{subequations}
\begin{alignat}{2}
 &\mbf{V} = \mbf{U} \odot \invbreve{\mbf{X}}, \quad 
 &&\mbf{J} = \mbf{X} - \mbf{A} \invbreve{\mbf{X}} - \mbf{B} \mbf{U} - \mbf{H}\mbf{V}, \\
 &\mbf{Z} = \mbf{L} \odot \candra{\mbf{X}}, \quad
 &&\mbf{T} = \mbf{Y}-\mbf{C}\mbf{Z}.
\end{alignat}
\end{subequations}
This yields the following expressions that can be solved for $\mbf{A}$, $\mbf{B}$, $\mbf{C}$, and $\mbf{H}$:
\begin{subequations}\label{eq:abhcqr}
\begin{gather}
\begin{bmatrix}
\invbreve{\mbf{X}}\invbreve{\mbf{X}}^T + P \tau_A \mbf{1} &
\invbreve{\mbf{X}}\mbf{U}^T &
\invbreve{\mbf{X}}\mbf{V}^T \\
\mbf{U}\invbreve{\mbf{X}}^T &
\mbf{U}\mbf{U}^T + P \tau_B \mbf{1} &
\mbf{U} \mbf{V}^T \\
\mbf{V}\invbreve{\mbf{X}}^T &
\mbf{V} \mbf{U}^T &
\mbf{V}\mbf{V}^T + P \tau_H \mbf{1}
\end{bmatrix}
\begin{bmatrix} \mbf{A}^T \\ \mbf{B}^T \\ \mbf{H}^T \end{bmatrix}
=
\begin{bmatrix} \invbreve{\mbf{X}}\mbf{X}^T \\ \mbf{U}\mbf{X}^T \\ \mbf{V}\mbf{X}^T \end{bmatrix}, \\
\mbf{C} = (\mbf{Y}\mbf{Z}^T)(\mbf{Z}\mbf{Z}^T + S\tau_C \mbf{1})^{-1},
\end{gather}
after which we obtain
\begin{align}
\mbf{Q} & = \frac{1}{P} \mbf{J} \mbf{J}^T + \tau_A \mbf{A} \mbf{A}^T + \tau_B \mbf{B} \mbf{B}^T + \tau_H \mbf{H} \mbf{H}^T + \tau_Q \mbf{1} , \\
\mbf{R} & = \frac{1}{S} \mbf{T}\mbf{T}^T + \tau_{C}\mbf{C}\mbf{C}^T + \tau_R \mbf{1},
\end{align}
\end{subequations}
where $\mbf{1}$ represents the identity operator of the appropriate domains. The time complexity of solving for $\mbf{A}$, $\mbf{B}$, $\mbf{H}$, $\mbf{Q}$ and $\mbf{R}$ is $\mathcal{O}(N^3(P+S))$, linear in the amount of training~data.

\subsection{Data Augmentation}
If training data is scarce, system invariances can be exploited if they are known. The general system in~\eqref{eq:bilin-inv} allows the model behaviour to vary arbitrarily with the states and with the landmark locations, but this freedom is not necessary for many systems. A robot may drive similarly within a room, and a rangefinding sensor may measure distances based only on the relative location of landmarks with respect to the robot. Although the injection of known invariances into~\eqref{eq:bilin-inv} is an interesting avenue for future work, for now we can use data augmentation~\cite{goodfellow} to improve model accuracy. We make copies of the gathered data, perturb them based on any known invariances, and add them into the dataset matrices in~\eqref{eq:block-matrix-train}. See the dataset experiments in Section~\ref{sec:experiments} for an example.

%% file: sections/5_UKL.tex
\section{Unconstrained Koopman Linearization (UKL)}
\label{sec:batch-est}
We now outline the procedure to set up and solve a batch estimation problem with UKL. This approach is in the same spirit as in~\cite{kooplin}, which solves batch state estimation with fixed landmarks through unconstrained optimization. UKL generalizes~\cite{kooplin} in that the formulation includes new landmark positions at test time, allowing for general localization (UKL-Loc) when the test landmarks are known but are different than during training time, and also allowing for SLAM (UKL-SLAM) when the test landmarks are unknown. We then discuss the limitations of using the minimum-cost solution of UKL, and how it leads to the introduction of manifold constraints for CKL, presented in the next section. Although the constraints turn out to be crucial for performance, we start by outlining UKL as it serves as a basis for CKL.

\subsection{UKL Batch Estimation Problem Setup}
Having learned the system matrices from training data in Section~\ref{sec:system-id}, we now move to an environment with a new set of landmarks, $\{\mbs{\psi}_j\}_{j=1}^V$, where we wish to solve for a sequence of states, $\{\mbs{\xi}_k\}_{k=0}^K$, given a series of inputs, $\{\mbs{\nu}_k\}_{k=1}^K$, and measurements, $\{\{\mbs{\gamma}_{k,j}\}_{j=1}^V\}_{k=0}^K$. If the landmarks are all known, the problem is simply localization. If any landmarks are unknown, the problem is SLAM and we also wish to estimate the unknown landmarks. We do this in the lifted space, where the quantities become $\{\mbs{\ell}_j\}_{j=1}^V$, $\{\mbf{x}_k\}_{k=0}^K$, $\{\mbf{u}_k\}_{k=1}^K$, and $\{\{\mbf{y}_{k,j}\}_{j=1}^V\}_{k=0}^K$, respectively. Since the inputs are completely determined at test time, we can simplify the bilinear process model of~\eqref{eq:bilin-inv-1} into a linear model. Observe that $\mbf{u}_k \otimes \mbf{x}_{k-1} = (\mbf{u}_k \otimes \mbf{1}) \mbf{x}_{k-1}$, where here $\mbf{1} \col \mathcal{X} \to \mathcal{X}$. Using the same trick as in~\cite{kooplin}, we use the known input at test time, $\mbf{u}_k$, to define a new time-varying system matrix, $\mbf{A}_{k-1}$, and a new input, $\mbf{v}_k$,~as
\begin{align}\label{eq:convert-to-ltv}
  \mbf{A}_{k-1} &= \mbf{A} + \mbf{H}(\mbf{u}_k \otimes \mbf{1}), \quad \mbf{v}_k = \mbf{B}\mbf{u}_k.
\end{align}
With this, we have a system in which the process model is linear-Gaussian:
\begin{subequations}\label{eq:ltv-system}
\begin{align}
 \mbf{x}_{k} & = \mbf{A}_{k-1} \mbf{x}_{k-1} + \mbf{v}_k + \mbf{w}_k, \quad k=1,\dots,K, \\
 \mbf{y}_{k,j} & = \mbf{C}(\mbs{\ell}_{j} \otimes \mbf{x}_k) + \mbf{n}_{k,j}, \quad k=0,\dots,K, \; j = 1,\dots,V,
\end{align}
\end{subequations}
where $\mbf{w}_k \sim \mathcal{N}(\mbf{0},\mbf{Q})$ and $\mbf{n}_{k,j} \sim \mathcal{N}(\mbf{0},\mbf{R})$. As the input is always given at test time, this form applies to both localization and SLAM.

We will now describe the process for solving UKL-SLAM, then briefly outline UKL-Loc as a special case of UKL-SLAM. A more efficient method for solving UKL-Loc using the RTS smoother~\cite{barfoot-txtbk} is described in Appendix~\ref{sec:unconstr-loc}.

\subsection{Solving UKL-SLAM}
\label{sec:subsec:unconstr-slam}

If some or all landmarks are unknown, we can solve for both the poses and the unknown landmarks by formulating and solving a batch linear SLAM problem. We first describe the case where all landmark positions are unknown and where the robot receives a measurement for all landmarks at each timestep, and later describe how to modify the method to allow for variations in the problem. As is often done, we assume that $\mbf{x}_0$ is known in order to render a unique solution to the SLAM problem. We define
\begin{gather}\label{eq:slam-block-1}
 \mbs{x} = \bbm \mbf{x}_1 \\ \vdots \\ \mbf{x}_K \ebm, \quad
 \mbs{\ell} = \bbm \mbs{\ell}_1 \\ \vdots \\ \mbs{\ell}_{V} \ebm, \quad
 \mbs{v} = \bbm \mbf{v}_1 \\ \vdots \\ \mbf{v}_K \ebm, \quad
 \mbs{y} = \bbm \mbf{y}_1 \\ \vdots \\ \mbf{y}_K \ebm, \quad
 \mbf{y}_k = \bbm \mbf{y}_{k,1} \\ \vdots \\ \mbf{y}_{k,V} \ebm.
\end{gather}
The pose errors and measurement errors are, respectively,
\begin{subequations}\label{eq:pose-errs}
\begin{align}
 &\mbf{e}_{\mbf{v},k}(\mbf{x}_k) = (\mbf{A}_{k-1} \mbf{x}_{k-1} + \mbf{v}_k) - \mbf{x}_k, \\
 &\mbf{e}_{\mbf{y},k,j}(\mbf{x}_k, \mbs{\ell}_j) = \mbf{C}(\mbs{\ell}_j \otimes \mbf{x}_k) - \mbf{y}_{k,j}.
\end{align}
\end{subequations}
We can formulate the cost function as
\begin{align}\label{eq:cost-orig-slam}
 J(\mbs{x}, \mbs{\ell}) = &\frac{1}{2} \sum_{k=1}^K \mbf{e}_{\mbf{v},k}(\mbf{x}_k)^T \mbf{Q}^{-1} \mbf{e}_{\mbf{v},k}(\mbf{x}_k) \\
 + &
 \frac{1}{2} \sum_{k=1}^K \sum_{j=1}^{V} \mbf{e}_{\mbf{y},k,j}(\mbf{x}_k, \mbs{\ell}_j)^T \mbf{R}^{-1} \mbf{e}_{\mbf{y},k,j}(\mbf{x}_k, \mbs{\ell}_j) \nonumber,
\end{align}
or, in block-matrix form,
\begin{gather}\label{eq:cost-orig}
 J(\mbs{q}) = \frac{1}{2} \mbs{e}(\mbs{q})^T \mbs{W}^{-1} \mbs{e}(\mbs{q}),
\end{gather}
where
\begin{subequations}
 \begin{gather}
  \mbs{q} = \bbm \mbs{x} \\ \mbs{\ell} \ebm, \quad
  \mbs{W} = \bbm \mbs{Q} & \mbf{0} \\ \mbf{0} & \mbs{R} \ebm, \\
  \mbs{Q} = \diag(\mbf{Q},\dots,\mbf{Q}), \quad
  \mbs{R} = \diag(\mbf{R},\dots,\mbf{R}), \label{eq:qr-struct} \\
  \mbs{e} = \bbm \mbs{e}_{\mbs{v}} \\ \mbs{e}_{\mbs{y}} \ebm, \quad
  \mbs{e}_{\mbs{v}} = \bbm \mbf{e}_{\mbf{v},1} \\ \vdots \\ \mbf{e}_{\mbf{v},K} \ebm, \quad
  \mbs{e}_{\mbs{y}} = \bbm \mbf{e}_{\mbf{y},1} \\ \vdots \\ \mbf{e}_{\mbf{y},K} \ebm, \quad
  \mbs{e}_{\mbf{y},k} = \bbm \mbf{e}_{\mbf{y},k,1} \\ \vdots \\ \mbf{e}_{\mbf{y},k,V} \ebm. \label{eq:e-struct}
 \end{gather}
\end{subequations}
Note that the sparsity pattern in $\mbs{W}$ implicitly represents a co-visibility graph, since there are only entries corresponding to process priors among adjacent poses, or measurements from poses to visible landmarks.
The UKL-SLAM optimization objective is
\begin{equation}\label{eq:unconstr-prob}
\begin{aligned}
 \min_{\mbs{q}} \; J(\mbs{q}).
\end{aligned}
\end{equation}
Note that although $J(\mbs{q})$ is quadratic with respect to $\mbs{e}$, it is generally non-quadratic with respect to $\mbs{q}$, and thus~\eqref{eq:unconstr-prob} cannot be solved in one shot. However, we can still use classic Gauss-Newton optimization to efficiently solve~\eqref{eq:unconstr-prob} iteratively by exploiting sparsity structures. See Appendix~\ref{sec:batch-slam-setup} and Appendix~\ref{sec:batch-slam} for details on solving~\eqref{eq:unconstr-prob} with classic Gauss-Newton. The optimal update, $\delta \mbs{q}$, can be computed in complexity $\mathcal{O}(N^3(V^3+V^2K))$ when we have more timesteps than landmarks, or $\mathcal{O}(N^3(K^3+K^2V))$ when we have more landmarks than timesteps. We then update the operating point with
\begin{gather}
 \mbs{x}_{\text{op}} \leftarrow \mbs{x}_{\text{op}} + \alpha \delta \mbs{x}, \quad
 \mbs{\ell}_{\text{op}} \leftarrow \mbs{\ell}_{\text{op}} + \alpha \delta \mbs{\ell},
\end{gather}
for a step size $\alpha$. An appropriate step size can be found using a backtracking line search~\cite{nocedal}. Iterating until convergence yields $(\mbs{x}^\star, \mbs{\ell}^\star)$, or $\mbs{q}^\star$ in batch form.

The distribution of the batch system's solution is $\mbs{q} \sim \mathcal{N}(\hat{\mbs{q}}, \hat{\mbs{P}}_{\mbs{q}})$, where $\hat{\mbs{q}} = \bbm \hat{\mbs{x}} \\ \hat{\mbs{\ell}} \ebm$ and $\hat{\mbs{P}}_{\mbs{q}} = \bbm \hat{\mbs{P}}_{\mbs{x}} & \hat{\mbs{P}}_{\mbs{x}\mbs{\ell}} \\ \hat{\mbs{P}}_{\mbs{x}\mbs{\ell}}^T & \hat{\mbs{P}}_{\mbs{\ell}} \ebm$. We have the batch mean from the Gauss-Newton solution: $\hat{\mbs{q}} = {\mbs{q}^\star}$. We can compute the individual distributions for $\mbf{x}_k$ and $\mbs{\ell}_j$ as follows. The means, $\hat{\mbf{x}}_k$ and $\hat{\mbs{\ell}}_j$, are simply the corresponding blocks from $\hat{\mbs{x}}$ and $\hat{\mbs{\ell}}$. As a feature of classic Gauss-Newton SLAM, we can also compute the covariances, $\hat{\mbf{P}}_{\mbf{x},k}$ and $\hat{\mbf{P}}_{\mbs{\ell},j}$, without raising the computation complexity of SLAM. See Appendix~\ref{sec:batch-slam} for details on computing covariances. The final output is $\mbf{x}_k \sim \mathcal{N}(\hat{\mbf{x}}_k, \hat{\mbf{P}}_{\mbf{x},k})$, $\mbs{\ell}_j \sim \mathcal{N}(\hat{\mbs{\ell}}_j, \hat{\mbf{P}}_{\mbs{\ell},j})$.

\subsection{Recovering Estimates from Lifted Space}
\label{sec:recover-estimates}
After solving for the lifted means and covariances of the states and landmarks, we now convert them back into the original space. We define the block structures of the means and covariances of states and landmarks in the original space for future reference:
\begin{subequations}\label{eq:orig-var-sol}
\begin{gather}
 \mbs{\zeta} = \bbm \mbs{\xi} \\ \mbs{\psi} \ebm, \quad
 \mbs{\zeta} \sim \mathcal{N}(\hat{\mbs{\zeta}}, \hat{\mbs{\Sigma}}_{\mbs{\zeta}}), \\
 \mbs{\xi} \sim \mathcal{N} (\hat{\mbs{\xi}}, \hat{\mbs{\Sigma}}_{\mbs{\xi}}), \quad
 \mbs{\psi} \sim \mathcal{N} (\hat{\mbs{\psi}}, \hat{\mbs{\Sigma}}_{\mbs{\psi}}), \\
 \mbs{\xi} = \bbm \mbs{\xi}_1 \\ \vdots \\ \mbs{\xi}_K \ebm, \quad
 \mbs{\psi} = \bbm \mbs{\psi}_1 \\ \vdots \\ \mbs{\psi}_V \ebm, \\
 \mbs{\xi}_k \sim \mathcal{N} (\hat{\mbs{\xi}}_k, \hat{\mbs{\Sigma}}_{\mbs{\xi},k}), \quad
 \mbs{\psi}_j \sim \mathcal{N} (\hat{\mbs{\psi}}_j, \hat{\mbf{\Sigma}}_{\mbs{\psi},j}).
\end{gather}
\end{subequations}
In general, recovering $\mbs{\xi}_k$ from $\mbf{x}_k$ involves a retraction step~\cite{reprojection}. However, we will restrict the lifting functions to include the original variables as the top parts of the lifted states later in Section \ref{sec:constrained-koopman}. The recovery procedure for the states then becomes simply picking off the top part of $\hat{\mbf{x}}_k$ to get $\hat{\mbs{\xi}}_k$, and picking off the top-left block of $\hat{\mbf{P}}_{\mbf{x},k}$ to get $\hat{\mbs{\Sigma}}_{\mbf{x},k}$. The procedure for the landmarks is analogous. See~\cite{kooplin} for a recovery procedure for general lifting functions.

\subsection{Other UKL Estimation Problems}
\label{sec:ukl-slam-vars}
We can leverage the flexibility of classic Gauss-Newton SLAM to handle variations to UKL-SLAM, including localization and mapping as special cases. It is possible that $\mbf{y}_{k,j}$ is missing, meaning the robot does not receive a measurement from $\mbs{\ell}_j$ at timestep $k$. Or, it is possible that a pose, $\mbf{x}_k$, or a landmark position, $\mbs{\ell}_j$, is known. In both cases, we can modify~\eqref{eqn:slam-eqn} to incorporate missing or known variables without affecting complexity. See Appendix~\ref{sec:batch-slam-setup} for more details. If all of the landmarks are known, SLAM reduces to localization (i.e., UKL-Loc). If all of the robot poses are known, SLAM reduces to mapping. In both of these cases, the cost in~\eqref{eq:cost-orig} becomes quadratic in the remaining unknowns, and using the sparse Cholesky solver (see Appendix~\ref{sec:batch-slam}) yields the minimum-cost solution after just one iteration.

\subsection{Limitations of UKL-Loc and UKL-SLAM}
\label{sec:complications}
At first, solving localization and SLAM through unconstrained optimization seems tempting, especially for UKL-Loc whose solution can be found in one shot. However, observe that there are no conditions that enforce the solution to be on the manifold defined by the lifting functions. In the case of UKL-Loc, we automatically enforce that $\mbf{u}_k \in \mathcal{U}$, $\mbf{y}_{k,j} \in \mathcal{Y}$, $\mbs{\ell}_j \in \mathcal{L}$, as these quantities are given and are directly lifted at test time. However, there is no guarantee that the minimum-cost solution of~\eqref{eq:cost-orig} satisfies $\mbf{x}_k^* \in \mathcal{X}$. Similarly, for UKL-SLAM, there is no guarantee that $\mbf{x}_k^* \in \mathcal{X}$ and $\mbs{\ell}_j^* \in \mathcal{L}$. In both cases, the minimum-cost solution could be very far from the lifting-function manifold. This can be problematic because the trained models are likely poor for governing the behaviour of off-manifold states and landmarks. The states and landmarks within the training data are always on the manifold since they are being lifted from data~\eqref{eq:block-matrix-train} to their lifted versions in~\eqref{eq:block-matrix-train-lifted}. If the minimum-cost solution for~\eqref{eq:cost-orig} happens to be far away from the manifold, the estimates may not be reflective of the training data.

For localization, our experiments suggest that the minimum-cost solution of UKL-Loc tends to be close enough to the manifold when there are frequent measurements. In the simulations in Section~\ref{sec:experiments}, we see that the UKL-Loc trajectory outputs are nearly as good as those of the model-based localizers. We hypothesize that the measurements are `pulling' the estimates back onto the manifold, although the exact pulling mechanism is still unclear. When the measurements are sporadic, however, there are gaps in the measurement stream during which the system can drift far away from the manifold from purely dead reckoning. Thus, although unconstrained Koopman localization can work in scenarios where the measurements are regular, such as in~\cite{kooplin}, it cannot work with sporadic measurements.

\begin{figure}[!t]
 \begin{widepage}
 \centering
\includegraphics[width=0.99\columnwidth,trim={0 0 0 0},clip]{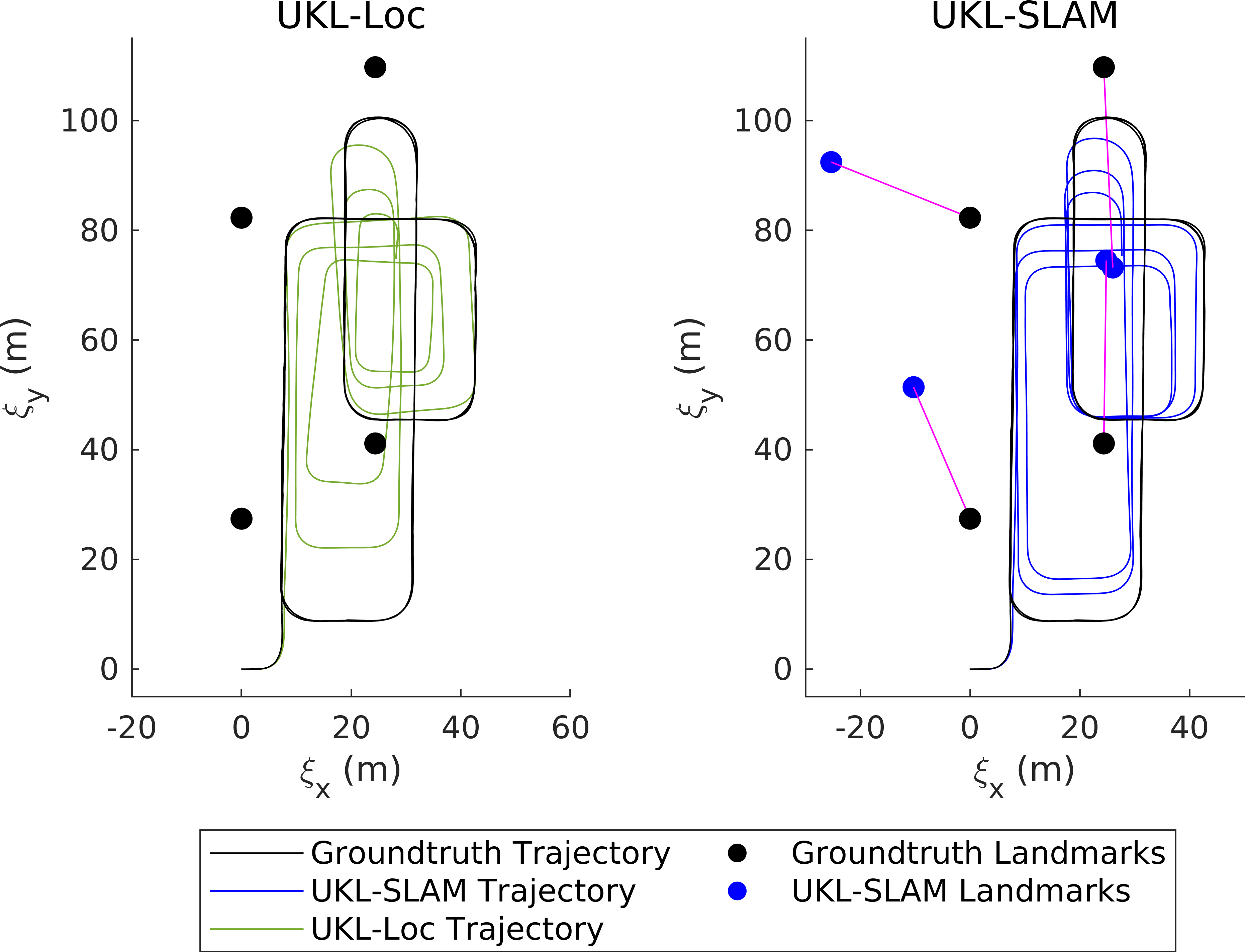}
\end{widepage}
\caption{Visualization of the trajectory output of UKL-Loc (left), and the trajectory and landmark output of UKL-SLAM (right), on Experiment 2 described in Section~\ref{sec:exp-golf-cart}, showing the estimators' mean states and mean landmark positions compared to the groundtruth. The pink lines show the landmark correspondances between the groundtruth and the output of UKL-SLAM. We observe that the trajectory estimate of UKL-Loc is poor, which we typically see when the measurements are sporadic. For UKL-SLAM, both the trajectory and the landmark estimates are poor, which we typically see regardless of the regularity of measurements.}
\label{fig:gesling-unconstkoop}
\end{figure}

On the other hand, the minimum-cost solutions of UKL-SLAM tend to yield very poor trajectory and landmark estimates in general as a result of deviating too much from the manifold. This trend appears to be present irregardless of the type or dimension of the lifting functions used in~\eqref{eq:embeddings}. See \figref{fig:gesling-unconstkoop} for a visualization of the outputs of UKL-Loc and of UKL-SLAM on a dataset with sporadic measurements.

The manifold deviation is related to a known challenge of finding Koopman-invariant subspaces for Koopman process models~\cite{dmd-big-book-compact}. For a noiseless system, a Koopman-invariant subspace is such that any on-manifold state stays on the manifold after passing through the lifted process model. For our case, this means that given lifting functions $\mbf{p}_{\mbs{\xi}}$ and $\mbf{p}_{\mbs{\mu}}$, the original process model, $\mbs{\xi}_k = \mbf{f}(\mbs{\xi}_{k-1}, \mbs{\mu}_k)$, can be written exactly as a lifted model, $\mbf{x}_k = \mbf{A} \mbf{x}_{k-1} + \mbf{B} \mbf{u}_k + \mbf{H}(\mbf{u}_k \otimes \mbf{x}_{k-1})$, such that
\begin{gather}\label{eq:koopman-inv-def}
 \mbf{x}_k \in \mathcal{X} \quad 
 \forall \mbf{x}_{k-1} \in \mathcal{X}, \;
 \forall \mbf{u}_k \in \mathcal{U}.
\end{gather}
Finding Koopman-invariant subspaces for real-world control problems is challenging and an active area of research~\cite{learn-koop-inv-dmd},~\cite{learn-koop-eigs},~\cite{stable-koop-for-prediction}. In the literature, the Koopman operator framework is commonly used for problems with short time horizons such as model-predictive control~\cite{koopman-mpc}, where the manifold deviations are small and can be removed by a retraction step. For example, we could drop the solution back into the original space and then re-lift it at the beginning of a new time window~\cite{reproject-eg},~\cite{reprojection}. However, the solution quality deteriorates with longer time horizons~\cite{koopman-worse-with-time}. In our case, ensuring invariance is further complicated with the addition of the measurement model and the addition of lifted landmarks as unknowns. For batch optimization over very long time horizons such as in SLAM, we find that reprojections cannot sufficiently improve the solution quality after converging to an off-manifold solution.

There is some work on converting nonconvex state-estimation problems to high-dimensional convex problems with only linear constraints~\cite{koopman-const-convex},~\cite{koopman-nonlinear-se}. However, the conversions assume noiseless systems and ideal, possibly infinite-dimensional Koopman operators, both of which cannot be achieved in practice. As a result, these conversions have yet to be applied on real-world datasets.

In the next section, we solve the same estimation problem for the system~\eqref{eq:bilin-inv}, with the addition of constraints to enforce the solution to be on the manifold of the lifting functions.

%% file: sections/6_CKL.tex
\section{Constrained Koopman Linearization (CKL)}
\label{sec:constrained-koopman}
In this section, we present the setup and the solution process for CKL. We present the setup of the optimization problem, the conversion to an SQP, and finally the algorithm for solving the SQP in linear time. We first focus on CKL-SLAM, where we will reuse much of the development for UKL-SLAM. We then show the necessary modifications for other CKL estimation problems including dead reckoning, mapping, and localization (CKL-Loc), followed by a note on reasonable SQP initializations for these problems.

\subsection{CKL-SLAM Problem Setup}
For CKL-SLAM, the optimization objective is identical to~\eqref{eq:unconstr-prob}, with the added constraint that the solution must lie on the manifold of the lifting functions. The objective is
\begin{equation}\label{eq:constr-prob-orig}
\begin{aligned}
 \min_{\mbs{q}} \quad & J(\mbs{q}) \\
 \textrm{s.t.} \quad & \mbf{x}_k \in \mathcal{X}, \quad k = 1,\dots,K, \\
 & \mbs{\ell}_j \in \mathcal{L}, \quad j = 1,\dots,V,
\end{aligned}
\end{equation}
where $J(\mbs{q})$ is defined in~\eqref{eq:cost-orig}. In order to simplify the form of the constraints, we enforce that the estimated quantities' lifting functions, $\mbf{p}_{\mbs{\xi}}(\cdot)$ and $\mbf{p}_{\mbs{\ell}}(\cdot)$, consist of the original quantities stacked on top of the actual lifted features:
\begin{subequations}\label{eq:new-embeddings}
\begin{gather}
\mbf{x}_k = \mbf{p}_{\mbs{\xi}} (\mbs{\xi}_k ) = \bbm \mbs{\xi}_k \\ \tilde{\mbf{p}}_{\mbs{\xi}}(\mbs{\xi}_k) \ebm = \bbm \mbs{\xi}_k \\ \tilde{\mbf{x}}_k \ebm, \label{eq:new-embeddings-x} \\
\mbs{\ell}_j = \mbf{p}_{\mbs{\psi}} (\mbs{\psi}_j) = \bbm \mbs{\psi}_j \\ \tilde{\mbf{p}}_{\mbs{\psi}} (\mbs{\psi}_j) \ebm = \bbm \mbs{\psi}_j \\ \tilde{\mbs{\ell}}_j \ebm, \label{eq:new-embeddings-l}
\end{gather}
\end{subequations}
where
\begin{gather}
 \tilde{\mbf{p}}_{\mbs{\xi}} : \mathbb{R}^{N_\xi} \rightarrow \mathbb{R}^{N_x - N_\xi}, \quad
 \tilde{\mbf{p}}_{\mbs{\psi}} : \mathbb{R}^{N_\psi} \rightarrow \mathbb{R}^{N_\ell - N_\psi}.
\end{gather}
We can write the manifold constraints as equality constraints:
\begin{subequations}\label{eq:eq-constr}
\begin{gather}
 \mbf{x}_k \in \mathcal{X} \; \Rightarrow \; \mbf{h}_{\mbf{x}}(\mbf{x}_k) = \tilde{\mbf{x}}_k - \tilde{\mbf{p}}_{\mbs{\xi}}(\mbs{\xi}_k) = \mbf{0}, \label{eq:eq-constr-x} \\
 \mbs{\ell}_j \in \mathcal{L} \; \Rightarrow \; \mbf{h}_{\mbs{\ell}}(\mbs{\ell}_j) = \tilde{\mbs{\ell}}_j - \tilde{\mbf{p}}_{\mbs{\psi}}(\mbs{\psi}_j) = \mbf{0}. \label{eq:eq-constr-l}
\end{gather}
\end{subequations}
The optimization objective in~\eqref{eq:constr-prob-orig} becomes
\begin{equation}\label{eq:constr-prob}
\begin{aligned}
 \min_{\mbs{q}} \quad & J(\mbs{q}) \\
 \textrm{s.t.} \quad & \mbs{h}(\mbs{q}) = \mbf{0},
\end{aligned}
\end{equation}
where
\begin{gather}\label{eq:h-struct}
 \mbs{h}(\mbs{q}) = \bbm \mbs{h}_{\mbs{x}}(\mbs{x}) \\ \mbs{h}_{\mbs{\ell}}(\mbs{\ell}) \ebm, \quad
 \mbs{h}_{\mbs{x}}(\mbs{x}) = \bbm \mbf{h}_{\mbf{x}}(\mbf{x}_1) \\ \vdots \\ \mbf{h}_{\mbf{x}}(\mbf{x}_K) \ebm, \quad
 \mbs{h}_{\mbs{\ell}}(\mbs{\ell}) = \bbm \mbf{h}_{\mbs{\ell}}(\mbs{\ell}_1) \\ \vdots \\ \mbf{h}_{\mbs{\ell}}(\mbs{\ell}_V) \ebm.
\end{gather}
This objective is identical to the UKL-SLAM objective in~\eqref{eq:unconstr-prob}, with the added manifold constraints. Note that $\mbs{h}(\mbs{q})$ is generally nonlinear, but it has an exploitable structure: each of the blocks in $\mbs{h}_{\mbs{x}}(\mbs{x})$ can be nonlinear but affects only one robot state, and each of the blocks in $\mbs{h}_{\mbs{\ell}}(\mbs{\ell})$ affects only one landmark. This blockwise structure will be especially important for solving the optimization problem efficiently.

\subsection{Solving CKL-SLAM with a Sequential Quadratic Program}
\label{sec:solve-ckl-slam}
We formulate an SQP~\cite{nocedal} to solve~\eqref{eq:constr-prob}. The Lagrangian is
\begin{gather}
 L(\mbs{q}, \mbs{\lambda}) = J(\mbs{q}) - \mbs{\lambda}^T \mbs{h}(\mbs{q}),
\end{gather}
where $\mbs{\lambda}$ is the Lagrange multiplier. Given the Lagrangian, we can solve for the SQP's optimal update, $(\delta\mbs{q},\delta\mbs{\lambda})$. The solution process involves similar matrix structures as the unconstrained problem in~\eqref{eq:unconstr-prob}. See Appendix~\ref{sec:sqp-formulation} for more details on the SQP formulation for CKL-SLAM. We then update the operating point of the primal variable and the multiplier as
\begin{gather}
 \mbs{q}_{\text{op}} \leftarrow \mbs{q}_{\text{op}} + \alpha \delta \mbs{q}, \quad
 \mbs{\lambda}_{\text{op}} \leftarrow \mbs{\lambda}_{\text{op}} + \alpha \delta\mbs{\lambda},
\end{gather}
with an appropriate step size $\alpha$. With~\eqref{eq:constr-prob} containing the same $J(\mbs{q})$ as~\eqref{eq:unconstr-prob} and a blockwise structure on $\mbs{h}(\mbs{q})$, we can extend the sparsity exploitation process of UKL-SLAM to CKL-SLAM. See Appendix~\ref{sec:solve-sqp-lin-time} for details. With this, the optimal updates can be solved in the same time complexity as the unconstrained problem: $\mathcal{O}(N^3(V^3 + V^2 K))$ when we have more timesteps than landmarks, or $\mathcal{O}(N^3(K^3 + K^2 V))$ when we have more landmarks than timesteps.

To extract the covariances of the estimates, $\hat{\mbs{\Sigma}}_{\mbs{\xi},k}$ and $\hat{\mbs{\Sigma}}_{\mbs{\psi},j}$, we need to take into account the effect of the added constraints. We show in Appendix~\ref{sec:extract-cov} that at convergence, the batch covariance of the original variable satisfies
\begin{gather}
 \hat{\mbs{\Sigma}}_{\zeta}^{-1} = \mbs{S}_{||}^T \mbs{F}  \mbs{S}_{||},
\end{gather}
where $\mbs{F}$ is the Gauss-Newton approximation of the Hessian of the Lagrangian~\cite[\S2]{constrained-cov},~\cite[\S3.2]{opt-survey}, and $\mbs{S}_{||} = \mathrm{null}(\mbs{S})$ is a matrix constructed by a basis of the nullspace of $\mbs{S}$. We can then use a similar procedure as for UKL-SLAM to extract the required covariances from $\hat{\mbs{\Sigma}}_{\zeta}^{-1}$. This can be done without raising the computational complexity of SLAM.

\subsection{Other CKL Estimation Problems and SQP Initializations}
\label{sec:sqp-init-short}
CKL-SLAM can be easily modified for other estimation problems including dead reckoning, localization (CKL-Loc), and mapping. We would make similar adjustments to the cost as described in Section~\ref{sec:ukl-slam-vars} for UKL-SLAM, and also analogous adjustments to the constraints. For all of these problems, the matrix sparsity structures are preserved, and we can still use the nullspace method to efficiently solve the SQP.
 
In the presence of nonlinear constraints, the problems of dead reckoning, mapping, localization, and SLAM are all nonlinear optimization problems, and good initial points are often required to avoid convergence to local minima. For dead reckoning, mapping, and localization, we acquire an initial point by dropping the constraints (i.e., the UKL solution). For SLAM, we initialize the robot trajectory by dead reckoning, then initialize the landmarks by mapping from the dead-reckoned trajectory.

%% file: sections/7_RCKL.tex
\section{Reduced Constrained Koopman Linearization (RCKL)}
\label{sec:redkoop}
With the manifold constraints introduced, CKL-SLAM yields much better results than UKL-SLAM. However, as we will explain, CKL is sensitive to poorly fit process models of the lifted features. As a result, CKL's outputs are often still less accurate than those of the model-based methods under experimental evaluation. We now introduce RCKL, a framework that further improves upon CKL by removing the portion of the learned process model that tends to be poorly fit. We first present the framework, and then discuss our hypothesis for its improvement over CKL.

\subsection{Reducing the Koopman Process Model}
We analyze the Koopman process model in~\eqref{eq:bilin-inv-1} with the form of the lifting function, $\mbf{p}_{\mbs{\xi}}(\cdot)$, enforced in~\eqref{eq:new-embeddings-x}. We break down $\mbf{A}$, $\mbf{B}$, and $\mbf{H}$ into two components:
\begin{gather}
 \mbf{A} = \bbm \mbf{A}_{\mbs{\xi}} \\ \tilde{\mbf{A}} \ebm, \quad
 \mbf{B} = \bbm \mbf{B}_{\mbs{\xi}} \\ \tilde{\mbf{B}} \ebm, \quad
 \mbf{H} = \bbm \mbf{H}_{\mbs{\xi}} \\ \tilde{\mbf{H}} \ebm,
\end{gather}
where
\begin{subequations}
\begin{alignat}{2}
 &\mbf{A}_{\mbs{\xi}} \in \mathbb{R}^{N_\xi \times N_x}, \quad &&\tilde{\mbf{A}} \in \mathbb{R}^{(N_x-N_\xi) \times N_x}, \\
 &\mbf{B}_{\mbs{\xi}} \in \mathbb{R}^{N_\xi \times N_u}, \quad &&\tilde{\mbf{B}} \in \mathbb{R}^{(N_x-N_\xi) \times N_u}, \\
 &\mbf{H}_{\mbs{\xi}} \in \mathbb{R}^{N_\xi \times N_x N_u}, \quad &&\tilde{\mbf{H}} \in \mathbb{R}^{(N_x-N_\xi) \times N_x N_u}.
\end{alignat}
\end{subequations}
Then, the process model \eqref{eq:bilin-inv-1}, along with the form of $\mbf{p}_{\mbs{\xi}}(\mbs{\xi}_k)$ in \eqref{eq:new-embeddings-x}, yields
 \begin{subequations}
 \begin{numcases}{}
  \mbs{\xi}_k = \mbf{A}_{\mbs{\xi}} \mbf{x}_{k-1} + \mbf{B}_{\mbs{\xi}} \mbf{u}_k + \mbf{H}_{\mbs{\xi}} (\mbf{u}_k \otimes \mbf{x}_{k-1}) + \mbf{w}_{\mbs{\xi},k}, \label{eq:reduced-1} \\
  \tilde{\mbf{x}}_k = \tilde{\mbf{A}} {\mbf{x}}_{k-1} + \tilde{\mbf{B}} \mbf{u}_k + \tilde{\mbf{H}} (\mbf{u}_k \otimes \mbf{x}_{k-1}) + \tilde{\mbf{w}}, \label{eq:reduced-2}
 \end{numcases}
 \end{subequations}
where $\mbf{w}_k = \bbm \mbf{w}_{\mbs{\xi},k} \\ \tilde{\mbf{w}}_k \ebm \sim \mathcal{N}\left(\mbf{0}, \bbm \mbf{Q}_{\mbs{\xi}} & \mbf{Q}_{\mbs{\xi}\mbf{x}} \\ \mbf{Q}_{\mbs{\xi}\mbf{x}}^T & \mbf{Q}_{\mbf{x}} \ebm \right)$.

If there are no constraints in place, both models are necessary to fully determine $\mbf{x}_k$. With the manifold constraints introduced, however, \eqref{eq:reduced-1} alone is sufficient to determine $\mbf{x}_k$, since the constraint $\mbf{h}_{\mbf{x}}(\mbf{x}_k) = \mbf{0}$ in~\eqref{eq:eq-constr-x} enforces that $\tilde{\mbf{x}}_k = \tilde{\mbf{p}}_{\mbs{\xi}}(\mbs{\xi}_k)$. This means that when~\eqref{eq:reduced-2} is a poor model, we can use~\eqref{eq:reduced-1} to only determine the original state variables, and let the lifting-function constraints determine the lifted features.

To clarify, the lifted features are still used in the process model of~\eqref{eq:reduced-1} as part of the previous state, $\mbf{x}_{k-1}$, but just not determined by the model for the current state. Note that reducing the model does not break the theoretical justifications previously mentioned in Section~\ref{sec:prelim}, including the fact that the conversion of a control-affine model to a lifted bilinear model is exact in the noiseless case.

To train the reduced process model, we train for $\mbf{A}_{\mbs{\xi}},\mbf{B}_{\mbs{\xi}},\mbf{H}_{\mbs{\xi}},\mbf{Q}_{\mbs{\xi}}$. The procedure is similar to the one described in Section~\ref{sec:system-id}, except that the transitioned state is not lifted in the process model. That is, the model in~\eqref{eq:train-model-eqns} becomes
\begin{subequations}
\begin{alignat}{2}
&\mbs{\xi}_i&& = \mbf{A}_{\mbs{\xi}}\invbreve{\mbf{x}}_i + \mbf{B}_{\mbs{\xi}}\mbf{u}_i + \mbf{H}_{\mbs{\xi}} \left(\mbf{u}_i \otimes \invbreve{\mbf{x}}_i \right) + \mbf{w}_i, \label{eq:reduced-sys-1} \\
&\mbf{y}_{k,j} && = \mbf{C} (\mbs{\ell}_{j} \otimes \mbf{x}_i) + \mbf{n}_{i,j}, \label{eq:reduced-sys-2}
\end{alignat}
\end{subequations}
where $\mbf{x}_i$ is replaced with $\mbs{\xi}_i$ in the process model. After making similar modifications to the loss function in~\eqref{eq:loss-fn}, the solutions for $\mbf{A}_{\mbs{\xi}},\mbf{B}_{\mbs{\xi}},\mbf{H}_{\mbs{\xi}},\mbf{Q}_{\mbs{\xi}}$ can be found with
\begin{subequations}\label{eq:abhq-small}
\begin{gather}
 \mbf{V} = \mbf{U} \odot \invbreve{\mbf{X}}, \quad 
 \mbf{J} = \mbs{\Xi} - \mbf{A}_{\mbs{\xi}} \invbreve{\mbf{X}} - \mbf{B}_{\mbs{\xi}} \mbf{U} - \mbf{H}_{\mbs{\xi}}\mbf{V}, \\
\begin{bmatrix}
\invbreve{\mbf{X}}\invbreve{\mbf{X}}^T + P \tau_A \mbf{1} &
\invbreve{\mbf{X}}\mbf{U}^T &
\invbreve{\mbf{X}}\mbf{V}^T \\
\mbf{U}\invbreve{\mbf{X}}^T &
\mbf{U}\mbf{U}^T + P \tau_B \mbf{1} &
\mbf{U} \mbf{V}^T \\
\mbf{V}\invbreve{\mbf{X}}^T &
\mbf{V} \mbf{U}^T &
\mbf{V}\mbf{V}^T + P \tau_H \mbf{1}
\end{bmatrix}
\begin{bmatrix} \mbf{A}_{\mbs{\xi}}^T \\ \mbf{B}_{\mbs{\xi}}^T \\ \mbf{H}_{\mbs{\xi}}^T \end{bmatrix}
=
\begin{bmatrix} \invbreve{\mbf{X}}\mbs{\Xi}^T \\ \mbf{U}\mbs{\Xi}^T \\ \mbf{V}\mbs{\Xi}^T \end{bmatrix}, \\
\mbf{Q}_{\mbs{\xi}} = \frac{1}{P} \mbf{J} \mbf{J}^T + \tau_A \mbf{A}_{\mbs{\xi}} \mbf{A}_{\mbs{\xi}}^T + \tau_B \mbf{B}_{\mbs{\xi}} \mbf{B}_{\mbs{\xi}}^T + \tau_H \mbf{H}_{\mbs{\xi}} \mbf{H}_{\mbs{\xi}}^T + \tau_Q \mbf{1}.
\end{gather}
\end{subequations}
At test time, we modify the time-varying quantities defined in~\eqref{eq:convert-to-ltv},
\begin{gather}
  \mbf{A}_{\mbs{\xi},k-1} = \mbf{A}_{\mbs{\xi}} + \mbf{H}_{\mbs{\xi}} (\mbf{u}_k \otimes \mbf{1}), \quad \mbf{v}_k = \mbf{B}_{\mbs{\xi}}\mbf{u}_k,
\end{gather}
and modify error functions in~\eqref{eq:pose-errs} to be
\begin{subequations}
\begin{align}
 &\mbf{e}_{\mbf{v},k}(\mbf{x}_k) = (\mbf{A}_{\mbs{\xi},k-1} \mbf{x}_{k-1} + \mbf{v}_k) - \mbs{\xi}_k, \\
 &\mbf{e}_{\mbf{y},k,j}(\mbf{x}_k, \mbs{\ell}_j) = \mbf{C}(\mbs{\ell}_j \otimes \mbf{x}_k) - \mbf{y}_{k,j},
\end{align}
\end{subequations}
where the pose error is modified to weigh only the $\mbs{\xi}_k$ component of $\mbf{x}_k$, while the measurement error is still using the full set of features in $\mbf{x}_k$. The cost function is modified from~\eqref{eq:cost-orig-slam} to be
\begin{align}\label{eq:cost-reduced}
 J(\mbs{x}, \mbs{\ell}) = &\frac{1}{2} \sum_{k=1}^K \mbf{e}_{\mbf{v},k}(\mbf{x}_k)^T \mbf{Q}_{\mbs{\xi}}^{-1} \mbf{e}_{\mbf{v},k}(\mbf{x}_k) \\
 + & \;
 \frac{1}{2} \sum_{k=1}^K \sum_{j=1}^{V} \mbf{e}_{\mbf{y},k,j}(\mbf{x}_k, \mbs{\ell}_j)^T \mbf{R}^{-1} \mbf{e}_{\mbf{y},k,j}(\mbf{x}_k, \mbs{\ell}_j) \nonumber,
\end{align}
which can be written in the familiar block-matrix form of $J(\mbs{q}) = \frac{1}{2} \mbs{e}(\mbs{q})^T \mbs{W}^{-1} \mbs{e}(\mbs{q})$ in~\eqref{eq:cost-orig}. This means that the RCKL-SLAM objective is the same form as~\eqref{eq:h-struct}, the CKL-SLAM objective. To solve for $\mbs{q}$, we can follow the same procedure as Section~\ref{sec:solve-ckl-slam}. The complexity of the SQP iterations are the same, possibly with a lower coefficient owing to the smaller process model.

\subsection{Theoretical Justification of RCKL}
\label{sec:redkoop-justification}
\begin{figure*}[!t]
\begin{widepage}
\centering
 \subcaptionbox*{}{
 \includegraphics[width=0.31\textwidth]{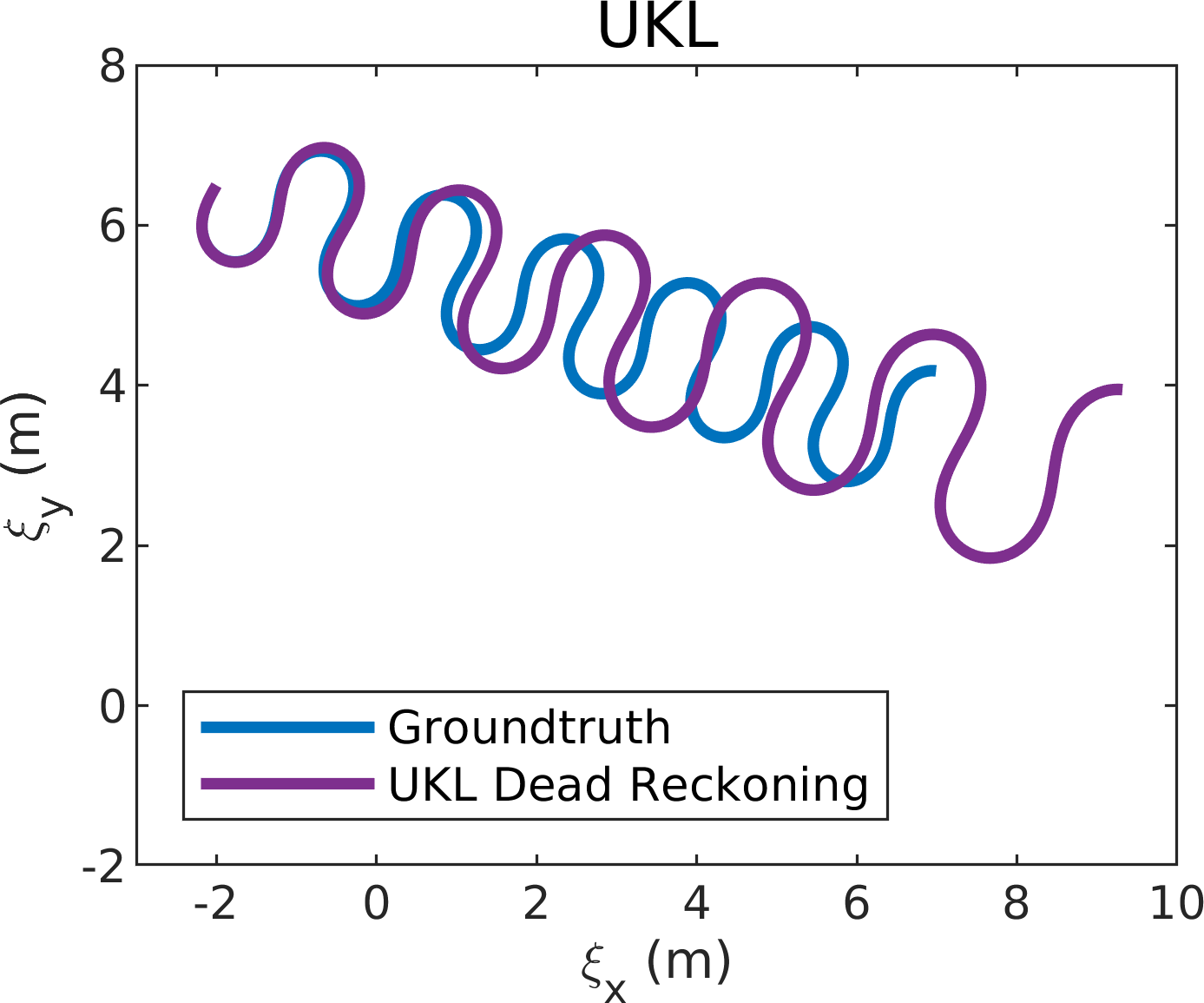}
 }
  \subcaptionbox*{}{
 \includegraphics[width=0.31\textwidth]{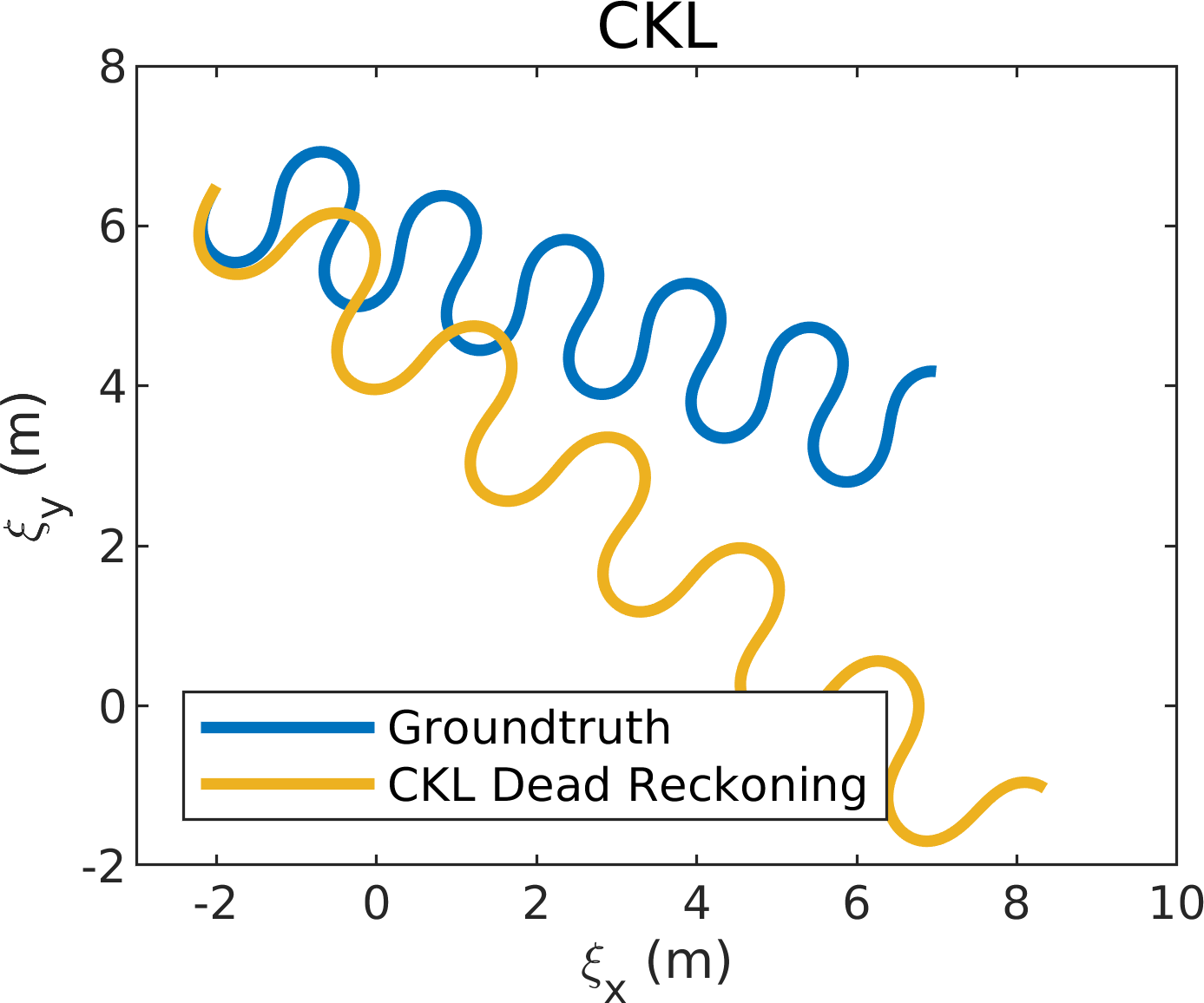}
 }
  \subcaptionbox*{}{
 \includegraphics[width=0.31\textwidth]{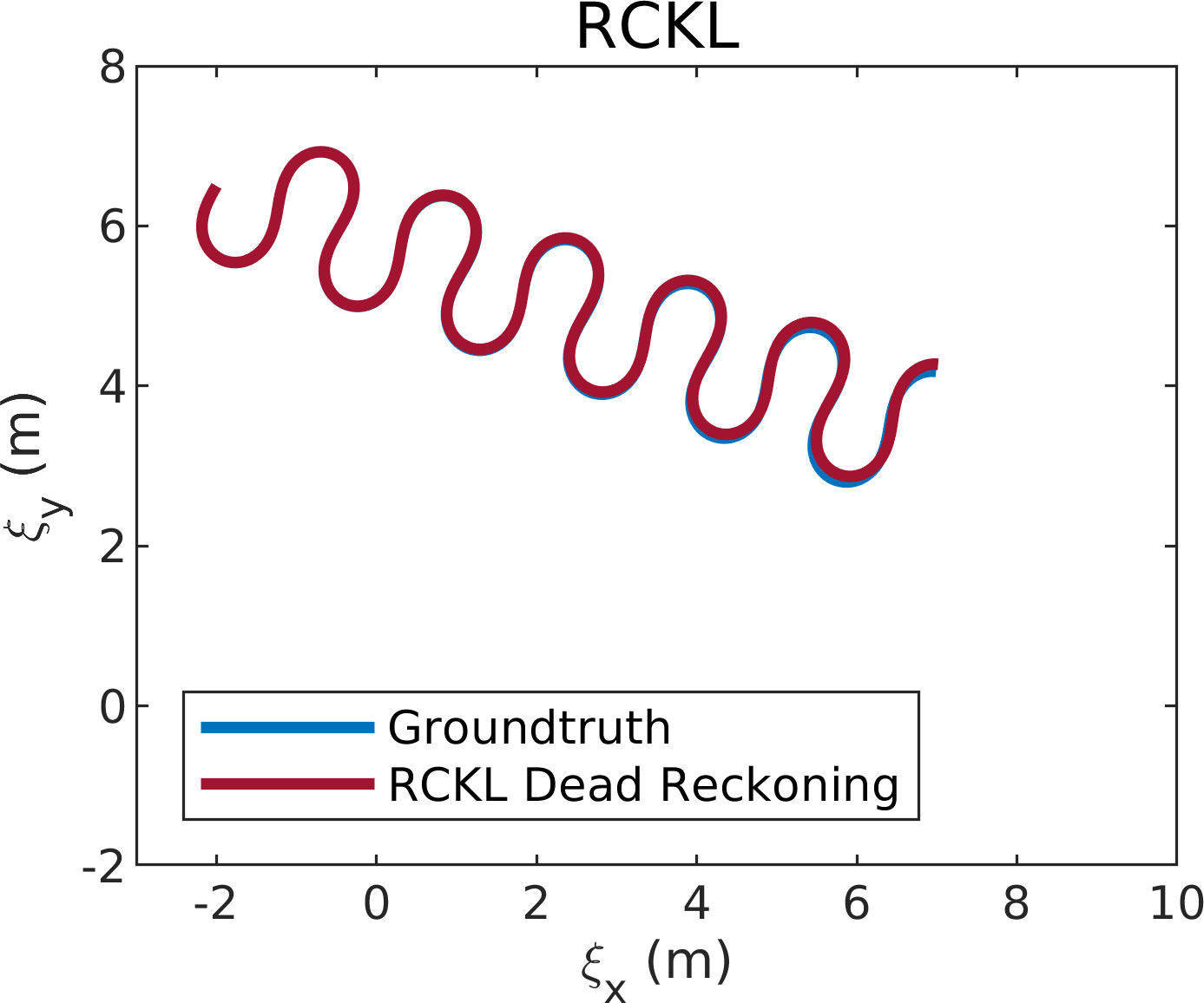}
 }
\end{widepage}
\vspace{-10pt}
\caption{\textit{UKL vs. CKL vs. RCKL in noiseless dead reckoning}. We train the three estimators on noiseless data corresponding to Simulation 1 described in Section~\ref{sec:simu-1}, then compare their dead-reckoning outputs on a noiseless test trajectory. UKL's output is on top of groundtruth at first but eventually drifts off as the states deviate from the feature manifold. CKL's output remains on the feature manifold, but solution is worsened by the poor process models on the features. Meanwhile, RCKL stays on the manifold and its output corresponds to the groundtruth almost exactly.}
\label{fig:dr-comparison}
\vspace{-10pt}
\end{figure*}
For the experimental scenarios presented Section~\ref{sec:experiments}, RCKL's process model always performs better than CKL's model. See \figref{fig:dr-comparison} for a visual example of noiseless dead reckoning with UKL, CKL, and RCKL, where RCKL's output is much closer to the groundtruth trajectory. We now discuss our hypothesis for this improvement in performance.

The main difference between RCKL and CKL is that RCKL avoids fitting a model for the evolution of the lifted features. A poor process model of the features would negatively affect the whole system through the constraints. In our experiments, we see that when a model contains all of the lifting functions required to fit the system, adding extraneous lifting functions to CKL degrades its performance. Even when $\tilde{\mbf{x}}_{k-1}$ is essential for determining $\mbs{\xi}_k$ in~\eqref{eq:reduced-1}, its own evolution to $\tilde{\mbf{x}}_k$ in~\eqref{eq:reduced-2} may be poor, which would affect $\mbs{\xi}_k$ through the constraint $\mbf{h}_{\mbf{x}}(\mbf{x}_k) = \tilde{\mbf{x}}_k - \tilde{\mbf{p}}_{\mbs{\xi}}(\mbs{\xi}_k) = \mbf{0}$.

This concept is related to how Koopman invariance~\eqref{eq:koopman-inv-def} is hard to satisfy especially for the process models of the lifted features. See~\cite[p.~263]{dmd-big-book-compact} for an example of where the Koopman representation of even a simple system becomes intractable when using polynomial features. The higher-order polynomials added require even higher-order polynomials to fit their own process model, ad infinitum. 

After selecting the features from a large pool of possible lifting functions, we find that the linear process models on the features tend to be poorly fit. RCKL, in contrast, fits a linear model on only the original variables and relies on the nonlinear manifold constraints for the evolution of the features. Assuming that we have enough training data to avoid overfitting, adding more lifting functions in RCKL would only add more expressiveness for fitting the model.

Note that RCKL's reduced model follows the same spirit as the original sparse identification of nonlinear dynamics (SINDy) algorithm~\cite{sindy}. For a nonlinear system, $\mbs{\xi}_k = \mbf{f}(\mbs{\xi}_{k-1})$, SINDy optimizes for a sparse $\mbf{A}$ and a low-dimensional $\mbf{p}_{\mbs{\xi}}(\cdot)$ such that the model can be written as $\mbs{\xi}_k = \mbf{A} \mbf{p}_{\mbs{\xi}}(\mbs{\xi}_{k-1})$. That is, SINDy identifies the nonlinear functions in $\mbf{p}_{\mbs{\xi}}(\cdot)$ necessary to reconstruct the original system. This form is similar to our reduced process model in~\eqref{eq:reduced-sys-1} with $\mbf{p}_{\mbs{\xi}}$ as the lifting function. However, SINDy's nonlinear model is not originally compatible with Koopman methods (without constraints). Instead, many Koopman methods use a modified version of SINDy to reconstruct $\mbf{p}_{\mbs{\xi}}(\mbs{\xi}_k) = \mbf{A} \mbf{p}_{\mbs{\xi}}(\mbs{\xi}_{k-1})$~\cite{koop-inv-with-sindy}, which is essentially finding the full process model. With the manifold constraints introduced, the reduced model is once again possible for Koopman systems.

One potential drawback of RCKL is that for some systems, the process model of some features could be simpler to construct than that of the original variables. In this case, a hybrid method could be used. As this is likely very problem-specific, we leave the investigation of this idea for future work. 

%% file: sections/8_ExperimentsResults.tex
\section{Experiments and Results}
\label{sec:experiments}
\begin{figure*}[!t]
\begin{widepage}
\centering
\subcaptionbox{Simulation 1 Results}{
\includegraphics[width=0.95\columnwidth]{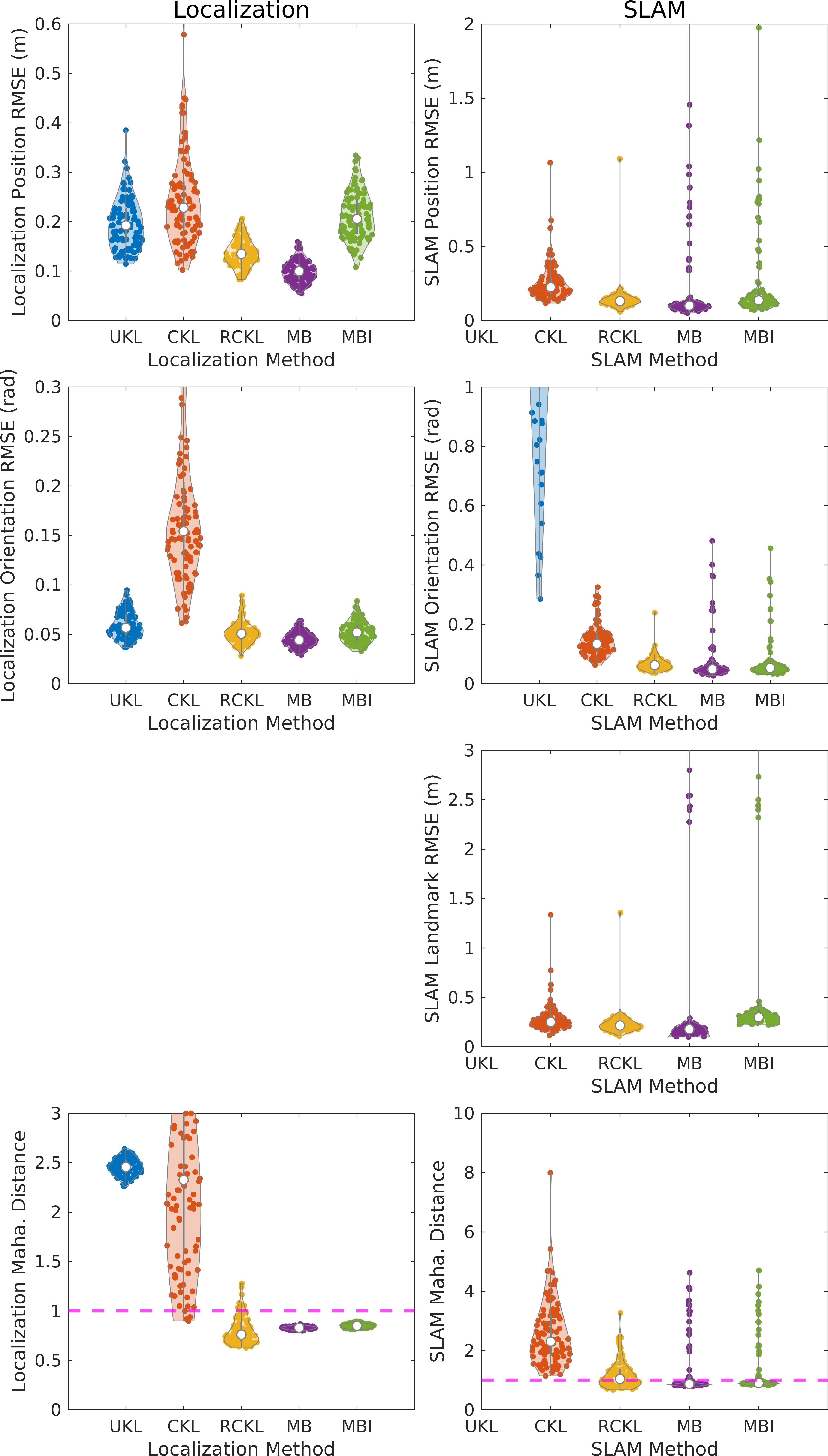}
}
\hspace{20pt}
\subcaptionbox{Simulation 2 Results}{
\includegraphics[width=0.95\columnwidth]{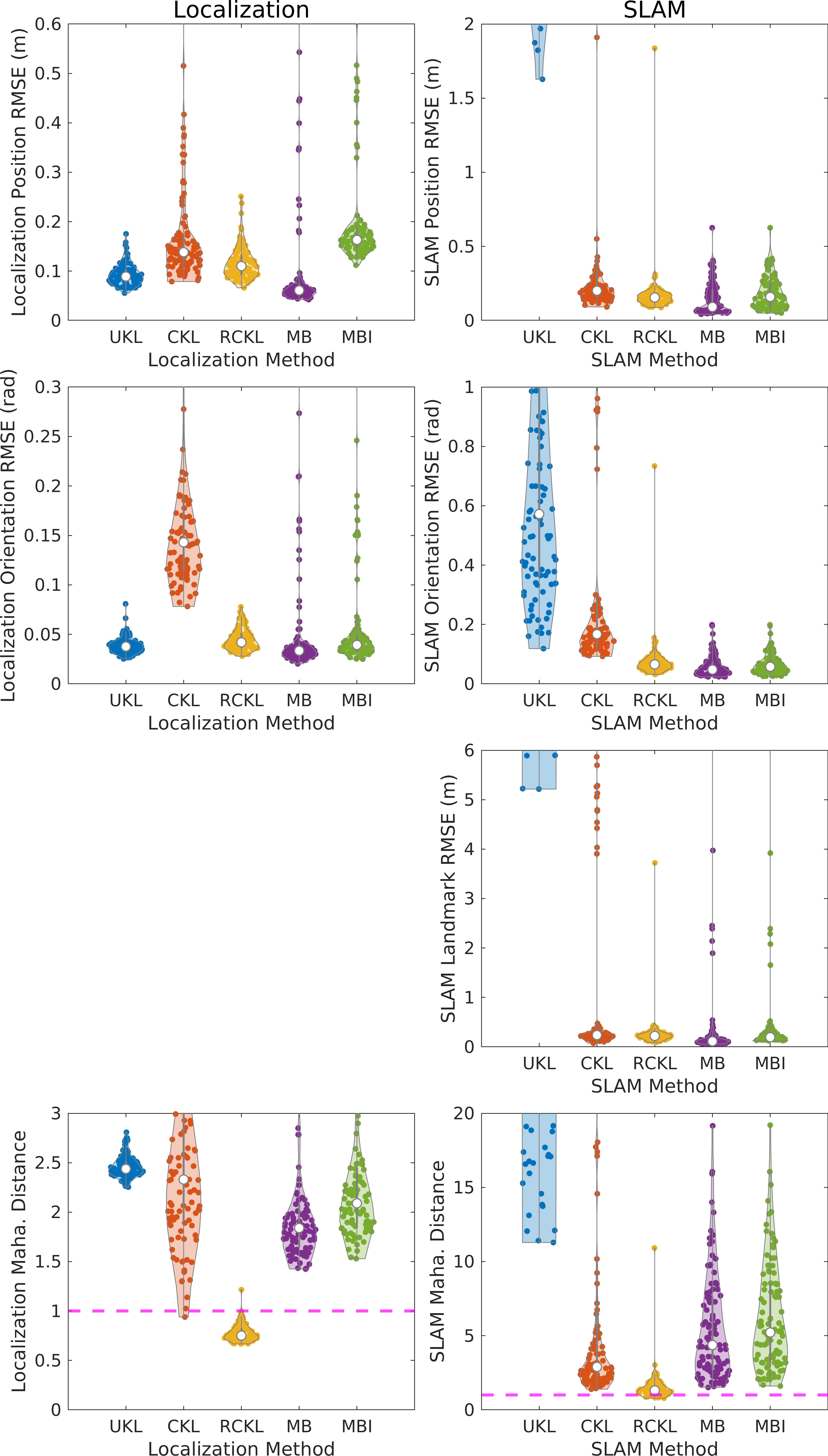}
}
\end{widepage}
\caption{\textit{Localization and SLAM results for Simulation 1 (left) and Simulation 2 (right)}. RMSEs and Mahalanobis distances for localization (left plots) and SLAM (right plots) for UKL, CKL, RCKL, MB (model-based with correct $\mu$), and MBI (model-based with imperfect $\mu$). In both simulations, RCKL has slightly higher RMSEs than MB but encounters local minima less frequently, and it has lower RMSEs than MBI. UKL-SLAM RMSEs are off the charts, and CKL-SLAM RMSEs are more reasonable than UKL-SLAM but still higher than RCKL-SLAM, demonstrating the necessity of the constraints and the reduced process model in RCKL. RCKL is also more consistent than MB and MBI.}
\label{fig:simu-res}
\vspace{-15pt}
\end{figure*}

We evaluate the performance of the Koopman estimators for localization and SLAM in two simulation scenarios and on two real-world datasets. We investigate the performances of UKL, CKL, and RCKL in simulation, and focus on RCKL on the datasets, owing to its superior performance in simulation. We compare our results with those of a classic model-based nonlinear batch estimator optimized using Gauss-Newton~\cite[\S8/9]{barfoot-txtbk}. We term this estimator ``MB'' for model-based. To demonstrate the advantages of our data-driven approach, we also compare with a classic estimator where the model parameters are imperfect, and we term this estimator ``MBI'' for model-based imperfect. We will show that our Koopman estimators can outperform MBI by learning the model parameters. Both the Koopman estimators and the model-based estimators follow the initialization procedure outlined in Section~\ref{sec:sqp-init-short}: initializing with dead reckoning for localization, and initializing with mapping from dead reckoning for SLAM.

To evaluate accuracy, we compute the root-mean-squared-error (RMSE) of an estimator's mean output. To evaluate consistency, we compute the normalized trajectory-level Mahalanobis distance of an estimator's mean and covariance outputs. This Mahalanobis distance measures the consistency of the entire batch solution and is computed with $\sqrt{(\delta \mbs{\zeta})^T \mbs{\Sigma}^{-1} (\delta \mbs{\zeta}) / N}$, where $\delta \mbs{\zeta}$ is the error of the estimator's mean outputs compared to the groundtruth, $\mbs{\Sigma}^{-1}$ is the inverse covariance output, and $N$ is the degrees of freedom of the system. An accurate estimator has an RMSE close to 0, and a consistent estimator has a Mahalanobis distance close to 1. For evaluating SLAM, we first perform an alignment step where the groundtruth is aligned to the output trajectory through a rigid transformation before computing the RMSEs and the Mahalanobis distances. This is because SLAM is used to derive a relative map between the trajectory and the landmarks, and the global SLAM error within the frame of the first pose is usually of lesser interest.

In the experiments below, the main lifting functions used are squared-exponential Random Fourier Features (SERFFs) \cite{rff}, which take in any number of vector-space inputs and generate a user-specified number of sinusoidal functions. SERFFs are common lifting functions for Koopman methods used for their experimentally determined high performance~\cite{kooplin},~\cite{koopman-with-rff}, and for their connection to the squared-exponential kernel~\cite{kooplin}, \cite{rff-koopman}. The formulas for generating the SERFFs can be found in~\cite{rff}. Note that the generation of SERFFs involve picking hyperparameter values that can be tuned based on data if necessary. Below, we use the notation $\mbf{r}_{\mbf{x}}(\mbf{x})$ to denote the SERFFs for input $\mbf{x}$.

\subsection{Simulation Setup}
For both simulation scenarios, the setup consists of a robot driving in a 2D plane, receiving measurements from landmarks scattered around the plane. We evaluate localization and SLAM on 100 test instances, each with a 1000-timestep trajectory and 10 landmarks. The measurements are scaled by a constant factor of $\mu=1.05$, which our Koopman estimators will learn from data. We compare the performance of the Koopman estimators against MB, whose measurement model uses the correct value of $\mu=1.05$, and MBI, whose measurement model uses $\mu=1$. To clarify, MB and MBI are using the same raw data, but they are doing estimation with slightly different prior measurement models, thus affecting their solutions. See~\cite[\S8/9]{barfoot-txtbk} for how the prior measurement models are used for computing costs and Jacobians in classic Gauss-Newton estimation.

\subsubsection{Simulation 1: Unicycle Model, Range Measurements}
\label{sec:simu-1}
The inputs for the unicycle model~\cite{unicycle} are the translational and rotation speeds of the robot, and the measurements are distances. For timestep $k$ and landmark $j$, the robot state, input, landmark position, and measurement are, respectively, 
\begin{equation}\label{eq:simu-1-defs}
 \mbs{\xi}_k = \bbm \xi_{x,k} \\ \xi_{y,k} \\ \xi_{\theta,k} \ebm, \quad
 \mbs{\nu}_k = \bbm u_k \\ \omega_k \ebm, \quad
 \mbs{\psi}_j = \bbm \psi_{x,j} \\ \psi_{y,j} \ebm, \quad
 {\gamma}_{k,j} = \mu r_{k,j},
\end{equation}
where $(\xi_{x,k},\xi_{y,k})$ is the robot's global position, $\xi_{\theta,k}$ is the robot's global orientation, $u_k$ is the robot's linear speed, $\omega_k$ is its angular speed, $(\psi_{x,j}, \psi_{y,j})$ is the landmark's global position, $r_{k,j}$ is the range measurement from the robot to the $j$th landmark, and $\mu$ is the constant scaling factor. Contrary to the assumption made in~\eqref{eq:new-embeddings}, the state is not within a vector space since it includes orientation as a circular quantity. As such, we convert it to an augmented state, $\mbs{\xi}_k^'$, with the substitutions $\xi_{\cos\theta,k} = \cos \xi_{\theta,k}$ and $\xi_{\sin\theta,k} = \sin \xi_{\theta,k}$. For CKL and RCKL, we introduce an additional state constraint, $h_\xi(\mbs{\xi}^\prime_k) = \xi_{\cos \theta,k}^2 + \xi_{\sin \theta,k}^2 - 1 = 0$.
See Appendix~\ref{sec:orient} for more details on handling the orientation.
We used SERFFs for the state, the landmark position, the measurements, and input:
\begin{subequations}\label{eq:simu-1-lifts}
\begin{gather}
 \mbs{\xi}_k^\prime = \bbm \xi_{x,k} \\ \xi_{y,k} \\ \xi_{\cos\theta,k} \\ \xi_{\sin\theta,k} \ebm, \quad
 \mbf{p}_{\mbs{\xi}^'}(\mbs{\xi}^'_k) =
 \bbm \mbs{\xi}_k^' \\ 
 \mbf{r}_{x,y}(\xi_{x,k}, \xi_{y,k}) \\
 \mbf{r}_{\cos\theta,\sin\theta}(\xi_{\cos\theta,k}, \xi_{\sin\theta,k}) \\
 \mbf{r}_{x,\cos\theta}(\xi_{x,k}, \xi_{\cos\theta,k}) \\
 \mbf{r}_{x,\sin\theta}(\xi_{x,k}, \xi_{\sin\theta,k}) \\
 \mbf{r}_{y,\cos\theta}(\xi_{y,k}, \xi_{\cos\theta,k}) \\
 \mbf{r}_{y,\sin\theta}(\xi_{y,k}, \xi_{\sin\theta,k}) \ebm, \\
 \mbf{p}_{\mbs{\psi}}(\mbs{\psi}_j) = \bbm \mbs{\psi}_j \\ \mbf{r}_{\mbs{\psi}}(\mbs{\psi}_{j}) \ebm, \quad
 \mbf{p}_{\gamma}(\gamma_{k,j}) = \bbm \gamma_{k,j} \\ \mbf{r}_{\gamma}(\gamma_{k,j}) \ebm, \quad
 \mbf{p}_{\mbs{\nu}}(\mbs{\nu}_k) = \bbm \mbs{\nu}_k \\ \mbf{r}_{\mbs{\nu}}(\mbs{\nu}_k) \ebm,
\end{gather}
\end{subequations}
where for the state, SERFFs are constructed on all possible pairs of variables in $\mbs{\xi}_k^\prime$ rather than constructing $\mbf{r}_{\mbs{\xi}^\prime}(\xi_{x,k}, \xi_{y,k}, \xi_{\cos\theta,k}, \xi_{\sin\theta,k})$. We find this form to be sufficient for our environment, without the burden of a vastly high-dimensional lifted state and requiring much more training data to avoid overfitting.

\subsubsection{Simulation 2: Bicycle Model, Range-Squared Measurements}
The setup is the same as for Simulation 1, except that the robot has a bicycle process model~\cite{bicycle} and the measurement is the squared range (to show the generalizability of our data-driven methods):
\begin{gather}
 \mbs{\nu}_k = \bbm u_k \\ \phi_k \ebm, \quad
 \gamma_{k,j} = \mu r_{k,j}^2,
\end{gather}
where $u_k$ is still the linear speed but $\phi_k$ is the steering angle of the bicycle's front axle. The lifting functions used are in the same form as in~\eqref{eq:simu-1-lifts} except for the input, for which we account for the circular input of $\phi_k$ by using
\begin{gather}
 \mbf{p}_{\mbs{\nu}}(\mbs{\nu}_k) = \bbm u_k \\ \cos \phi_k \\ \sin \phi_k \\ \mbf{r}_{u,\cos\phi,\sin\phi}(u_k, \cos \phi_k, \sin \phi_k) \ebm.
\end{gather}
\begin{figure}[!t]
\begin{widepage}
\centering
\includegraphics[width=0.9\columnwidth,trim={0 30pt 0 0},clip]{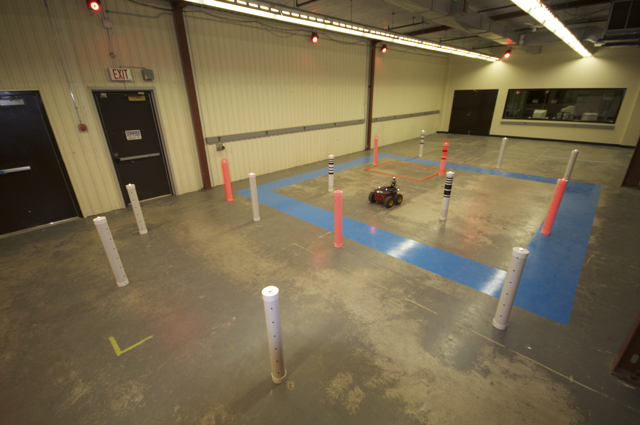}
\end{widepage}
 \caption{Setup for Experiment 1. A wheeled robot drives around 17 cylindrical landmarks (2 are not visible in this photo) in an indoor environment. It logs wheel odometry and measures the range to its surrounding landmarks using a laser rangefinder. The landmarks for testing are highlighted in red.}
 \label{fig:robot-pic}
 \vspace{5pt}
\end{figure}
\begin{figure*}[!h]
 \centering
   \includegraphics[width=0.90\textwidth]{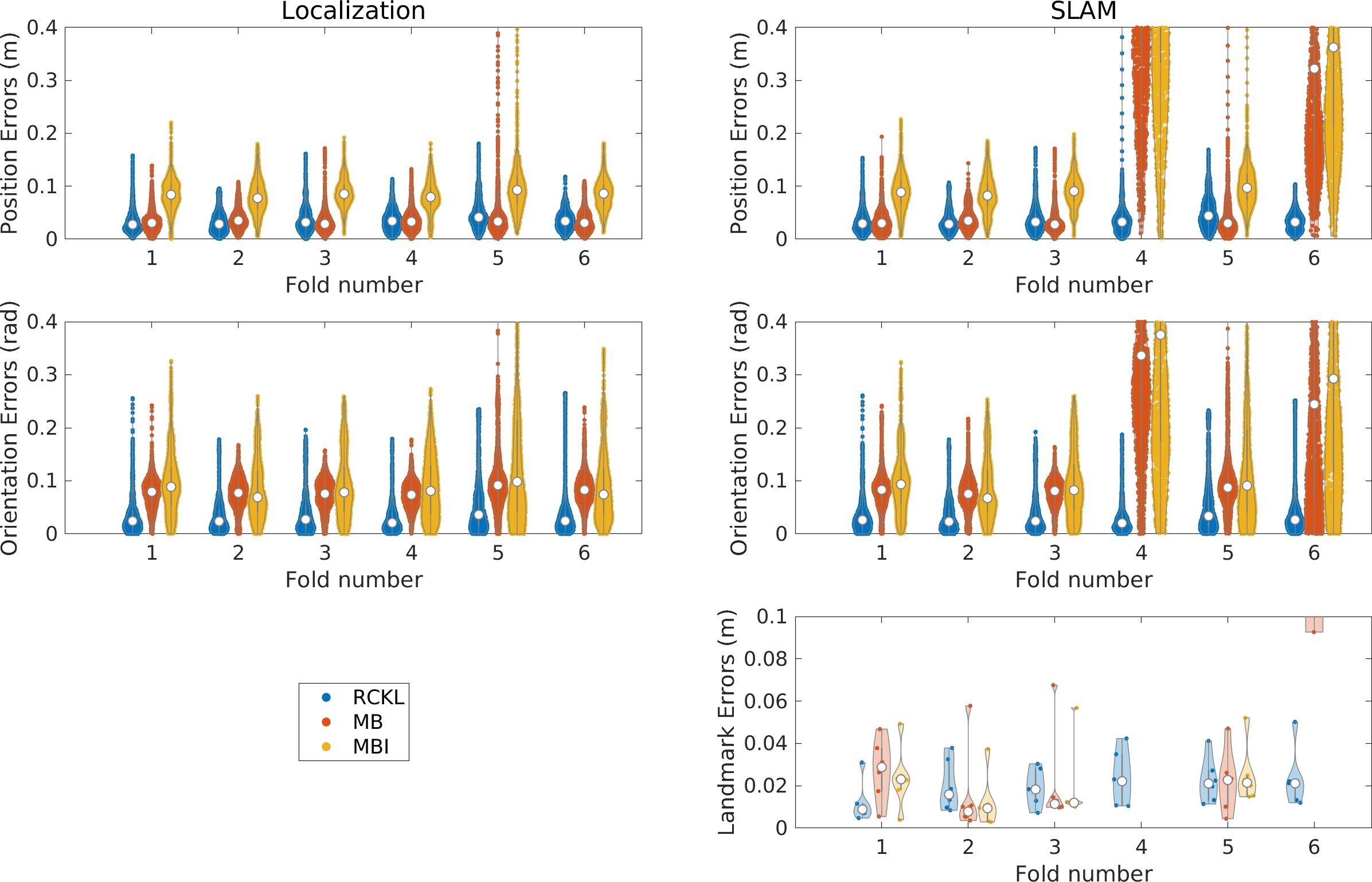}
 \caption{\textit{Localization and SLAM errors for Experiment 1}: position and orientation errors of the mean trajectory and the mean landmark outputs of RCKL, MB, and MBI (whose robot-to-rangefinder distance is offset by 10 cm), on the 6 dataset folds. For localization, RCKL has similar position errors and generally lower orientation errors compared to MB. Against MBI, RCKL has lower position and orientation errors, demonstrating the advantage of the data-driven approach. These trends are also present for SLAM on folds 1, 2, 3, and 5. On folds 4 and 6, the model-based SLAM errors are much higher as a result of convergence to local minima (see \figref{fig:lost-in-woods-local-min} for a visualization).}
 \label{fig:lost-in-woods-errors}
 \vspace{-5pt}
 \end{figure*}

\subsection{Simulation Results Discussion}
The results for Simulation 1 and Simulation 2 are shown in \figref{fig:simu-res}. We see that MB tends to have the lowest errors for localization and SLAM, whereas RCKL has similar or slightly higher errors. This outcome is expected since the models and the noise distributions are perfectly known for MB, whereas the Koopman estimators have learned the model through noisy data. However, RCKL has similar or better accuracy compared to MBI, demonstrating the advantages of the data-driven approach. In addition, while the model-based estimators, MB and MBI, encounter local minima in both localization and SLAM, RCKL appears to encounter local minima much less frequently.\footnote{We have verified the optimality of our solutions by running our optimization problems two times: the first time is after following our standard initialization procedure described in Section~\ref{sec:sqp-init-short}, and the second time is after initializing at groundtruth. Through this procedure, we can verify that a local minimum has occurred when the first solution (standard initialization) has a higher cost than the second solution (groundtruth initialization). Global optimality is harder to verify, but we assume that the groundtruth initialization converges near the global minimum for our considered noise levels. Then, a global mimimum has likely occurred when the first and second solution has the same cost.} This suggests that another potential benefit of reformulating nonlinear estimation problems in a lifted space is improved convergence properties.\footnote{We chose not to compare to globally optimal estimation methods because they tend to be prohibitively expensive for even moderately sized problems~\cite{holmes2023semidefinite}. If desired, one could attempt to add a global optimality certificate~\cite{safe-and-smooth} to either MB or RCKL.}

For the other Koopman estimators, we see that UKL-Loc is viable in our environment where a measurement is received from every landmark at every timestep. However, UKL-SLAM is not at all viable, and only becomes viable when the constraints are added to yield CKL-SLAM. CKL still has higher errors than the model-based estimators, MB and MBI. It only become comparable to MB for RCKL, that is, after the reduced process model is employed.

\subsection{Experiment 1: Indoor Navigation with Laser Rangefinder}
\label{sec:exp-1}

The setup consists of a wheeled robot driving around in a 2D indoor environment, with 17 tubes scattered throughout the space to act as landmarks (see \figref{fig:robot-pic}). The robot has an odometer to measure its translational and rotational speed, and uses a laser rangefinder to measure its range to the cylindrical landmarks. The groundtruth positions of the robot are recorded using a Vicon motion capture system. All data are logged at 10 Hz. There are approximately 4 visible landmarks at any given time. The robot state, input, landmark position, and measurement are in the same form as in~\eqref{eq:simu-1-defs}, where we adopt the common practice of using interoceptive measurments as inputs in the process model~\cite{probabilistic-robotics}. We use the lifting functions given in~\eqref{eq:simu-1-lifts}.

 \begin{figure}[!h]
 \centering
 \vspace{5pt}
   \includegraphics[width=0.97\columnwidth]{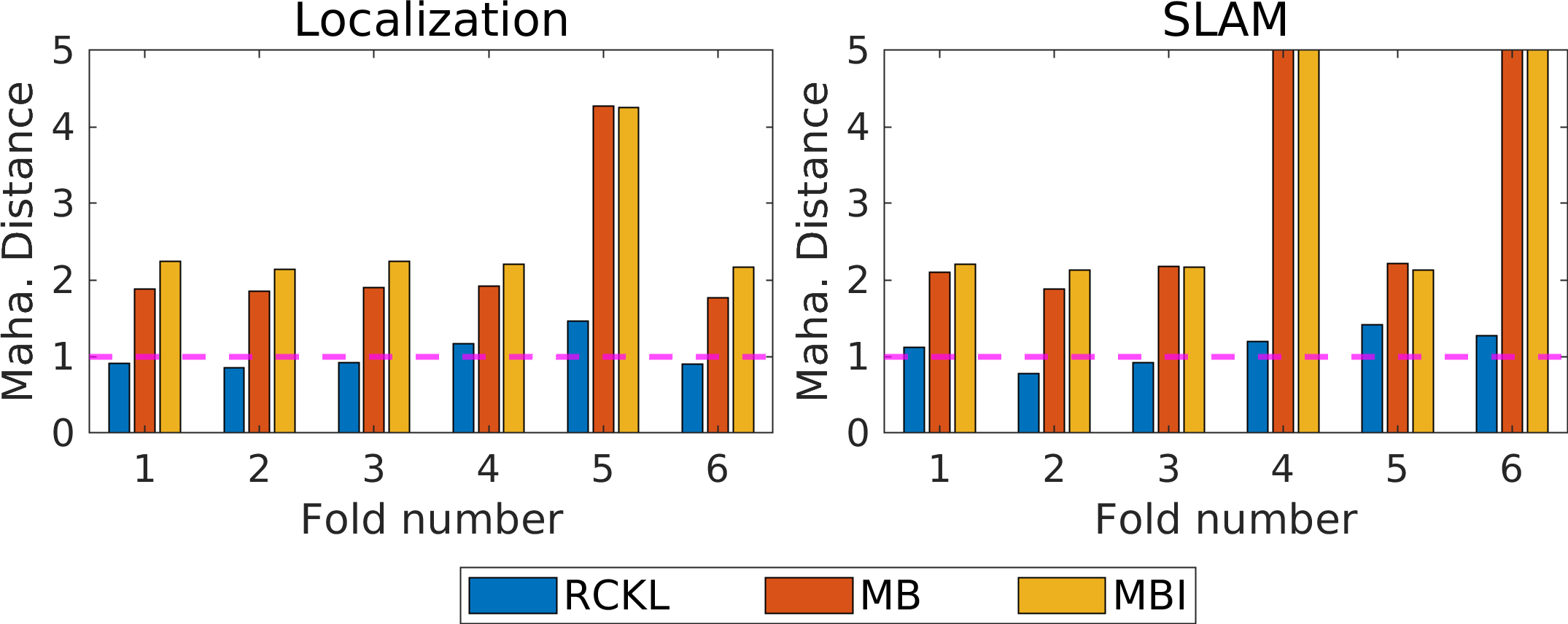}
\caption{\textit{Mahalanobis distances of localization and SLAM for Experiment 1}: Mahalanobis distances of the outputs of RCKL, MB, and MBI (whose robot-to-rangefinder distance is offset by 10 cm), on the 6 folds of the dataset. For localization, RCKL's Mahalanobis distances are close to 1, signifying that RCKL-Loc is consistent. MB's Mahalanobis distances are slighly higher than 1, signifying that its estimates are slightly overconfident, and MBI is similarly to marginally more overconfident than MB. For localization, MB and MBI are more overconfident in fold 5 than in the other folds as a result of converging to local minima. These trends are also present for SLAM on folds 1, 2, 3, and 5. On folds 4 and 6, the estimates of the model-based estimators, MB and MBI, are way overconfident as a result of convergence to local minima.}
\label{fig:lost-in-woods-maha}
\vspace{5pt}
\end{figure}
\begin{figure}[!h]
 \centering
  \subcaptionbox*{}{
 \includegraphics[width=0.85\columnwidth]{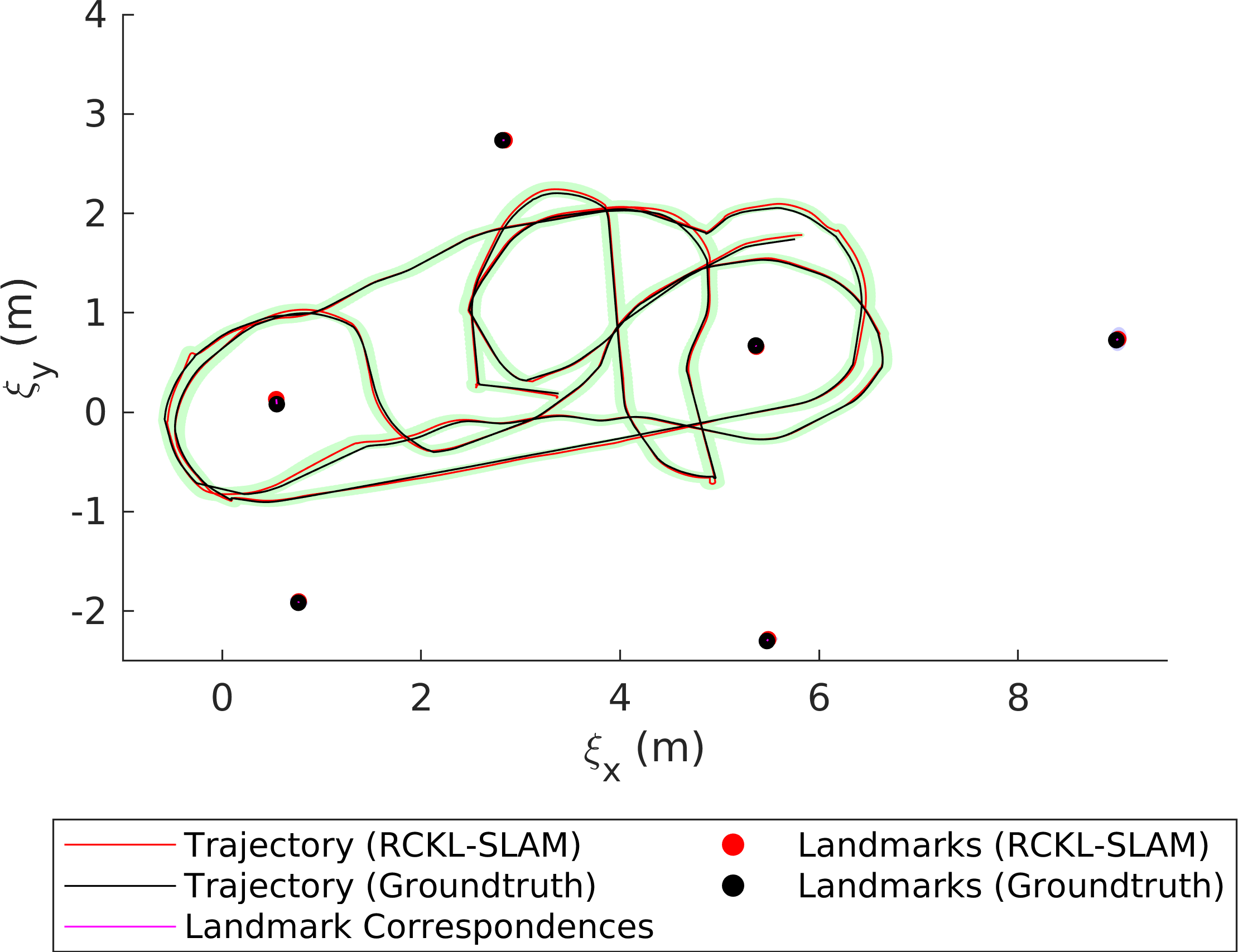}
   }
 \subcaptionbox*{}{
 \includegraphics[width=0.85\columnwidth]{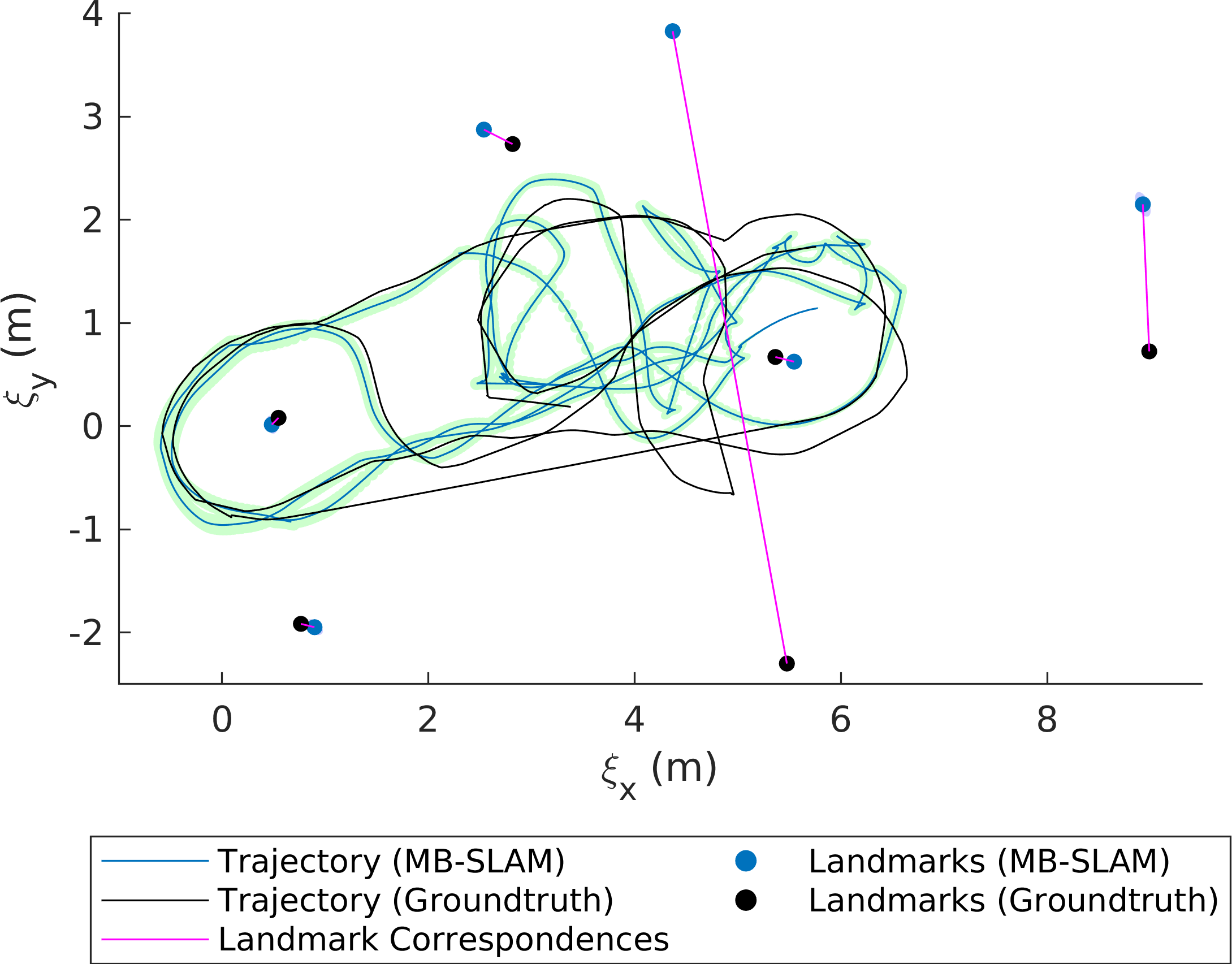}
}
 \caption{Visualization of SLAM output of Experiment 1 fold 6 for RCKL-SLAM (top) and MB-SLAM (bottom), showing the estimators' mean states and mean landmark positions compared to the groundtruth. The green regions and the grey regions are, respectively, the $3\sigma$ covariances of the trajectory and of the landmarks, though the $3\sigma$ bounds of the landmarks are barely visible. The pink lines show the landmark correspondences between the groundtruth and the estimators' outputs. RCKL-SLAM's trajectories and landmarks are close to the groundtruth and are generally within the estimated $3\sigma$ bounds. On the other hand, MB-SLAM has converged to a local minimum with one landmark on the opposite side of the trajectories, and its estimate is far from being within $3\sigma$ bounds of the groundtruth.}
 \label{fig:lost-in-woods-local-min}
\end{figure}

\begin{figure*}[!t]
\centering
 \subcaptionbox{RCKL-Loc}{
 \includegraphics[width=0.23\textwidth]{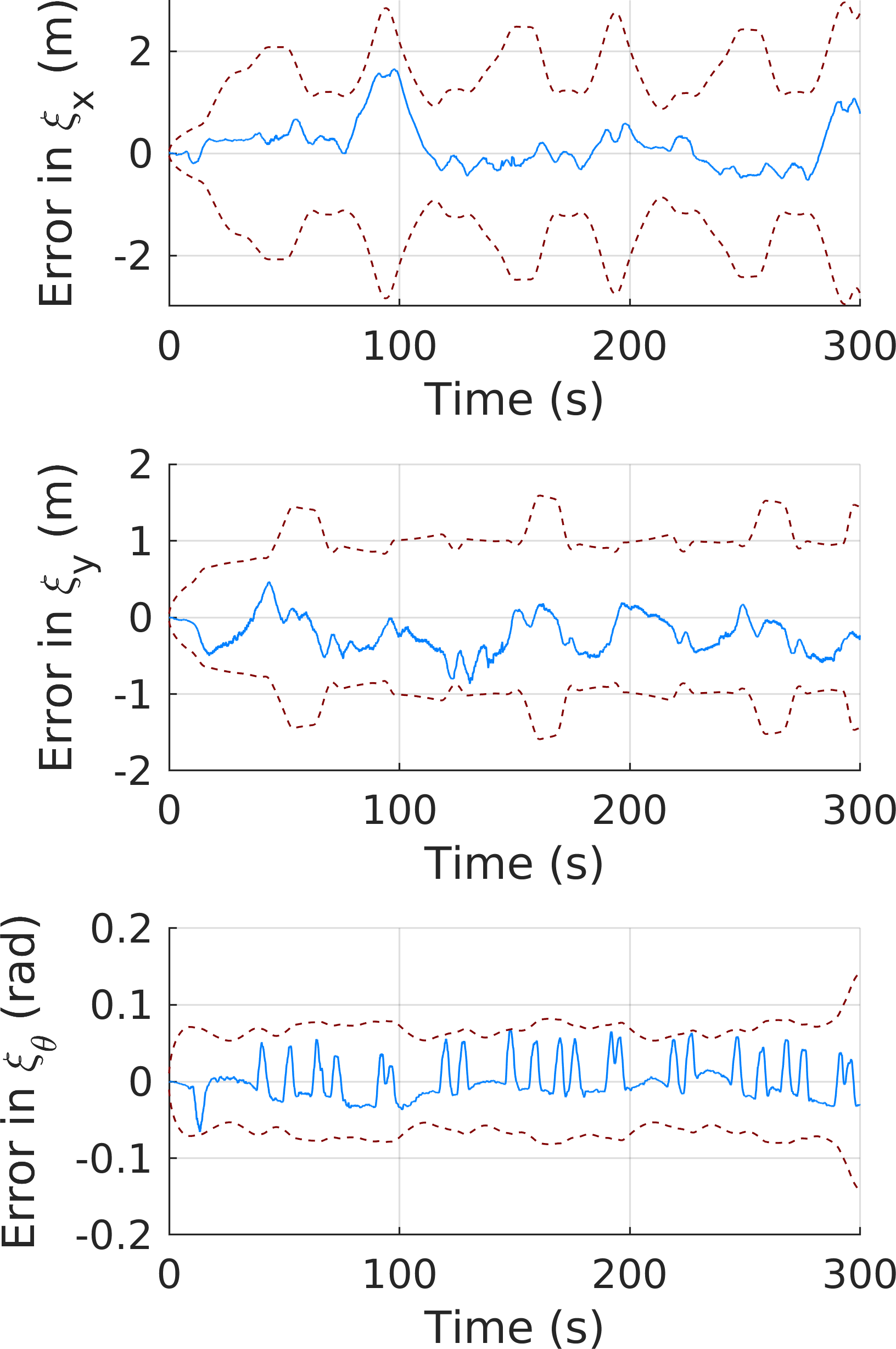}
 }
  \subcaptionbox{MB-Localization}{
 \includegraphics[width=0.23\textwidth]{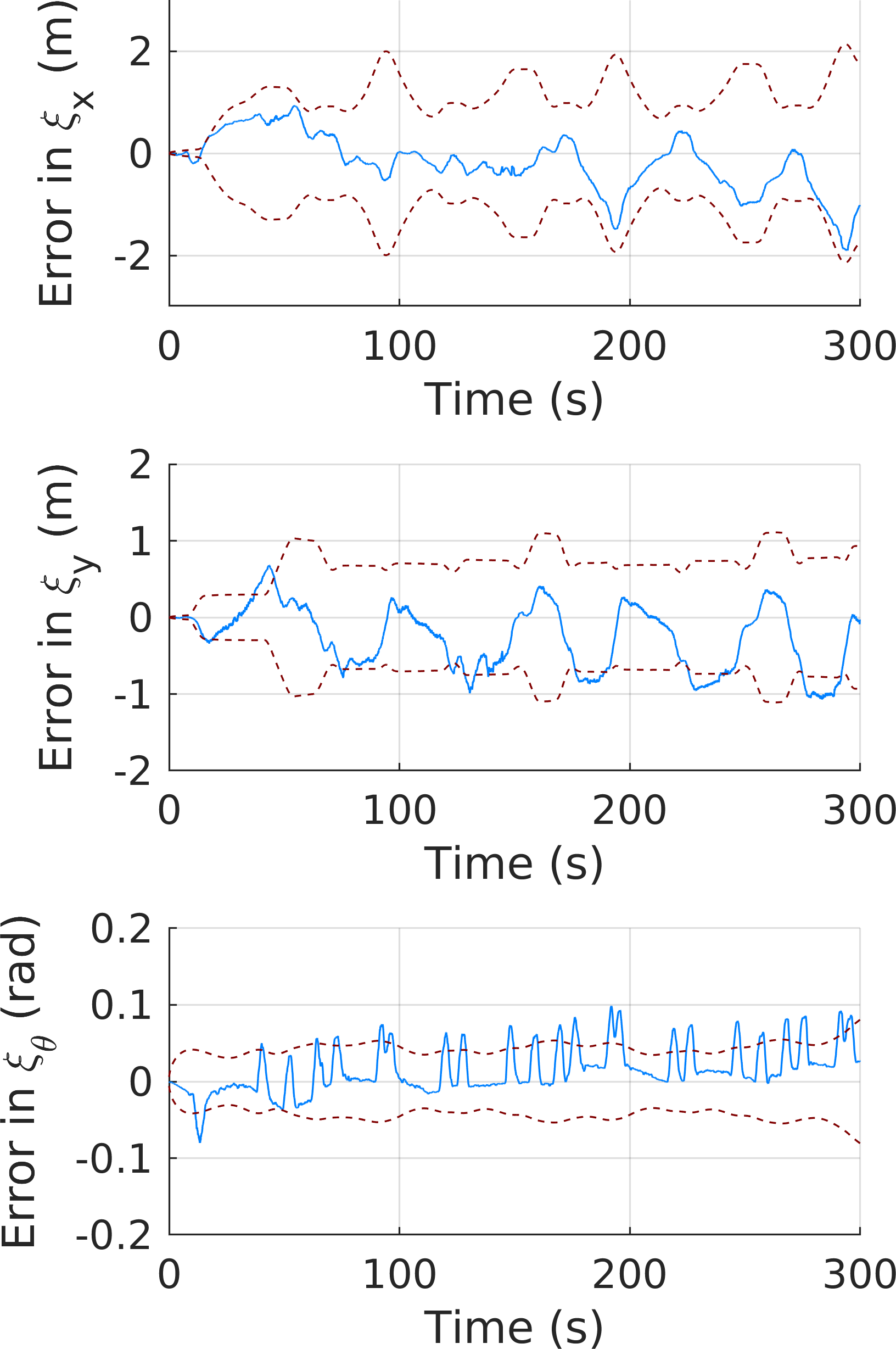}
 }
  \subcaptionbox{RCKL-SLAM}{
 \includegraphics[width=0.23\textwidth]{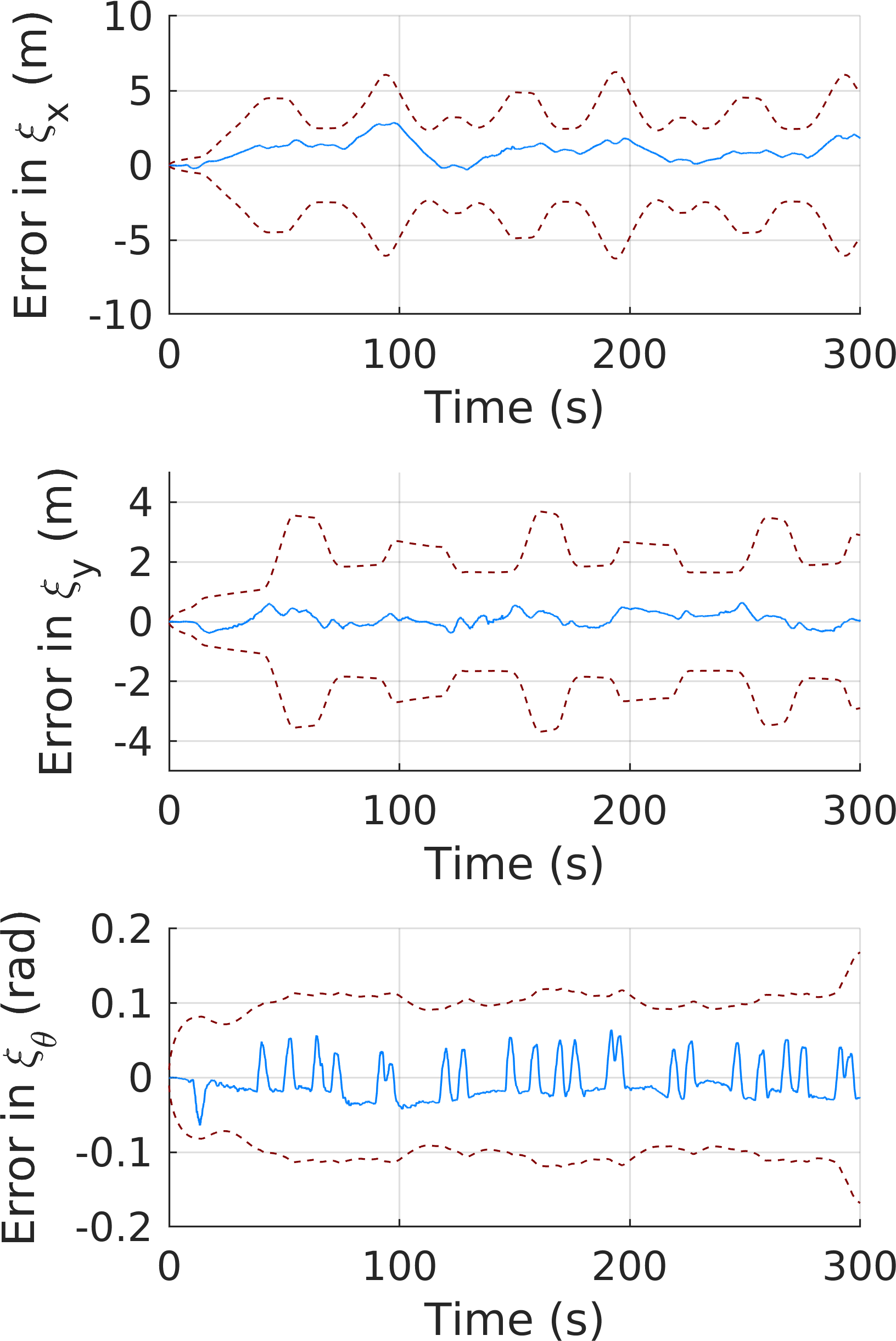}
 }
  \subcaptionbox{MB-SLAM}{
 \includegraphics[width=0.23\textwidth]{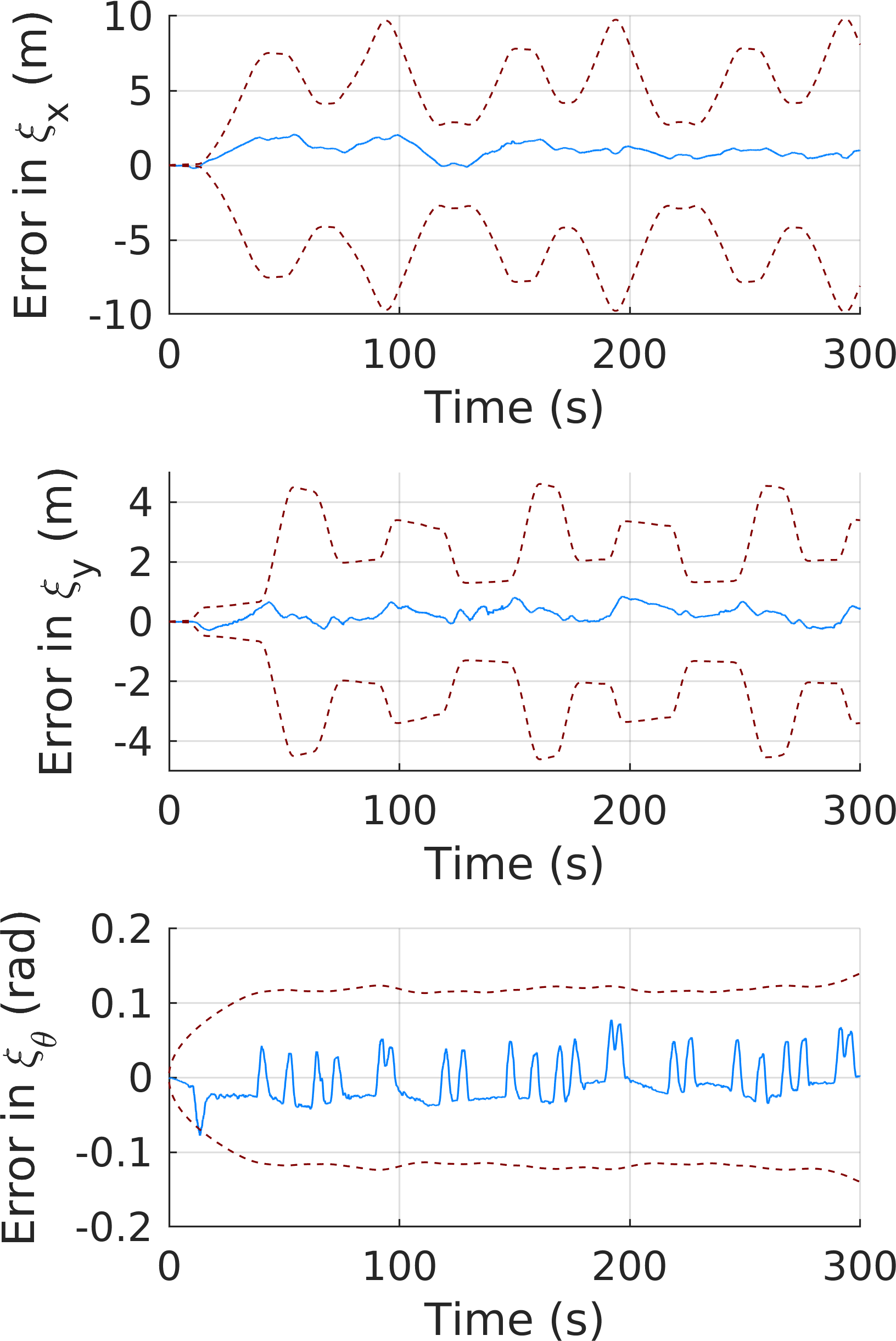}
 }
 \caption{Error plots for Experiment 2, comparing RCKL-Loc with MB-Localization (left two plots) and comparing RCKL-SLAM with MB-SLAM (right two plots). The blue lines represent the errors of the estimated trajectories, and the red envelopes represent the estimated $3\sigma$ covariance bounds. For both localization and SLAM, RCKL's errors are similar to or smaller than those of MB. RCKL's errors are bounded by the $3\sigma$ bounds, while MB's errors sometimes exceeds the $3\sigma$ bounds especially in localization. The shapes of the errors and the covariance envelopes of RCKL are similar to those of MB, suggesting that RCKL's outputs reflect the model reasonably well.}
 \label{fig:err-plots}
\end{figure*}

To empirically evaluate RCKL, we split the entire 20-minute dataset into training data and testing data. We use $\nicefrac{5}{6}$ of the trajectories for training, where measurements derived from only 11 of the 17 landmarks are used to train the measurement model. We test on the remaining $\nicefrac{1}{6}$ trajectories using measurements derived from only the 6 remaining landmarks. We assume that there are no factors such as slippage that would cause the robot to drive differently in one area of the room compared to another, and that the measurements are dependent only on the landmark positions relative to the robot poses. Thus, we perform data augmentation by adding translational and rotational transformations of the original training data to the training set.

We do 6-fold cross-validation by repeating the training/testing procedure for different sections of the dataset's trajectories, and compare the results of RCKL to MB. To demonstrate the advantages of the model-free framework, we also compare these results with MBI, whose robot-to-rangefinder distance in the measurement model is offset by 10 cm with respect to the generated measurements. Again, MB and MBI are using the same raw measurements but different prior models. The errors for RCKL, MB, and MBI are shown in \figref{fig:lost-in-woods-errors}, and their respective Mahalanobis distances are shown in \figref{fig:lost-in-woods-maha}.

In \figref{fig:lost-in-woods-errors}, we see that for both localization and SLAM, RCKL has similar error levels as MB but outperforms MBI. The Mahalanobis distances in \figref{fig:lost-in-woods-maha} show that RCKL is fairly consistent. Meanwhile, the model-based estimators are overconfident, and even more overconfident with the model offset. These results suggest that RCKL has learned a more accurate and more consistent model than MB and especially MBI. While RCKL appears to have not encountered any local minima in localization or SLAM, the model-based estimators have encountered local minima in 1 of the 6 folds for localization, and 2 of the 6 folds for SLAM (see \figref{fig:lost-in-woods-local-min}). This suggests that RCKL has better convergence properties.

 \begin{table}[!t]
\footnotesize
\caption{\textit{Ablation results of Experiment 2 for localization (top) and SLAM (bottom)}. The first row corresponds to RCKL, the second row corresponds to MB, the third row corresponds to RCKL with MB's measurement model, and the fourth row corresponds to RCKL with MB's process model. The fifth row is MBI, the model-based estimator whose measurement model has imperfect scaling factors. For both localization and SLAM, RCKL has similar or lower RMSEs than MB, and a more consistent Mahalanobis distance. RCKL is especially better than MBI, demonstrating the advantage of the data-driven method.}
\label{tab:gesling-loc-slam}
\begin{subtable}[h]{0.98\columnwidth}
\centering
\begin{NiceTabular}{|c|c|c|c|c|}
\hline
\textbf{Localization Method}
& \makecell{Position \\RMSE (m)} & \makecell{Orientation \\RMSE (rad)} 
& \makecell{Maha. \\distance} \\
\hline
\makecell{RCKL Process,\\RCKL Measurement} & $0.592$ & $\mbf{0.0246}$ & $\mbf{0.834}$ \\
\hline
\makecell{MB Process,\\MB Measurement} & $0.768$ & $0.0339$ & $6.167$ \\
\hline
\makecell{RCKL Process,\\MB Measurement} & $0.790$ & $0.0359$ & $1.092$ \\
\hline
\makecell{MB Process,\\RCKL Measurement} & $\mbf{0.582}$ & $0.0253$ & $6.161$ \\
\hline
\hline
\makecell{MBI\\(imperfect model)} & $8.187$ & $0.1249$ & $6.422$ \\
\hline
\end{NiceTabular}
\end{subtable}
\newline
\vspace{10pt}
\newline
\begin{subtable}[h]{\columnwidth}
\centering
\begin{NiceTabular}{|c|c|c|c|c|}
\hline
\textbf{SLAM Method}
& \makecell{Position \\RMSE (m)} & \makecell{Orientation \\RMSE (rad)} 
& \makecell{Landmark \\RMSE (m)}
& \makecell{Maha. \\distance} \\
\hline
\makecell{RCKL Process,\\RCKL Measurement} & $0.552$ & $0.0293$ & $\mbf{1.073}$ & $0.865$ \\
\hline
\makecell{MB Process,\\MB Measurement} & $\mbf{0.495}$ & $\mbf{0.0280}$ & $1.233$ & $6.154$ \\
\hline
\makecell{RCKL Process,\\MB Measurement} & $0.553$ & $0.0289$ & $1.298$ & $\mbf{0.971}$ \\
\hline
\makecell{MB Process,\\RCKL Measurement} & $1.781$ & $0.0699$ & $2.307$ & $6.956$ \\
\hline
\hline
\makecell{MBI\\(imperfect model)} & $1.428$ & $0.0638$ & $7.773$ & $6.240$ \\
\hline
\end{NiceTabular}
\end{subtable}
\end{table}

\subsection{Experiment 2: Golf Cart with RFID Measurements}
\label{sec:exp-golf-cart}
For this experiment, we train on Dataset A3 and test on 3000 steps of Dataset A1 of~\cite{gesling}. The setup for these two datasets consists of a cart driving on a golf course. The robot is equipped with a transponder with four radio-frequency (RF) antennae mounted on the four corners of the robot. As it moves, it measures ranges to RF tags placed on top of ten traffic cones that are scattered around the course. The range measurements are sporadic, occuring only at certain points on the trajectory. The measurements for each transponder contain a rather large scaling factor, corresponding to $\mu$ in~\eqref{eq:simu-1-defs}, of around 1.2. The groundtruth positions of the robot and of the cones are found with GPS. The control input consists of linear and angular velocities found with a fibre-optic gyro and wheel encoders. With just dead reckoning, the test trajectory drifts in orientation owing to repeated turns in the same direction~\cite{gesling}. This drift will be corrected by localization or SLAM.

Similar to Experiment 1, the robot state, input, landmark position, and measurement are in the same form as in~\eqref{eq:simu-1-defs}, and the same form of lifting functions are used as in~\eqref{eq:simu-1-lifts}. The landmark locations in A1 and A3 are the same. As such, we train on A3 with measurements from only 5 landmarks, and test on A1 with measurements from the 4 remaining landmarks, ignoring the one landmark with no measurements. We treat each RFID tag as an independent measurement model, so there are three sets of $\{\mbf{C},\mbf{R}\}$ to train, ignoring the one tag with only two measurements. We do a similar data augmentation procedure as Experiment 1, adding translational and rotational transformations on the original training data into the training set. We compare against MB, whose model contains the correct scaling factors, and against MBI, whose model uses a scaling factor of 1. The error plots for RCKL and the model-based estimators are shown in \figref{fig:err-plots}. A visualization of the outputs of RCKL-Loc and RCKL-SLAM can be seen in \figref{fig:gesling_redkooploc_redkoopslam}. As an ablation study, we also perform estimation with parts of RCKL's learned models swapped out with MB's prior models. The RMSEs and Mahalanobis distances of the above estimators are shown in \tabref{tab:gesling-loc-slam}.

\figref{fig:err-plots} qualitatively validates RCKL, since the errors and covariances of RCKL appear similar to those of MB. Quantitatively, we see from \tabref{tab:gesling-loc-slam} that RCKL has similar or better accuracy and consistency than MB in both localization and SLAM. This suggests that RCKL's learned models are better than MB's prior models. The ablation results in \tabref{tab:gesling-loc-slam} also show that swapping to RCKL's measurement model reduces more error than swapping to RCKL's process model. This suggests that RCKL's measurement model is especially better than MB's measurement model, while the two process models are of similar quality. Finally, RCKL is vastly better than MBI, demonstrating the advantage of the data-driven framework.

\subsection{Computation Time}
\begin{figure}[t!]
\centering
 \includegraphics[width=0.95\columnwidth]{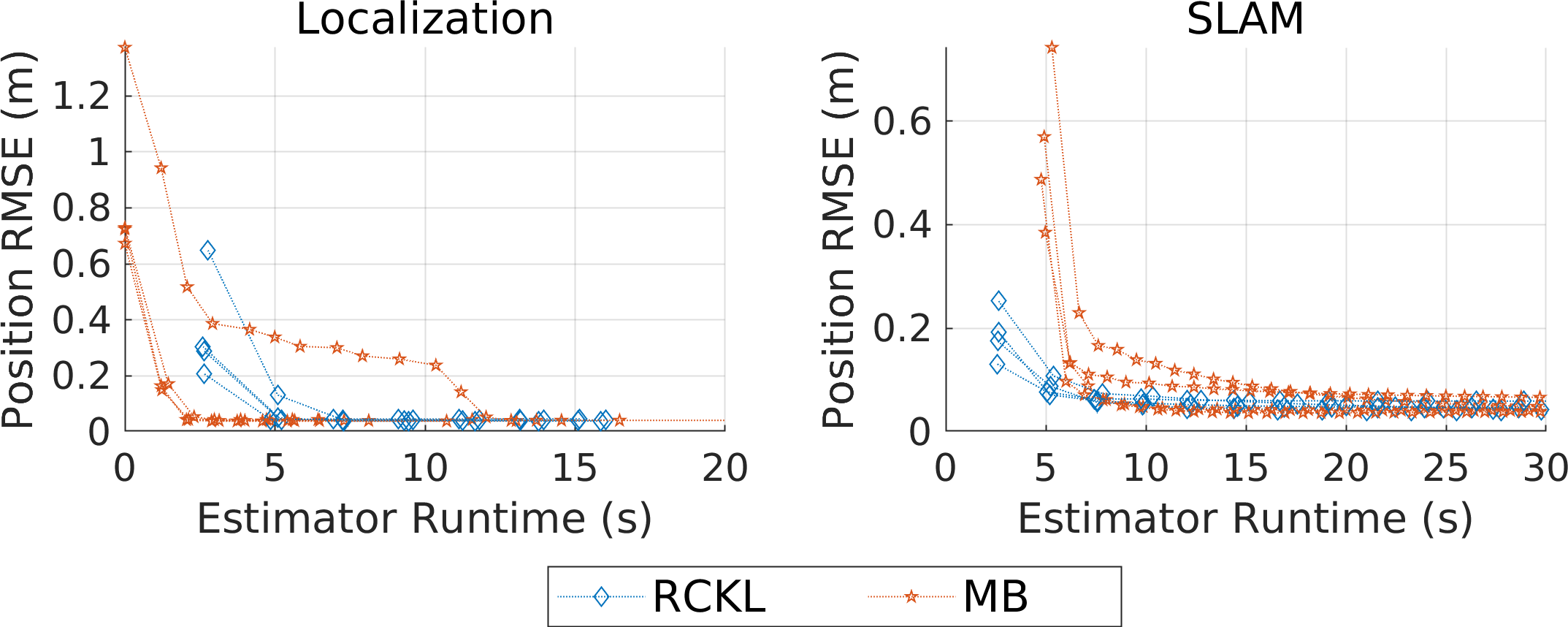}
 \caption{Position RMSE vs. estimator runtime for Experiment 1 for folds 1, 2, 3, 4, and 6 for localization, and folds 1, 2, 3, and 5 for SLAM, ignoring the folds where MB converges to a local minimum. For both RCKL and MB, markers are placed at every iteration, and the first markers are placed at the end of their initialization procedures. On all of these 200-second trajectories, both estimators converged to near the groundtruth within 30 seconds.}
 \label{fig:computation} 
\end{figure}
We briefly discuss the computation time of RCKL compared with that of the model-based estimators. It is difficult to compare the convergence speed of RCKL vs. MB based on cost tolerances since the two algorithms have vastly different cost functions. Thus, we instead view convergence as being sufficiently close to the groundtruth. In this sense, RCKL takes 1-3 times longer to converge than MB and MBI, depending on the environment setup. As an example, we compare the computation time of CPU-based implementations of RCKL and MB on an Intel Core i7-9750H Processor. We show in \figref{fig:computation} the estimator runtime for the folds where both estimators converged near the groundtruth in Experiment 1. Each fold contains 200 seconds of data, consisting of 2000 timesteps (and inputs) and about 8000 range measurements. As seen in \figref{fig:computation}, both estimators converge near the groundtruth in less than 30 seconds. RCKL has the same Big-O inference complexity as the model-based estimators for both localization and SLAM, scaling linearly with the number of timesteps. The model-based estimators work with smaller matrices and thus have faster iterations than RCKL. However, they are not substantially faster overall because they usually require more iterations to converge near the groundtruth.

\subsection{Hyperparameter Tuning}
\begin{figure}[t!]
\centering
 \includegraphics[width=0.95\columnwidth]{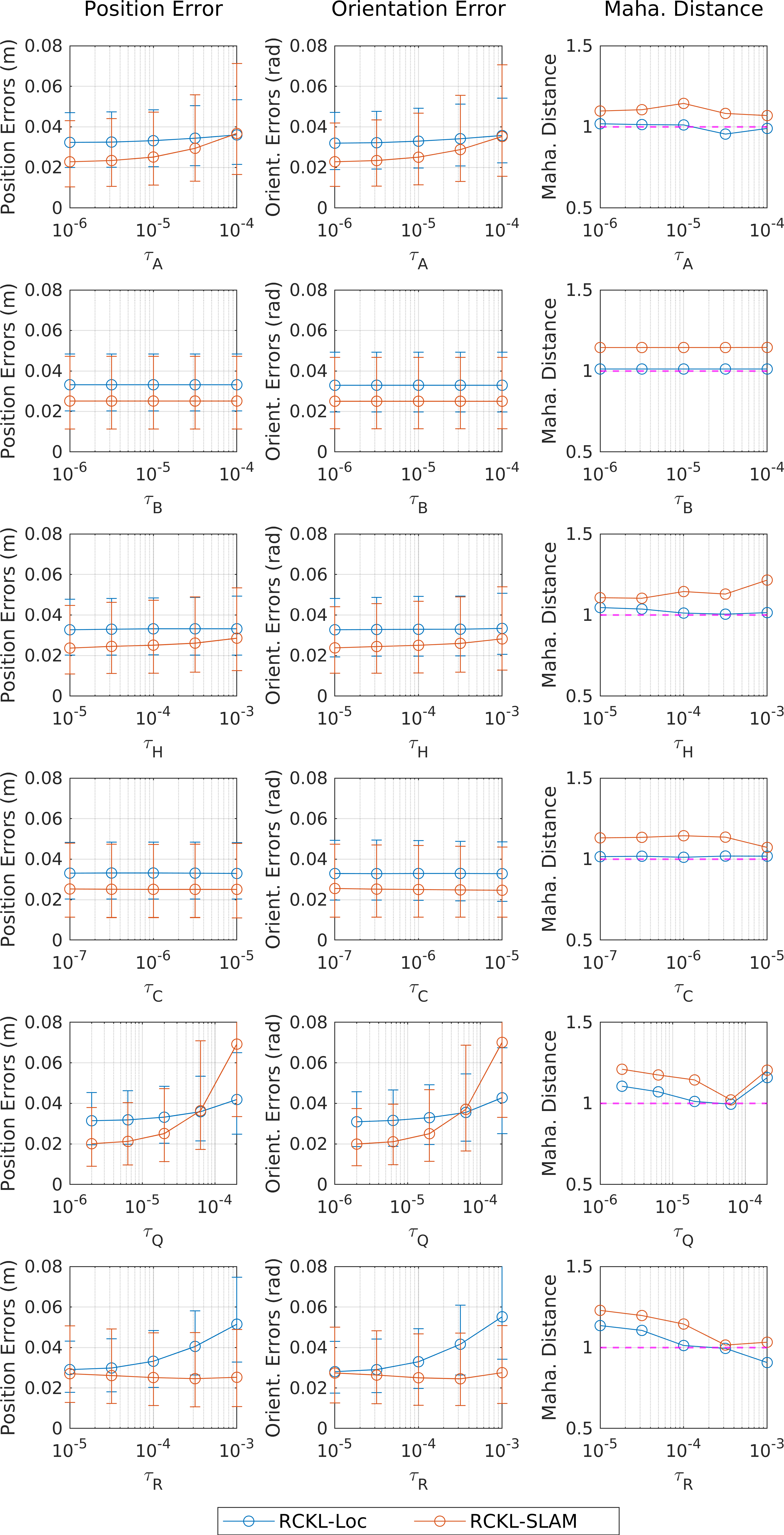}
 \caption{\textit{Hyperparameter sensitivity test}: Position errors, orientation errors, and Mahalanobis distances of RCKL-Loc and RCKL-SLAM results on Experiment 1 (all 6 folds) vs. different hyperparameter values ($\tau_A,\tau_B,\tau_H,\tau_C,\tau_Q,\tau_R$). For all plots, the middle marker is placed at the hand-tuned hyperparameter value used to generate the results in Section~\ref{sec:exp-1}. In each row, a hyperparameter is varied an order of magnitude above and below its hand-tuned value, while all other hyperparameters are set to their hand-tuned values. In the first two columns, we show the 25th, 50th, and 75th percentiles of the errors by representing them with the error bars' lower bounds, the circle markers, and the error bars' upper bounds. In the third column, the markers represent the mean Mahalanobis distances.}
 \label{fig:hyper-sens-test} 
\end{figure}
We discuss our procedure and recommendations for tuning the hyperparameters involved in RCKL. The main hyperparameters unique to this algorithm are the regularizers in~\eqref{eq:loss-fn-2}, namely $\tau_A,\tau_B,\tau_H,\tau_C,\tau_Q,\tau_R$. We used a different set of hyperparameter values for each of the experimental scenarios presented (Simulation 1, Simulation 2, Experiment 1, Experiment 2). We hand-tuned these values by manually adjusting them until a desired performance is reached on a validation dataset. Hyperparameters were never trained nor tuned on the test dataset. To achieve maximum performance, we recommend further refining the hyperparameters through a grid search or other hyperparameter optimization methods~\cite{hyper-opt}. In our experiments, we saw that RCKL's performance is sensitive to only a few hyperparameter values. To show this, we performed a sensitivity analysis on the $\tau$'s used in Experiment 1. \figref{fig:hyper-sens-test} shows the accuracy and consistency of RCKL-Loc and RCKL-SLAM as we individually vary each of the $\tau$'s within an order of magnitude above and below our hand-tuned values. We see that RCKL's accuracy and consistency does not change significantly across the tested ranges of $\tau_A, \tau_B, \tau_H,$ and $\tau_C$. The covariance regularizers, $\tau_Q$ and $\tau_R$, have larger effects on performance. Notably, the estimates are generally less consistent if $\tau_Q$ and $\tau_R$ are set either too low or too high, and generally less accurate if $\tau_Q$ and $\tau_R$ are set too high.

There are other settings that can be classified as hyperparameters, such as the lifting functions used and parameters within the SQP (e.g., $\mu$ in the line search acceptance criterion; see \eqref{eq:accept-crit} in Appendix~\ref{sec:sqp-formulation}). We recommend using standard methods found in literature to choose these parameters. For example, we chose SERFFs as our lifting functions because they are universal~\cite{rff}, but more methodical Koopman identification methods~\cite{dahdah_pykoop_2022},~\cite{pysindy} can be used as well.

%% file: sections/9_Conclusion.tex
\section{Conclusion}
\label{sec:discussion_conclusion}
The results highlight the advantages of RCKL's data-driven approach to localization and SLAM. When the classic model-based estimator that relies on Gauss-Newton has access to perfect models in simulation, it has similar or slightly better accuracy than RCKL. However, RCKL has similar or slightly better accuracy than the model-based estimator with an imperfect model. On real-world datasets, RCKL is generally more accurate and more consistent than the classic model-based estimator, and even more so when comparing with classic estimators with imperfect models. While the performance of the model-based estimators depends on the validity of their prior models, RCKL simply applies the high-dimensional model learned through data. In addition, RCKL appears to have the unanticipated benefit of being less prone to local minima.


Many avenues deserve interest for future work. It is worth investigating the origin of RCKL's improvement in convergence properties as a result of lifting estimation into a high-dimensional space. Another direction is to investigate updating the solution incrementally for real-time localization or SLAM, possibly by applying factor graph solvers~\cite{incopt},~\cite{constrained-opt-nav} or smoothing methods~\cite{ics} in the lifted space. As well, we can broaden the model types that can be converted into lifted forms, making a further step towards data-driven state estimation for general robotics systems. Moreover, many of the techniques developed in this paper could potentially be applied for other applications, such as data-driven optimal control of nonlinear systems.

%% file: sections/Appendix_arxiv.tex
\appendix
\subsection{Solving UKL-Loc with the RTS smoother}
\label{sec:unconstr-loc}
Here, we outline the process of applying the RTS smoother to the system in~\eqref{eq:ltv-system} to solve UKL-Loc. This may be a more efficient solution process than the sparse Cholesky solver. Note that the two methods are algebraically equivalent~\cite[\S3]{barfoot-txtbk}.

With the landmarks given, we convert the measurement model to be linear-Gaussian using the same trick as in the process model. Observe that
\begin{equation}\label{eq:meas-trick}
\mbs{\ell}_j \otimes \mbf{x}_k = (\mbs{\ell}_j \otimes \mbf{1}) \mbf{x}_k = (\mbf{1} \otimes \mbf{x}_k) \mbs{\ell}_j. 
\end{equation}
With this, we can rewrite the measurement equation at each timestep $k$ by stacking the measurements, $\mbf{y}_k$, and defining matrices, $\mbf{C}^\prime$ and $\mbf{R}^\prime$ as
\begin{gather}\label{eq:stack}
 \mbf{y}_k = \bbm \mbf{y}_{k,1} \\ \vdots \\ \mbf{y}_{k,V} \ebm, \quad
 \mbf{C}^\prime = \bbm \mbf{C} (\mbs{\ell}_1 \otimes \mbf{1}) \\ \vdots \\ \mbf{C}(\mbs{\ell}_V \otimes \mbf{1}) \ebm, \quad
 \mbf{R}^\prime = \diag(\mbf{R},\dots,\mbf{R}),
\end{gather}
where $\mbf{y}_k$ is filled with known values of the landmarks and $\mbf{R}^\prime$ is $\mbf{R}$ stacked diagonally $V$ times. The system in~\eqref{eq:ltv-system} can then be written as
\begin{subequations}
\label{eqn:loc-sys}
\begin{align}
 \mbf{x}_{k} & = \mbf{A}_{k-1} \mbf{x}_{k-1} + \mbf{v}_k + \mbf{w}_k, \quad &k=1,\dots,K, \\
 \mbf{y}_k & = \mbf{C}^\prime \mbf{x}_k + \mbf{n}_k, \quad &k=0,\dots,K,
\end{align}
\end{subequations}
where $\mbf{w}_k \sim \mathcal{N}(\mbf{0},\mbf{Q})$, $\mbf{n}_{k} \sim \mathcal{N}(\mbf{0},\mbf{R}^\prime)$. If any measurement is missing, we can make $\mbf{C}^\prime$ and $\mbf{R}^\prime$ be time-varying and omit the corresponding entries when constructing $\mbf{y}_k$, $\mbf{C}^\prime_k$, and $\mbf{R}^\prime_k$. In any case, we have converted the bilinear time-invariant system in~\eqref{eq:bilin-inv} into a linear time-varying (LTV) system. We can then apply the standard RTS smoother~\cite{barfoot-txtbk} to~\eqref{eqn:loc-sys} to solve for the states in $\mathcal{O}(N^3KV)$ time. The output is $\mbf{x}_k \sim \mathcal{N}(\hat{\mbf{x}}_k, \hat{\mbf{P}}_{\mbf{x},k})$, where $\hat{\mbf{x}}_k$ is the state mean and $\hat{\mbf{P}}_{\mbf{x},k}$ is the covariance at timestep $k$, for all $k=0,\dots,K$. In contrast to~\cite{kooplin}, by using a measurement model that is individually applicable to each landmark, we have generalized localization to environments with new landmark positions.

\subsection{Gauss-Newton Setup for UKL-SLAM}
\label{sec:batch-slam-setup}
We use Gauss-Newton optimization to solve~\eqref{eq:unconstr-prob} for $\mbs{x}$ and $\mbs{\ell}$. Given an operating point, $\mbs{q}_{\text{op}}$, the Gauss-Newton approximation of~\eqref{eq:cost-orig} yields
\begin{gather}\label{eq:cost}
 J(\mbs{q}) = J(\mbs{q}_{\text{op}} + \delta \mbs{q}) \approx J(\mbs{q}_{\text{op}}) - \mbs{g}^T \delta\mbs{q} + \frac{1}{2} \delta\mbs{q}^T \mbs{F} \delta\mbs{q},
\end{gather}
where
\begin{subequations}\label{eq:slam-block-2}
\begin{gather}
 \mbs{F} = \mbs{H}^T \mbs{W}^{-1} \mbs{H}, \quad
 \mbs{g} = \mbs{H}^T \mbs{W}^{-1} \mbs{e}(\mbs{q}_{\text{op}}), \\
 \mbs{H} = \bbm \mbs{A}^{-1} & \mbf{0} \\ \mbs{G}_{\mbs{x}} & \mbs{G}_{\mbs{\ell}} \ebm, \quad
 \mbs{A}^{-1} = \bbm
  -\mbf{A}_0 & \mbf{1} & & \\
  & \ddots & \ddots & \\
  & & -\mbf{A}_{K-1} & \mbf{1}
  \ebm, \label{eq:slam-block-2-ha} \\
 \mbs{G}_{\mbs{x}} = \diag(\mbf{G}_{\mbf{x},1},\dots,\mbf{G}_{\mbf{x},K}), \quad
 \mbs{G}_{\mbs{\ell}}^T = \bbm \mbf{G}_{\mbs{\ell},1}^T & \cdots & \mbf{G}_{\mbs{\ell},K}^T \ebm, \\
 \mbf{G}_{\mbf{x},k}^T = \bbm \mbf{G}_{\mbf{x},k,1}^T & \cdots & \mbf{G}_{\mbf{x},k,V}^T \ebm, \quad
 \mbf{G}_{\mbs{\ell},k} = \diag(\mbf{G}_{\mbs{\ell},k,1},\dots,\mbf{G}_{\mbs{\ell},k,V}),
\end{gather}
\end{subequations}
where, using~\eqref{eq:meas-trick}, the measurement Jacobians are
\begin{subequations}
 \begin{gather}
 \mbf{G}_{\mbf{x},k,j} = \frac{\partial \mbf{y}_{k,j}}{\partial \mbf{x}} \biggr \rvert_{\mbf{x}_{\text{op},k}, \mbs{\ell}_{\text{op},j}} = \mbf{C}(\mbs{\ell}_{\text{op},j} \otimes \mbf{1}), \\
 \mbf{G}_{\mbs{\ell},k,j} = \frac{\partial \mbf{y}_{k,j}}{\partial \mbs{\ell}} \biggr \rvert_{\mbf{x}_{\text{op},k}, \mbs{\ell}_{\text{op},j}} = \mbf{C}(\mbf{1} \otimes \mbf{x}_{\text{op},k}).
 \end{gather}
\end{subequations}
Note that $\mbs{F}$ is at least positive semidefinite when $\mbf{Q}$ and $\mbf{R}$ are positive definite, and $\mbs{F}$ is positive definite when the lifted system is observable~\cite{barfoot-txtbk}. Observability of the lifted system depends on the test data and the Jacobians in $\mbs{H}$~\cite{slam-observable}. We found that in simulation and experimental testing, when UKL has a good initialization, $\mbs{F}$ is positive definite and successful convergence of the Gauss-Newton algorithm is realized. See Section~\ref{sec:sqp-init-short} for an initialization procedure. This empirically suggests that when the original system is observable, the lifted system is also observable once given good priors in~\eqref{eq:loss-fn-2}, reasonable lifting functions, and sufficient training data. An interesting avenue for future work is explicitly enforcing observability when learning $\mbf{A}$, $\mbf{B}$, $\mbf{H}$, and $\mbf{C}$.

The optimal update, $\delta \mbs{q}$, satisfies
\begin{equation}
\label{eqn:slam-eqn}
 \mbs{F} \delta {\mbs{q}} = \mbs{g},
\end{equation}
where we can solve for $\delta \mbs{q}$ by following the standard approach for classic batch SLAM~\cite{barfoot-txtbk}. As we will see in Appendix~\ref{sec:batch-slam}, the complexity of solving~\eqref{eqn:slam-eqn} is $\mathcal{O}(N^3(V^3+V^2K))$ when we have more timesteps than landmarks, or $\mathcal{O}(N^3(K^3+K^2V))$ when we have more landmarks than timesteps.

As mentioned in Section~\ref{sec:ukl-slam-vars}, we can easily accomodate for missing or known variables. If $\mbf{y}_{k,j}$ is missing, we simply remove $\mbf{e}_{\mbf{y},k,j}$ from the cost and delete the corresponding blocks in $\mbs{R}^{-1}$, $\mbf{G}_{\mbf{x},k}$, and $\mbf{G}_{\mbs{\ell},k}$. If $\mbf{x}_k$ or $\mbs{\ell}_j$ is already known, we modify~\eqref{eqn:slam-eqn} to incorporate the known variable by setting $\mbf{x}_k$ or $\mbs{\ell}_j$ to its value, removing it from the list of variables in $\mbs{q}$, and moving its corresponding equation in $\mbs{F}\delta\mbs{q}$ to the right-hand side as part of $\mbs{g}$.

The CKL-SLAM problem~\eqref{eq:constr-prob} contains the same objective function as UKL-SLAM~\eqref{eq:unconstr-prob}. As we will see in Appendices~\ref{sec:sqp-formulation} and~\ref{sec:solve-sqp-lin-time}, CKL-SLAM's solution process uses the same Gauss-Newton approximation as described above. The case is similar for RCKL-SLAM, where owing to its slightly-modified cost function~\eqref{eq:cost-reduced}, the only required changes are to replace $\mbf{Q}^{-1} = \bbm \mbf{Q}_{\mbs{\xi}}^{-1} & \mbf{0} \\ \mbf{0} & \mbf{0} \ebm$ in~\eqref{eq:qr-struct} and $\mbf{A}_{k-1} = \bbm \mbf{A}_{\mbs{\xi}, k-1} \\ \mbf{0} \ebm$ in~\eqref{eq:slam-block-2-ha}.

\subsection{Solving UKL-SLAM in Linear Time}
\label{sec:batch-slam}
Although the linear system in~\eqref{eqn:slam-eqn} is very large, it can be efficiently solved with Cholesky factorization and by exploiting sparsity patterns. $\mbs{F}$ has the structure
\begin{gather}\label{eq:slam-f}
 \mbs{F} =  \bbm
 \mbs{A}^{-T}\mbs{Q}^{-1}\mbs{A}^{-1} + \mbs{G}_{\mbs{x}}^T \mbs{R}^{-1} \mbs{G}_{\mbs{x}} &
 \mbs{G}_{\mbs{x}}^T \mbs{R}^{-1} \mbs{G}_{\mbs{\ell}} \\
 \mbs{G}_{\mbs{\ell}}^T \mbs{R}^{-1} \mbs{G}_{\mbs{x}} &
 \mbs{G}_{\mbs{\ell}}^T \mbs{R}^{-1} \mbs{G}_{\mbs{\ell}}
 \ebm = 
 \bbm \mbs{F}_{\mbs{x}} & \mbs{F}_{\mbs{x}\mbs{\ell}} \\ \mbs{F}_{\mbs{x}\mbs{\ell}}^T & \mbs{F}_{\mbs{\ell}} \ebm,
\end{gather}
which exhibits the usual SLAM arrowhead pattern where $\mbs{F}_{\mbs{x}}$ is block-tridiagonal and $\mbs{F}_{\mbs{\ell}}$ is block-diagonal~\cite{barfoot-txtbk}. As discussed in Section~\ref{sec:subsec:unconstr-slam}, $\mbs{F}$ is usually positive definite. Then, when we have more robot poses than landmarks, we can use the lower-Cholesky decomposition on $\mbs{F}$,
\begin{gather}\label{eq:chol}
 \mbs{F}
 = 
 \underbrace{\bbm \mbs{L}_{\mbs{x}} & \mbf{0} \\ \mbs{L}_{\mbs{x}\mbs{\ell}} & \mbs{L}_{\mbs{\ell}} \ebm}_{\mbs{L}}
 \underbrace{\bbm \mbs{L}_{\mbs{x}}^T & \mbs{L}_{\mbs{x}\mbs{\ell}}^T \\ \mbf{0} & \mbs{L}_{\mbs{\ell}}^T \ebm}_{\mbs{L}^T}
 =
 \bbm \mbs{L}_{\mbs{x}}\mbs{L}_{\mbs{x}}^T & 
 \mbs{L}_{\mbs{x}}\mbs{L}_{\mbs{x}\mbs{\ell}}^T \\
 \mbs{L}_{\mbs{x}\mbs{\ell}}\mbs{L}_{\mbs{x}}^T & 
 \mbs{L}_{\mbs{x}\mbs{\ell}}\mbs{L}_{\mbs{x}\mbs{\ell}}^T + \mbs{L}_{\mbs{\ell}}\mbs{L}_{\mbs{\ell}}^T \ebm,
\end{gather}
where we can use the fact that $\mbs{F}_{11}$ is block-tridiagonal to efficiently compute the nonzero blocks in $\mbs{L}_{\mbs{x}\mbs{x}}$ (see~\cite[\S3]{barfoot-txtbk}), and then compute $\mbs{L}_{\mbs{x}\mbs{\ell}}$ and $\mbs{L}_{\mbs{\ell}\mbs{\ell}}$. For the right-hand side of~\eqref{eqn:slam-eqn}, $\mbs{g}$ has the structure
\begin{gather}
  \mbs{g} = \mbs{H}^T \mbs{W}^{-1}\mbs{e}_{\text{op}} = 
 \bbm \mbs{A}^{-T} \mbs{Q}^{-1} \mbs{e}_{\mbs{v},\text{op}} + \mbs{G}_{\mbs{x}}^T \mbs{R}^{-1} \mbs{e}_{\mbs{y},\text{op}} \\
 \mbs{G}_{\mbs{\ell}}^T \mbs{R}^{-1} \mbs{e}_{\mbs{y},\text{op}} \ebm
 = \bbm \mbs{g}_{\mbs{x}} \\ \mbs{g}_{\mbs{\ell}} \ebm,
\end{gather}
for which we can efficiently compute using the appropriate blocks. The solution process for~\eqref{eqn:slam-eqn} becomes first solving $\mbs{L}\mbs{p} = \mbs{g}$ for a placeholder variable $\mbs{p}$ using forward substitution, then solving $\mbs{L}^T \hat{\mbs{q}} = \mbs{p}$ for $\hat{\mbs{q}}$ using backward substitution. Thus,~\eqref{eqn:slam-eqn} can be solved without ever constructing the large sparse system, rather just working with the required blocks. The exact solution for $\hat{\mbs{q}}$ can be computed in $\mathcal{O}(N^3(V^3+V^2K))$ time. On the other hand, if we have more landmarks than poses, we can use the upper-Cholesky decomposition with a similar procedure, solving in $\mathcal{O}(N^3(K^3+K^2V))$ time. See~\cite[\S9]{barfoot-txtbk} for an example.

After iterating until convergence, we can also efficiently compute the covariances of the system. To find $\hat{\mbf{P}}_{\mbf{x},k}$ and $\hat{\mbf{P}}_{\mbs{\ell},j}$, we compute the diagonal blocks of $\mbs{F}^{-1}$. Since $\hat{\mbs{P}}_{\mbs{q}} = \mbs{F} = \mbs{L}\mbs{L}^T$~\cite{barfoot-txtbk}, we can use \cite{takahashi} to compute only the blocks of $\mbs{F}^{-1}$ corresponding to the nonzero blocks of $\mbs{L}$. This can be done without raising the computational complexity of SLAM given the sparsity pattern of $\mbs{L}$ present in SLAM problems~\cite{esgvi}. The final output is $\mbf{x}_k \sim \mathcal{N}(\hat{\mbf{x}}_k, \hat{\mbf{P}}_{\mbf{x},k})$, $\mbs{\ell}_j \sim \mathcal{N}(\hat{\mbs{\ell}}_j, \hat{\mbf{P}}_{\mbs{\ell},j})$.

\subsection{Formulating the SQP for (R)CKL-SLAM}
\label{sec:sqp-formulation}
We formulate an SQP~\cite{nocedal} to solve~\eqref{eq:constr-prob}. The Lagrangian is $L(\mbs{q}, \mbs{\lambda}) = J(\mbs{q}) - \mbs{\lambda}^T \mbs{h}(\mbs{q})$, where $\mbs{\lambda}^T = \bbm \mbs{\lambda}_{\mbs{x}}^T & \mbs{\lambda}_{\mbs{\ell}}^T \ebm$, $ \mbs{\lambda}_{\mbs{x}}^T = \bbm \mbs{\lambda}_{\mbf{x},1}^T & \cdots & \mbs{\lambda}_{\mbf{x},K}^T \ebm$, $\mbs{\lambda}_{\mbs{\ell}}^T = \bbm \mbs{\lambda}_{\mbs{\ell},1}^T & \cdots & \mbs{\lambda}_{\mbs{\ell},V}^T \ebm$ are the Lagrange multipliers. Using the same matrix structures as~\eqref{eq:slam-block-2}, we can write the Jacobian of the Lagrangian at an operating point, $\mbs{q}_{\text{op}}$ and $\mbs{\lambda}_{\text{op}}$, as $\frac{\partial L}{\partial \mbs{q}^T} \rvert_{\mbs{q}_{\text{op}}} = -\mbs{g} - \mbs{S}^T \mbs{\lambda}_{\text{op}}$, where $\mbs{g}$ is defined in~\eqref{eq:slam-block-2}, and $\mbs{S}$ is the Jacobian of the constraints, defined as
\begin{subequations}\label{eq:s}
\begin{gather}
 \mbs{S} = \frac{\partial \mbs{h}}{\partial \mbs{q}} \biggr\rvert_{\mbs{q}_{\text{op}}} = \bbm \mbs{S}_{\mbs{x}} & \mbf{0} \\ \mbf{0} & \mbs{S}_{\mbs{\ell}} \ebm, \\
 \mbs{S}_{\mbs{x}} = \diag(\mbf{S}_{\mbf{x},1},\dots,\mbf{S}_{\mbf{x},K}), \quad
 \mbs{S}_{\mbs{\ell}} = \diag(\mbf{S}_{\mbs{\ell},1},\dots,\mbf{S}_{\mbs{\ell},V}), \\
 \mbf{S}_{\mbf{x},k} = \frac{\partial \mbf{h}_{\mbf{x}}(\mbf{x})}{\partial \mbf{x}} \biggr\rvert_{\mbf{x}_{\text{op},k}}, \quad
 \mbf{S}_{\mbs{\ell},j} = \frac{\partial \mbf{h}_{\mbs{\ell}}(\mbs{\ell})}{\partial \mbs{\ell}} \biggr\rvert_{\mbs{\ell}_{\text{op},j}}.
\end{gather}
\end{subequations}
Note that $\mbs{S}$ is block-diagonal owing to the blockwise structure of the constraints.

We use a generalized Gauss-Newton approximation of the Hessian of the Lagrangian~\cite[\S2]{constrained-cov},~\cite[\S3.2]{opt-survey}, which simply corresponds to the Gauss-Newton approximation of the objective function:
\begin{align}
 \frac{\partial^2 L}{\partial \mbs{q} \partial \mbs{q}^T} \biggr\rvert_{\mbs{q}_{\text{op}}} \approx \left(\frac{\partial J}{\partial \mbs{q}^T}\biggr\rvert_{\mbs{q}_{\text{op}}}\right) \left(\frac{\partial J}{\partial \mbs{q}}\biggr\rvert_{\mbs{q}_{\text{op}}}\right) = \mbs{F},
\end{align}
where $\mbs{F}$ is defined in~\eqref{eq:slam-block-2}. In contrast to the full Hessian of the Lagrangian, which may be indefinite, the Gauss-Newton approximation is guaranteed to be at least positive semidefinite. The condition for a valid descent direction from the SQP is that the reduced Hessian, or the Hessian projected onto the tangent space of the constraints, is positive definite \cite[p.~452]{nocedal}. With our Gauss-Newton approximation, the reduced Hessian is at least positive semidefinite. It is also empirically full rank in our experiments, thus meeting the SQP's requirement. See Appendix~\ref{sec:solve-sqp-lin-time} for more discussion on this condition.

We now follow the procedure in \cite{nocedal} to set up the SQP system of equations at each iteration as
\begin{gather}\label{eq:sqp}
 \bbm \mbs{F} & \mbs{S}^T \\ \mbs{S} & \mbf{0} \ebm
 \bbm \delta\mbs{q} \\ -\delta\mbs{\lambda} \ebm
 =
 \bbm \mbs{g} + \mbs{S}^T \mbs{\lambda}_{\text{op}} \\ -\mbs{h} \ebm,
\end{gather}
then update the operating point of the primal variable and the multiplier as $\mbs{q}_{\text{op}} \leftarrow \mbs{q}_{\text{op}} + \alpha \delta \mbs{q}$ and $\mbs{\lambda}_{\text{op}} \leftarrow \mbs{\lambda}_{\text{op}} + \alpha \delta\mbs{\lambda}$, with an appropriate step size $\alpha$. We find $\alpha$ using a backtracking line search with an acceptance criterion \cite{nocedal},
\begin{gather}\label{eq:accept-crit}
 \phi_1(\mbs{q}_{\text{op}} + \alpha \delta\mbs{q}, \mu) \leqslant \phi_1(\mbs{q}_{\text{op}}, \mu) + \eta \alpha \frac{\partial J}{\partial \mbs{q}^T}\biggr\rvert_{\mbs{q}_{\text{op}}} \delta\mbs{q},
\end{gather}
where $\eta \in (0,1)$, $\mu > 0$, and $\phi_1(\mbs{q}, \mu) = J(\mbs{q}) + \mu ||\mbs{h}(\mbs{q})||_1$ is the $\mathcal{L}_1$ merit function.

We can now solve for $(\delta\mbs{q}, \delta\mbs{\lambda})$ and iterate to convergence, yielding $(\mbs{q}^*, \mbs{\lambda}^*)$. We then recover the mean estimates in the original space, $\hat{\mbs{\zeta}}$, by picking off the top blocks of the lifted estimates in $\mbs{q}^*$. Although the linear system in~\eqref{eq:sqp} is quite large, as we will see in Appendix~\ref{sec:solve-sqp-lin-time}, we can solve it efficiently by exploiting the SLAM arrowhead structure in $\mbs{F}$ and the block-diagonal structure in $\mbs{S}$.

\subsection{Solving (R)CKL-SLAM in Linear Time}
\label{sec:solve-sqp-lin-time}
In this section, we efficiently solve the large linear system in \eqref{eq:sqp} for $\delta\mbs{q}$ and $\delta\mbs{\lambda}$, and also describe the requirements on the SQP Hessian for valid descent directions. We define $\mbs{t} = \delta\mbs{q}$ and $\mbs{\nu} = \delta\mbs{\lambda}$ for brevity.

We solve the SQP with the nullspace method~\cite{nocedal}. This method is recommended for systems where the degrees of freedom of the free variables is small. For our system, this number simply corresponds to the degrees of freedom of the original system in~\eqref{eq:control-affine}, since the rest of the variables are lifted from, and thus constrained by, the original variables. As we will see,~\eqref{eq:sqp} can be solved in the same time complexity as the unconstrained problem: $\mathcal{O}(N^3(V^3 + V^2 K))$ when we have more timesteps than landmarks, or $\mathcal{O}(N^3(K^3 + K^2 V))$ when we have more landmarks than timesteps.

We break down $\mbs{t}$ and $\mbs{v}$ as $\mbs{t} = \bbm \mbs{t}_{\mbs{x}} \\ \mbs{t}_{\mbs{\ell}} \ebm$, $\mbs{v} = \bbm \mbs{v}_{\mbs{x}} \\ \mbs{v}_{\mbs{\ell}} \ebm$, $\mbs{t}_{\mbs{x}}^T = \bbm \mbf{t}_{\mbf{x},1}^T & \cdots & \mbf{t}_{\mbf{x},K}^T \ebm$, $\mbs{t}_{\mbs{\ell}}^T = \bbm \mbf{t}_{\mbs{\ell},1}^T & \cdots & \mbf{t}_{\mbs{\ell},V}^T \ebm$, $\mbs{v}_{\mbs{x}}^T = \bbm \mbf{v}_{\mbf{x},1}^T & \cdots & \mbf{v}_{\mbf{x},K}^T \ebm$, $\mbs{v}_{\mbs{\ell}}^T = \bbm \mbf{v}_{\mbs{\ell},1}^T & \cdots & \mbf{v}_{\mbs{\ell},V}^T \ebm$.

Let $\mbs{S}_{||} = \mathrm{null}(\mbs{S})$ be a matrix constructed by a basis of the nullspace of $\mbs{S}$, where $\mbs{S}$ is defined in~\eqref{eq:s}. Then, $\mathrm{span} (\mbs{S}_{||})$ represents the tangent space of the linearized constraints. Let $\mbs{S}_{\perp}$ be a matrix that completes the basis for $\mbs{S}_{||}$, meaning that the square matrix $\bbm \mbs{S}_{||} | \mbs{S}_{\perp} \ebm$ is full rank\footnote{Note that one choice for $\mbs{S}_\perp$ that contains the desired block-diagonal sparsity structure is $\mbs{S}_\perp = \mbs{S}^T$. However, other choices of a block-diagonal $\mbs{S}_\perp$ are just as efficient, and fixing $\mbs{S}_\perp = \mbs{S}^T$ in particular does not simplify any future computations. Rather, this restriction could limit a user from freely constructing $\mbs{S}_\perp$ to resolve any numerical issues that may arise. Thus, in this section, we kept $\mbs{S}_\perp$ as a separate entity from $\mbs{S}^T$, in the same spirit as how the nullspace method is presented in~\cite{nocedal}.}. Since $\mbs{S}$ is block-diagonal, we can choose a block-diagonal construction of $\mbs{S}_{||}$ and $\mbs{S}_{\perp}$ as
\begin{subequations}\label{eq:sqp-st}
\begin{gather}
 \mbs{S}_{||} = \bbm \mbs{S}_{||,\mbs{x}} & \mbf{0} \\ \mbf{0} & \mbs{S}_{||,\mbs{\ell}} \ebm, \quad
 \mbs{S}_{\perp} = \bbm \mbs{S}_{\perp,\mbs{x}} & \mbf{0} \\ \mbf{0} & \mbs{S}_{\perp,\mbs{\ell}} \ebm, \\
 \mbs{S}_{||,\mbs{x}} = \diag(\mbf{S}_{||,\mbf{x},1},\dots,\mbf{S}_{||,\mbf{x},K}), \quad
 \mbs{S}_{||,\mbs{\ell}} = \diag(\mbf{S}_{||,\mbs{\ell},1},\dots,\mbf{S}_{||,\mbs{\ell},V}), \\
 \mbs{S}_{\perp,\mbs{x}} = \diag(\mbf{S}_{\perp,\mbf{x},1},\dots,\mbf{S}_{\perp,\mbf{x},K}), \quad
 \mbs{S}_{\perp,\mbs{\ell}} = \diag(\mbf{S}_{\perp,\mbs{\ell},1},\dots,\mbf{S}_{\perp,\mbs{\ell},V}), \\
 \mbf{S}_{||,\mbf{x},k} = \mathrm{null}(\mbf{S}_{\mbf{x},k}), \quad
 \mbf{S}_{||,\mbs{\ell},j} = \mathrm{null}(\mbf{S}_{\mbs{\ell},j}), \label{eq:sqp-st-sbar}\\
 \mbf{S}_{\perp,\mbf{x},k} = \mathrm{null}(\mbf{S}_{||,\mbf{x},k}^T), \quad
 \mbf{S}_{\perp,\mbs{\ell},j} = \mathrm{null}(\mbf{S}_{||,\mbs{\ell},j}^T).
\end{gather}
\end{subequations}
We decompose $\mbs{t}$ into two components, $\mbs{t}_{\perp}$ and $\mbs{t}_{||}$:
\begin{gather}\label{eq:t-decomp}
 \mbs{t} = \mbs{t}_{\perp} + \mbs{t}_{||}, \quad
 \mbs{t}_\perp = \mbs{S}_{\perp} \mbs{c}_\perp, \quad \mbs{t}_{||} = \mbs{S}_{||} \mbs{c}_{||}.
\end{gather}
Here, $\mbs{t}_{\perp}$ represents the update direction orthogonal to the constraints, and $\mbs{t}_{||}$ represents the update direction tangent to the constrants. $\mbs{c}_\perp$ and $\mbs{c}_{||}$ are, respectively, the coordinates of $\mbs{t}_\perp$ and $\mbs{t}_{||}$ in the basis of $\mbs{S}_{\perp}$ and $\mbs{S}_{||}$. We write $\mbs{c}_{\perp}$ and $\mbs{c}_{||}$ as $\mbs{c}_\perp = \bbm \mbs{c}_{\perp,\mbs{x}} \\ \mbs{c}_{\perp,\mbs{\ell}} \ebm$, $\mbs{c}_{||} = \bbm \mbs{c}_{||,\mbs{x}} \\ \mbs{c}_{||,\mbs{\ell}} \ebm$, $\mbs{c}_{\perp,\mbs{x}}^T = \bbm \mbf{c}_{\perp,\mbf{x},1}^T & \cdots & \mbf{c}_{\perp,\mbf{x},K}^T \ebm$, $\mbs{c}_{\perp,\mbs{\ell}}^T = \bbm \mbf{c}_{\perp,\mbs{\ell},1}^T & \cdots & \mbf{c}_{\perp,\mbs{\ell},V}^T \ebm$, $\mbs{c}_{||,\mbs{x}}^T = \bbm \mbf{c}_{||,\mbf{x},1}^T & \cdots & \mbf{c}_{||,\mbf{x},K} \ebm^T$, $\mbs{c}_{||,\mbs{\ell}}^T = \bbm \mbf{c}_{||,\mbs{\ell},1}^T & \cdots & \mbf{c}_{||,\mbs{\ell},V}^T \ebm$.

To make the coordinates $\mbf{c}_{||,\mbf{x},k}$ and $\mbf{c}_{||,\mbs{\ell},j}$ more interpretable, we can enforce that the choice of $\mbf{S}_{||,\mbf{x},k}$ and $\mbf{S}_{||,\mbs{\ell},j}$ has the identity matrix of the appropriate size on top:
\begin{gather}\label{eq:s-bar-struct}
 \mbf{S}_{||,\mbf{x},k} = \bbm \mbf{1}_{N^\xi} \\ \mbs{*} \ebm, \quad
 \mbf{S}_{||,\mbs{\ell},j} = \bbm \mbf{1}_{N^\psi} \\ \mbs{*} \ebm,
\end{gather}
where the $\mbs{*}$ block consists of any entries such that~\eqref{eq:sqp-st-sbar} is satisfied. Then, each element in $\mbf{c}_{||,\mbf{x},k}$ and $\mbf{c}_{||,\mbs{\ell},j}$ represents, respectively, the update in each dimension of $\mbs{\xi}_k$ and $\mbs{\psi}_j$ tangent to the constraints. This structure will also simplify the extraction of covariances later.

To solve for $\mbs{t}$, we first solve for $\mbs{t}_{\perp}$, then solve for $\mbs{t}_{||}$. From the second equation of \eqref{eq:sqp}, we have $\mbs{S} \mbs{t} = -\mbs{h}$, \\
$\Rightarrow \mbs{S} (\mbs{S}_{\perp} \mbs{c}_{\perp} + \mbs{S}_{||} \mbs{c}_{||}) = -\mbs{h}$,
\begin{gather}\label{eq:solve-for-t-perp}
 \Rightarrow (\mbs{S} \mbs{S}_{\perp}) \mbs{c}_{\perp} = -\mbs{h}.
\end{gather}
We assume that $\mbs{S}$ has full row rank, or else the SQP is not feasible. Note that $\mbs{S} \mbs{S}_{\perp}$ is nonsingular. To see why, notice that $\bbm \mbs{S}_{||} | \mbs{S}_{\perp} \ebm$ has full rank, so the product $\mbs{S} \bbm \mbs{S}_{||} | \mbs{S}_{\perp} \ebm = \bbm \mbs{0} | \mbs{S} \mbs{S}_{\perp} \ebm$ has full row rank, and thus $\mbs{S} \mbs{S}_{\perp}$ is a square nonsingular matrix. Therefore, we can uniquely solve for $\mbs{c}_\perp$ from~\eqref{eq:solve-for-t-perp}.
With our choice of $\mbs{S}_{\perp}$ in~\eqref{eq:sqp-st}, $\mbs{S}\mbs{S}_{\perp}$ also becomes block-diagonal with the same structure as in~\eqref{eq:sqp-st}. We can thus solve for $\mbs{c}_\perp$ in $\mathcal{O}(N^3(K+V))$ time with
\begin{subequations}
\begin{gather}
 (\mbf{S}_{\mbf{x},k}\mbf{S}_{\perp,\mbf{x},k}) \mbf{c}_{\perp,\mbf{x},k} = -\mbf{h}_{\mbf{x},k}, \quad k = 1,\dots,K, \\
 (\mbf{S}_{\mbs{\ell},j}\mbf{S}_{\perp,\mbs{\ell},j}) \mbf{c}_{\perp,\mbs{\ell},j} = -\mbf{h}_{\mbs{\ell},j}, \quad j = 1,\dots,V.
\end{gather}
\end{subequations}
To find $\mbs{t}_{||}$, we premultiply the first equation of~\eqref{eq:sqp} by $\mbs{S}_{||}^T$ to elimininate any $\mbs{S}_{||}^T \mbs{S}^T$ terms:
\begin{subequations}
\begin{gather}
\mbs{S}_{||}^T (\mbs{F} \mbs{t} - \mbs{S}^T \mbs{v}) = \mbs{S}_{||}^T (\mbs{g} + \mbs{S}^T \mbs{\lambda}), \\
\Rightarrow \mbs{S}_{||}^T \mbs{F} (\mbs{S}_{\perp} \mbs{c}_{\perp} + \mbs{S}_{||} \mbs{c}_{||}) = \mbs{S}_{||}^T \mbs{g}, \\
\Rightarrow (\mbs{S}_{||}^T \mbs{F} \mbs{S}_{||}) \mbs{c}_{||} = \mbs{S}_{||}^T \mbs{g} - \mbs{S}_{||}^T \mbs{F} \mbs{S}_{\perp} \mbs{c}_{\perp}, \\
\Rightarrow \mbs{F}_{||} \mbs{c}_{||} = \mbs{g}_{||}, \label{eq:solve-for-t-para}
\end{gather}
\end{subequations}
where $\mbs{F}_{||} = \mbs{S}_{||}^T \mbs{F} \mbs{S}_{||}$ and $\mbs{g}_{||} = \mbs{S}_{||}^T \mbs{g} - \mbs{S}_{||}^T \mbs{F} \mbs{S}_{\perp} \mbs{d}_{\perp}$. Here, $\mbs{F}_{||}$ represents the reduced Hessian~\cite{nocedal}, with its size just being the degrees of freedom of the original unlifted system.

The SQP solution is guaranteed to be a direction of descent if $\mbs{F}_{||}$ is positive definite \cite[p.~452]{nocedal}. Since $\mbs{F}$ is at least positive semidefinite, $\mbs{F}_{||}$ is at least positive semidefinite. We then only require that $\mbs{F}_{||}$ is full rank. This can be interpreted as observability of the lifted constrained system, where the solution of the states and landmarks is unique given the process prior, measurements, and constraints. Unlike in UKL, the constraints contribute to the observability of the system. That is, $\mbs{F}$ may not be full rank and is only positive semidefinite (i.e., the unconstrained system is unobservable), but $\mbs{F}_{||}$ is positive definite (i.e., the constrained system is observable). This happens when the unobservable space in $\mbs{F}$ is projected out by $\mbs{S}_{||}$. We see this commonly in RCKL-SLAM, where the evolution of the lifted features is `moved' from the process model to the nonlinear constraints, and the system is only observable with the constraints in place.

Similar to the case for UKL, the observability of (R)CKL depends on the priors, lifting functions, training data, and the observability of the original system. In our experiments, the rank requirement always holds empirically when the SQP is initialized according to Section~\ref{sec:sqp-init-short}.

Using $\mbs{F}$'s breakdown in~\eqref{eq:slam-f}, $\mbs{F}_{||}$ can be written as
\begin{gather}
 \mbs{F}_{||} = 
 \bbm \mbs{S}_{||,\mbs{x}}^T \mbs{F}_{\mbs{x}} \mbs{S}_{||,\mbs{x}} &
 \mbs{S}_{||,\mbs{x}}^T \mbs{F}_{\mbs{x}\mbs{\ell}} \mbs{S}_{||,\mbs{\ell}} \\
 \mbs{S}_{||,\mbs{\ell}}^T \mbs{F}_{\mbs{x}\mbs{\ell}}^T \mbs{S}_{||,\mbs{x}} &
 \mbs{S}_{||,\mbs{\ell}}^T \mbs{F}_{\mbs{\ell}} \mbs{S}_{||,\mbs{\ell}} \ebm
 =
 \bbm \mbs{F}_{||,\mbs{x}} & \mbs{F}_{||,\mbs{x}\mbs{\ell}} \\ \mbs{F}_{||,\mbs{x}\mbs{\ell}}^T & \mbs{F}_{||,\mbs{\ell}} \ebm,
\end{gather}
where, owing to the block-diagonal structure of $\mbs{S}_{||}$, the SLAM arrowhead sparsity structure is preserved. This is to say, $\mbs{F}_{||,\mbs{x}\mbs{x}}$ is block-tridiagonal and $\mbs{F}_{||,\mbs{\ell}\mbs{\ell}}$ is block-diagonal. The blocks in $\mbs{F}_{||}$ can be computed in parallel in $\mathcal{O}(N^3 KV)$ time.
The computation of $\mbs{g}_{||}$ is similarly efficient, since $\mbs{S}_{||}^T$ is block-diagonal and $\mbs{S}_{||}^T \mbs{F} \mbs{S}_{\perp}$ also preserves the sparsity pattern of $\mbs{F}$. Then, we can use a Cholesky decomposition on $\mbs{F}_{||}$, similar to the one done on $\mbs{F}$ in \eqref{eq:chol}, to efficiently solve for $\mbs{c}_{||}$ in~\eqref{eq:solve-for-t-para}. Again, the complexity is $\mathcal{O}(N^3(V^3 + V^2 K))$ when the lower Cholesky decomposition is used when there are more timesteps than landmarks, but the upper Cholesky decomposition can be used for the other case. Finally, we use $\mbs{c}_\perp$ and $\mbs{c}_{||}$ to form $\mbs{t}_\perp$ and $\mbs{t}_{||}$, and then form the full update variable, $\mbs{t}$, with \eqref{eq:t-decomp}, all in linear time.

To solve for the Lagrange multiplier update, $\mbs{v}$, we premultiply the first equation of~\eqref{eq:sqp} by $\mbs{S}_{\perp}^T$:
\begin{subequations}
\begin{gather}
 \mbs{S}_{\perp}^T (\mbs{F} \mbs{t} - \mbs{S}^T \mbs{v}) = \mbs{S}_{\perp}^T (\mbs{g} + \mbs{S}^T \mbs{\lambda}), \\
 \Rightarrow (\mbs{S}\mbs{S}_{\perp})^T \mbs{v} = \mbs{S}_{\perp}^T (\mbs{F}\mbs{t} - \mbs{g} - \mbs{S}^T \mbs{\lambda}),
\end{gather}
\end{subequations}
where we can solve for $\mbs{v}$ in linear time since the square matrix $\mbs{S}\mbs{S}_{\perp}$ is full rank and block-diagonal.  With this, we have efficiently solved for $\mbs{t}$ and $\mbs{v}$ in time $\mathcal{O}(N^3(V^3 + V^2 K))$.

\subsection{Proof of (R)CKL-SLAM Covariance Extraction}
\label{sec:extract-cov}

Below, we will show that when $\mbf{S}_{||,\mbf{x},k}$ and $\mbf{S}_{||,\mbs{\ell},j}$ have the structures shown in~\eqref{eq:s-bar-struct}, then the batch covariance of the original variables satisfies 
\begin{gather}\label{eq:constr-cov}
 \hat{\mbs{\Sigma}}_{\zeta} = \mbs{F}_{||}^{-1} = (\mbs{S}_{||}^T \mbs{F}  \mbs{S}_{||})^{-1},
\end{gather}
where $\mbs{F}_{||}$ is constructed at the solution $(\mbs{q}^\star,\mbs{\lambda}^\star)$. Since $\mbs{F}_{||}$ has the SLAM arrowhead sparsity structure, we use a similar procedure as for UKL-SLAM to extract the required covariances by finding the corresponding nonzero blocks of $\mbs{F}_{||}$.

We now show~\eqref{eq:constr-cov}. After convergence, we construct the left-hand side of the SQP in~\eqref{eq:sqp} once more, yielding $\mbs{M} = \bbm \mbs{F} & \mbs{S}^T \\ \mbs{S} & \mbf{0} \ebm$. $\mbs{M}$ is termed the Karush–Kuhn–Tucker (KKT) matrix~\cite{boyd-vmls}. Before inverting $\mbs{M}$, note that $\mbs{F}$ may not be invertible. As discussed in Appendix~\ref{sec:solve-sqp-lin-time}, this occurs when only the constrained system is observable, which is most commonly seen in RCKL-SLAM. However, as long as $\mbs{F}_{||}$ is invertible and $\mbs{S}$ has full row rank, then $\mbs{M}$ is invertible. This is because the SQP solution, $\delta \mbs{q}$ and $\delta \mbs{\lambda}$ in~\eqref{eq:sqp}, is unique, as seen in Appendix~\ref{sec:solve-sqp-lin-time} by construction. We now write the inverse of the KKT matrix as $\mbs{M}^{-1} = \bbm \mbs{P}_{\mbf{F}} & \mbs{P}_{\mbf{S}}^T \\ \mbs{P}_{\mbf{S}} & \mbs{P}_{\mbf{Z}} \ebm$, where $\mbs{P}_{\mbf{F}}$, $\mbs{P}_{\mbf{S}}$, and $\mbs{P}_{\mbf{Z}}$ has the same size as $\mbs{F}$, $\mbs{S}$, and the $\mbf{0}$ in $\mbs{M}$, respectively. Then, it is known that $\hat{\mbs{P}}_{\mbs{q}}$, the covariance of the primal variable, satisfies $\hat{\mbs{P}}_{\mbs{q}} = \mbs{P}_{\mbf{F}}$~\cite{constrained-cov}.

We cannot use the Schur complement on $\mbs{M}$ to solve for $\mbs{P}_{\mbf{F}}$ since $\mbs{F}$ is not necessarily invertible. However, computing $\mbs{P}_{\mbf{F}}$ is actually not necessary. When the constraints are satisfied at the minimum-cost solution, any uncertainty as a result of noisy information would only be within the constraint manifold's tangent space~\cite{constrained-cov}. That is, we can write $\mbs{q}^\star = \mbs{S}_{||} \mbs{c}^\star$, where $\mbs{c}^\star$ is the coordinate of $\mbs{q}^\star$ in the basis of $\mbs{S}_{||}$. The solution distribution in the basis of $\mbs{S}_{||}$ satisfies $\mbs{c} \sim \mathcal{N}(\hat{\mbs{c}}, \hat{\mbs{P}}_{\mbs{c}})$, where $\hat{\mbs{c}} = \mbs{c}^\star$ and $\hat{\mbs{P}}_{\mbf{q}} = \mbs{S}_{||} \hat{\mbs{P}}_{\mbs{c}} \mbs{S}_{||}^T$, affirming that the covariance of $\mbs{q}$ is within the space of $\mbs{S}_{||}$. Then, observe that $\hat{\mbs{\Sigma}}_{\zeta} = \hat{\mbs{P}}_{\mbs{c}}$, where $\hat{\mbs{\Sigma}}_{\zeta}$ is the covariance of the original state variables and is exactly the covariance we need. This is because owing to the structure of the bases in~\eqref{eq:s-bar-struct}, the dimensions in $\mbs{c}$ exactly represent the dimensions of the original variables, so the covariance $\mbs{\zeta}$ is equivalent to the covariance of $\mbs{c}$. 

We write out the expression $\mbs{M}\mbs{M}^{-1} = \mbf{1}$, yielding
\begin{gather}\label{eq:finding-m-inv}
\bbm \mbs{F} & \mbs{S}^T \\ \mbs{S} & \mbf{0} \ebm \bbm \mbs{P}_{\mbf{F}} & \mbs{P}_{\mbf{S}}^T \\ \mbs{P}_{\mbf{S}} & \mbs{P}_{\mbf{Z}} \ebm = \bbm \mbf{1} & \mbf{0} \\ \mbf{0} & \mbf{1} \ebm.
\end{gather}
We premultiply first expression of~\eqref{eq:finding-m-inv} by $\mbs{S}_{||}^T$:
\begin{subequations}
\begin{gather}
 \mbs{F} \mbs{P}_{\mbf{F}} + \mbs{S}^T \mbs{P}_{\mbf{S}} = \mbf{1}, \\
 \Rightarrow \mbs{S}_{||}^T \mbs{F} \mbs{S}_{||} \hat{\mbs{P}}_{\mbs{c}} \mbs{S}_{||}^T + \mbs{S}_{||}^T \mbs{S}^T \mbs{P}_{\mbf{S}} = \mbs{S}_{||}^T, \\
 \Rightarrow (\mbs{S}_{||}^T \mbs{F} \mbs{S}_{||}) \hat{\mbs{P}}_{\mbs{c}} \mbs{S}_{||}^T = \mbs{S}_{||}^T, \\
 \Rightarrow \hat{\mbs{P}}_{\mbs{c}} = (\mbs{S}_{||}^T \mbs{F}  \mbs{S}_{||})^{-1},
\end{gather}
\end{subequations}
where the last derivation uses the fact that $\mbs{S}_{||}$ has full column rank. Thus, $\hat{\mbs{\Sigma}}_{\zeta} = \hat{\mbs{P}}_{\mbs{c}} = (\mbs{S}_{||}^T \mbs{F} \mbs{S}_{||})^{-1}$.

\subsection{(R)CKL Estimation with Orientation}
\label{sec:orient}
Suppose the state of a 2D robot is $\mbs{\xi}_k = \bbm \xi_{x,k} & \xi_{y,k} & \xi_{\theta,k} \ebm^T$, where $(\xi_{x,k},\xi_{y,k})$ is its position and $\xi_{\theta,k}$ is its orientation. Since $\xi_{\theta,k}$ is circular, simply lifting with $\mbf{x}_k = \mbf{p}_{\mbs{\xi}}(\mbs{\xi}_k)$ in~\eqref{eq:new-embeddings-x} would not work: $\xi_{\theta,k}$ would be part of the lifted state but we cannot enforce that $\xi_{\theta,k} = \xi_{\theta,k} + 2m\pi$ for $m \in \mathbb{Z}$. Instead, we do a variable transformation on $\mbs{\xi}_k$ as $\mbs{\xi}^'_k = \bbm \xi_{x,k} & \xi_{y,k} & \xi_{\cos \theta,k} & \xi_{\sin \theta, k} \ebm^T$, where the conversion from $\mbs{\xi}_k$ to $\mbs{\xi}^'_k$ is $\xi_{\cos \theta,k} = \cos \xi_{\theta,k}$, $\xi_{\sin \theta,k} = \sin \xi_{\theta,k}$. We introduce the constraint $h_\xi(\mbs{\xi}^'_k) = \xi_{\cos \theta,k}^2 + \xi_{\sin \theta,k}^2 - 1 = 0$. We then select a lifting function $\tilde{\mbf{p}}_{\mbs{\xi}^'}(\mbs{\xi}^'_k)$, and then write the lifted state as $\mbf{x}_k = \bbm \mbs{\xi}^'_k \\ \tilde{\mbf{p}}_{\mbs{\xi}^'}(\mbs{\xi}^'_k) \ebm = \bbm \mbs{\xi}^'_k \\ \tilde{\mbf{x}}_k \ebm$, where the manifold constraint is now the lifting function constraint in addition to the orientation constraint on $\mbs{\xi}^'_k$:
\begin{gather}
 \mbf{x}_k \in \mathcal{X} \; \Rightarrow \; \mbf{h}_{\mbf{x}}(\mbf{x}_k) = \bbm \tilde{\mbf{x}}_k - \tilde{\mbf{p}}_{\mbs{\xi}^'}(\mbs{\xi}^'_k) \\ \xi_{\cos \theta,k}^2 + \xi_{\sin \theta,k}^2 - 1 \ebm = \mbf{0}.
\end{gather}
Note that the additional constraint still acts on each state individually at each timestep, meaning that $\mbs{h}_{\mbf{x}}(\mbs{x})$ is still block-diagonal. Thus, the rest of the procedure for (R)CKL estimation follows as usual. Afterwards, we get $\hat{\xi}_{\theta,k} = \mathrm{atan2}(\hat{\xi}_{\sin \theta,k}, \hat{\xi}_{\cos \theta,k})$.

With the additional constraint, the covariance extraction needs to be slightly modified. $\hat{\mbf{\Sigma}}_{\mbs{\xi},k}$, the covariance of $\mbs{\xi}_k$, is not the same as $\hat{\mbf{P}}_{\mbf{c},k}$, the covariance of $\mbf{x}_k$ in the coordinates of the constraint tangent space defined by ${\mbf{S}}_{||,\mbf{x},k}$. $\mbs{\xi}_k^\prime$ has 4 dimensions but only 3 degree of freedom. As such, we enforce a different structure on ${\mbf{S}}_{||,\mbf{x},k}$ than in~\eqref{eq:s-bar-struct}:
\begin{gather}
 {\mbf{S}}_{||,\mbf{x},k} \bbm 1 & 0 & 0 \\ 0 & 1 & 0 \\ 0 & 0 & {S}_{||,\cos\theta,k} \\ 0 & 0 & {S}_{||,\sin\theta,k} \\ * & * & * \\ \vdots & \vdots & \vdots \ebm,
\end{gather}
where the $*$ entries are any entries so that ${\mbf{S}}_{||,\mbf{x},k} = \mathrm{null}(\mbf{S}_{\mbf{x},k})$ is satisfied. Then, given $\hat{\mbf{P}}_{\mbf{c},k}$ in the form
\begin{gather}
 \hat{\mbf{P}}_{\mbf{c},k} =
 \bbm \hat{\sigma}^2_{\mbf{c},x,k} & \hat{\sigma}^2_{\mbf{c},xy,k} & \hat{\sigma}^2_{\mbf{c},x\theta,k} \\
 \hat{\sigma}^2_{\mbf{c},xy,k} & \hat{\sigma}^2_{\mbf{c},y,k} & \hat{\sigma}^2_{\mbf{c},y\theta,k} \\
 \hat{\sigma}^2_{\mbf{c},x\theta,k} & \hat{\sigma}^2_{\mbf{c},y\theta,k} & \hat{\sigma}^2_{\mbf{c},\theta,k} \ebm,
\end{gather}
the position covariance is still the upper-left $2\times2$ block, 
and the variance of $\xi_{\cos\theta,k}$ and $\xi_{\sin\theta,k}$ can be found with
\begin{gather}
  \hat{\Sigma}_{\mbs{\xi},\cos\theta,k} = {S}_{||,\cos\theta,k}^2 \hat{\sigma}^2_{\mbf{c},\theta,k}, \quad
  \hat{\Sigma}_{\mbs{\xi},\sin\theta,k} = {S}_{||,\sin\theta,k}^2 \hat{\sigma}^2_{\mbf{c},\theta,k},
\end{gather}
either of which can be converted to the orientation variance, $\hat{\Sigma}_{\mbs{\xi},k,\theta}$, through a linear transformation of a Gaussian at the mean, $\hat{\xi}_{\theta,k}$. We see in our experiments that when $\hat{\Sigma}_{\mbs{\xi},\cos\theta,k}$ and $\hat{\Sigma}_{\mbs{\xi},\sin\theta,k}$ are small, using either variance yield the same value for $\hat{\Sigma}_{\mbs{\xi},\theta,k}$.